\documentclass[10pt,twocolumn,letterpaper]{article}

\usepackage[pagenumbers]{cvpr}      % To produce the REVIEW version
\usepackage{times}
\usepackage{epsfig}
\usepackage{graphicx}
\usepackage{amsmath}
\usepackage{amssymb}
\usepackage{booktabs}
\usepackage{kotex}
\usepackage{verbatim}
\usepackage{adjustbox}
\usepackage{rotating}
\usepackage{epsfig}

\usepackage{cuted}
\usepackage{capt-of}
\usepackage{diagbox}
\usepackage{multirow}
\usepackage{bbm}
\usepackage{tabularx}
\usepackage{calc}
\usepackage{algorithm}
\usepackage{algorithmic}
\usepackage{tabu}
\usepackage{boldline}
\usepackage[pagebackref,breaklinks,colorlinks]{hyperref}

\usepackage[capitalize]{cleveref}
\crefname{section}{Sec.}{Secs.}
\Crefname{section}{Section}{Sections}
\Crefname{table}{Table}{Tables}
\crefname{table}{Tab.}{Tabs.}

\DeclareMathOperator*{\argmin}{argmin}

\def\cvprPaperID{5236} % *** Enter the CVPR Paper ID here
\def\confName{CVPR}
\def\confYear{2022}

\begin{document}

\title{ADAS: A Direct Adaptation Strategy for Multi-Target Domain Adaptive Semantic Segmentation}

% \title{MTDA-Net: Multi-Target Domain Adaptation Networks \\
% for Semantic Segmentation}

% \and

\author{%
Seunghun Lee, Wonhyeok Choi, Changjae Kim, Minwoo Choi, and Sunghoon Im\\
Department of Electrical Engineering \& Computer Science, DGIST, Daegu, Korea\\
% DGIST\\
% \email{\{lsh5688, smu06117, chang5434, subminu, sunghoonim \}@dgist.ac.kr}
% \affaddr{DGIST}%
{\tt\small \{lsh5688, smu06117, chang5434, subminu, sunghoonim\}@dgist.ac.kr}
}
\maketitle

\begin{strip}
    \newcommand\x{0.19}
	\centering
	\includegraphics[width=0.95\linewidth]{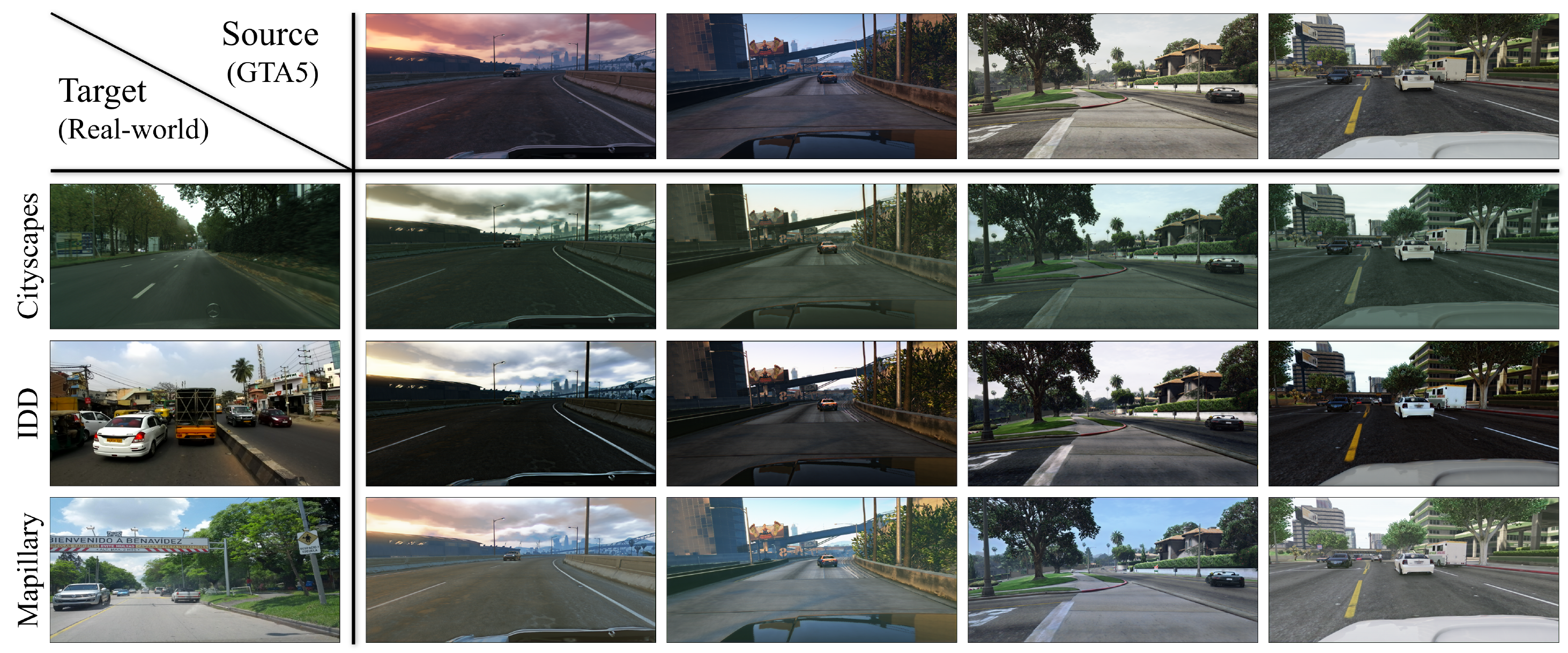}
    \vspace{-3.5mm}
	\captionof{figure}{Multi-target domain transfer results of our single MTDT-Net in the driving scenes.
	The top row and leftmost column represent source domain images and multiple target domain images, respectively.
	The other images are the domain transferred images generated by passing the source image at each column through our MTDT-Net.
	}
% 	\vspace{-3mm}
	\vspace{-4mm}
	\label{fig:teaser} % figure hyperlink caption부분으로 갑니다...
\end{strip}

\begin{abstract}
\vspace{-3mm}
In this paper, we present a direct adaptation strategy (ADAS), which aims to directly adapt a single model to multiple target domains in a semantic segmentation task without pretrained domain-specific models.
To do so, we design a multi-target domain transfer network (MTDT-Net) that aligns visual attributes across domains by transferring the domain distinctive features through a new target adaptive denormalization (TAD) module.
Moreover, we propose a bi-directional adaptive region selection (BARS) that reduces the attribute ambiguity among the class labels by adaptively selecting the regions with consistent feature statistics.
We show that our single MTDT-Net can synthesize visually pleasing domain transferred images with complex driving datasets, and BARS effectively filters out the unnecessary region of training images for each target domain. 
With the collaboration of MTDT-Net and BARS, our ADAS achieves state-of-the-art performance for multi-target domain adaptation (MTDA). 
To the best of our knowledge, our method is the first MTDA method that directly adapts to multiple domains in semantic segmentation.
\end{abstract}

% Introduction
%%%%%%%%% Introduction
\section{Introduction}
\label{sec:intro}
\vspace{-2mm}
Unsupervised domain adaptation (UDA)\cite{ lee2018diverse, shen2019towards, hoffman2018cycada, jeong2021memory, lee2021dranet} aims to alleviate the performance drop caused by the distribution discrepancy between domains.
It is widely utilized in synthetic-to-real adaptation for various computer vision applications that require a large number of labeled data.
Most of the works are designed for single-target domain adaptation (STDA), which allows a single network to adapt to a specific target domain.
It rarely addresses the variability in the real-world, particularly changes in driving region, illumination, and weather conditions in autonomous driving scenarios.
This issue can be tackled by adopting multiple target-specific adaptation models, but this limits the memory efficiency, scalability, and practical utility of embedded autonomous systems.

\begin{figure}[t] 
	\centering
	\begin{tabular}{c@{\hspace{2mm}}c@{\hspace{3mm}}c}
    \includegraphics[height=0.33\linewidth]{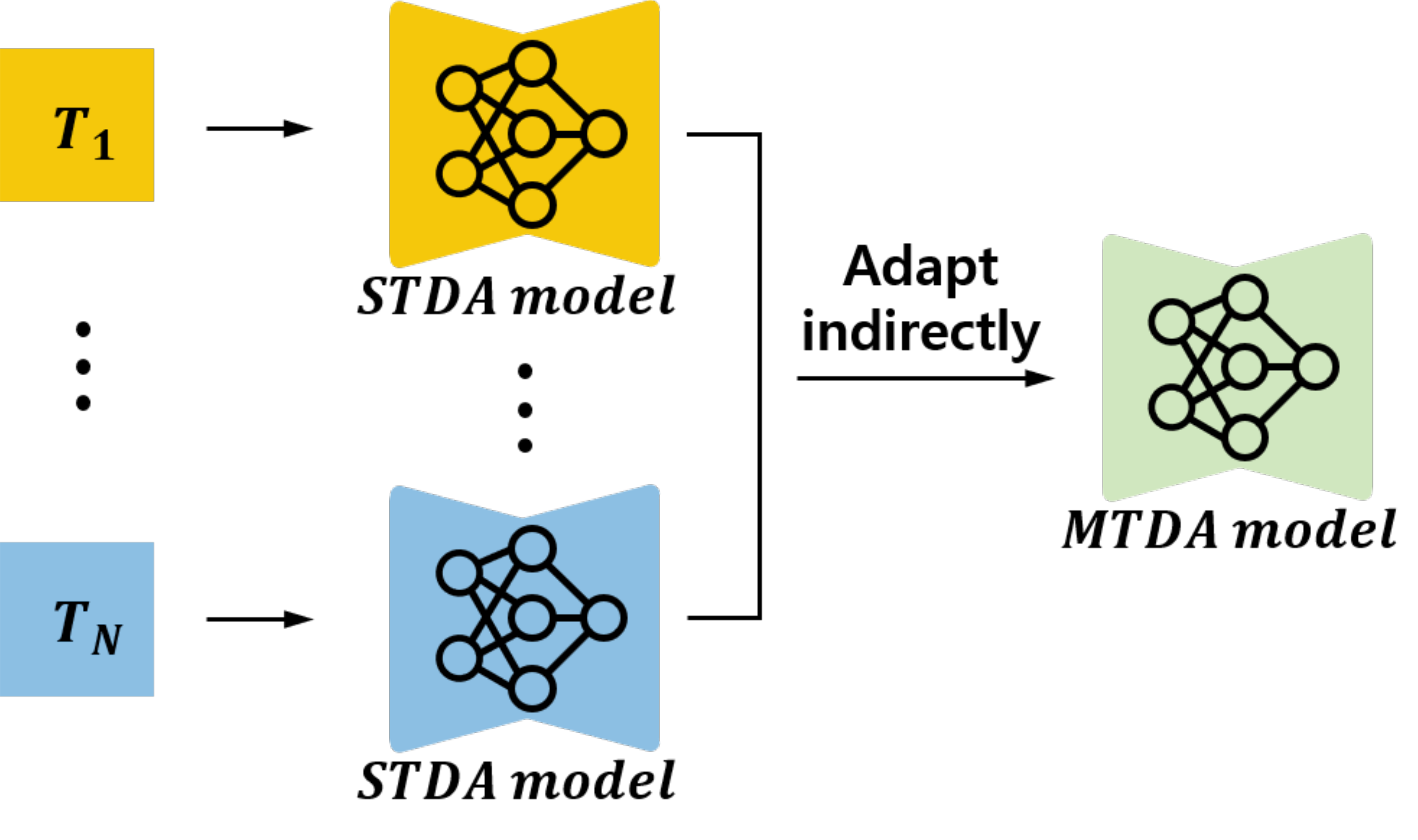} &
    \includegraphics[height=0.33\linewidth]{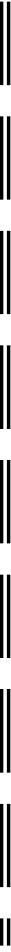} & \includegraphics[height=0.33\linewidth]{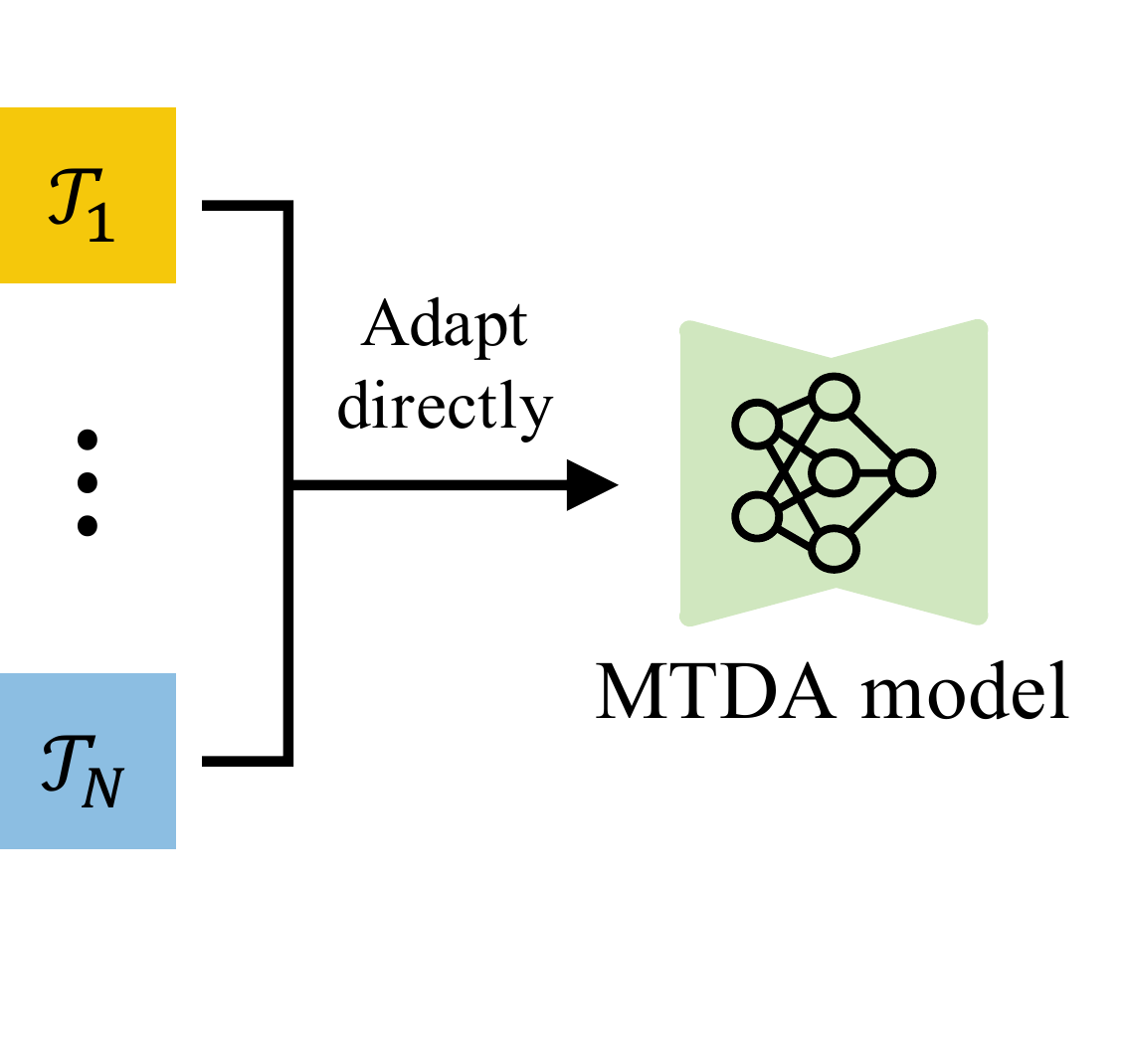} \\
    \small{(a) Existing MTDA method} &
    &
    \small{(b) Ours}
    \end{tabular}
	\caption{
	Illustration of the existing MTDA and our method.
	(a) Conventional MTDA methods pretrain each STDA model then distill the knowledge into a single MTDA model.
	(b) Our ADAS directly adapts multiple target domains.
	}
	\label{fig:Concept}
	\vspace{-3mm}
\end{figure}

Recently, multi-target domain adaptation (MTDA) methods \cite{yu2018multi, peng2019domain, gholami2020unsupervised, nguyen2021unsupervised, isobe2021multi, saporta2021multi} have been proposed, which enables a single model to adapt a labeled source domain to multiple unlabeled target domains.
Most of works train multiple STDA models and then distill the knowledge into a single multi-target domain adaptation network.
% \textcolor{red}{Most of works adopt indirect way to train a single MTDA model using knowledge distillation of pretraiend multiple STDA models.}
% Early work~\cite{yu2018multi} adapts model parameters using an adaptive learning scheme.
Recent approaches\cite{nguyen2021unsupervised, isobe2021multi, saporta2021multi} transfer the knowledge from label predictors as shown in \figref{fig:Concept}-(a).
These methods show impressive results, but their performance can be restricted by the performance of the pretrained models.
Moreover, inaccurate label predictions in the teacher network can degrade model performance, but none of works have deeply investigated them.
To address this problem, we propose A Direct Adaptation Strategy (ADAS) that directly adapts a single model to multiple target domains without pretrained STDA models, as shown in \figref{fig:Concept}-(b). 
Our approach achieves robust multi-domain adaptation by exploiting the feature statistics of training data on multiple domains.
The followings provides a detailed introduction of our sub-modules: Multi-Target Domain Transfer Networks (MTDT-Net) and a Bi-directional Adaptive Region Selection (BARS).

%It consists of Multi-Target Domain Transfer Networks (MTDT-Net) and a Bi-directional Adaptive Region Selection (BARS).
%Our main contributions are summarized as follows:

%All of the above methods that deal with semantic segmentation focus on how to build a unified MTDA model using multiple STDA models, which is an indirect approach that requires to pretrain all the adaptation models of each target domain.
%
%Unlike existing methods, we take another approach called A Direct Adaptation Strategy (ADAS) that directly adapts a single model to multiple target domains without pretrained STDA models. 
%
%
%Fig. 2. shows how our direct adaptation approach (that not using multiple STDA models) is differ from existing methods.

\noindent\textbf{MTDT-Net}
We present a Multi-Target Domain Transfer Network (MTDT-Net) that transfers the distinctive attribute of target domains to a source domain rather than learning all of the target domain distributions.
Our network consists of a novel Target Adaptive Denormalization (TAD) that helps to adapt the statistics of source feature to that of the target feature.
While the existing works on UDA~\cite{liu2016coupled, bousmalis2017unsupervised, murez2018image, hoffman2018cycada, chen2019learning, chen2019crdoco, ma2021coarse} require domain-specific encoders and generators for multi-target domain adaptation, the TAD module enables our single network to adapt to multiple domains. 
\figref{fig:teaser} shows how a single MTDT-Net can efficiently synthesize visually pleasing domain transferred images.
\definecolor{pink}{RGB}{247, 35, 235}
\definecolor{purple}{RGB}{130, 65, 130}
\begin{figure}[t]
    \newcommand\h{0.23}
	\centering
	\begin{tabular}{c@{\hspace{2mm}}c}
	GTA5 $\to$ Cityscapes &
	Cityscapes
	\\
    \includegraphics[height=\h\linewidth]{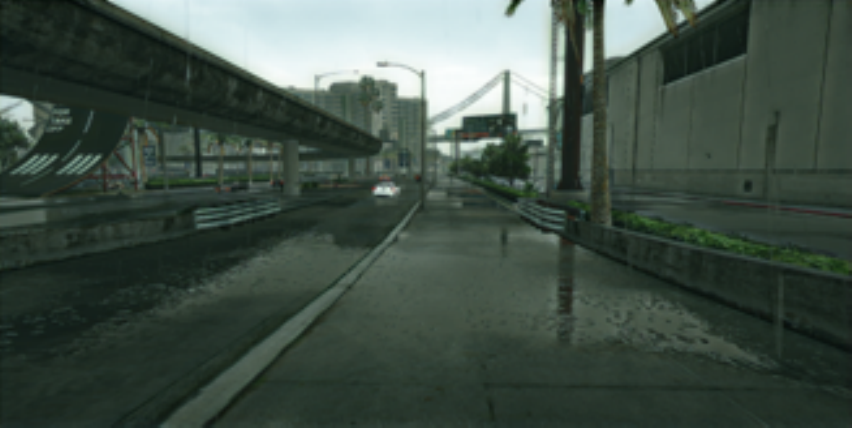} &
    \includegraphics[height=\h\linewidth]{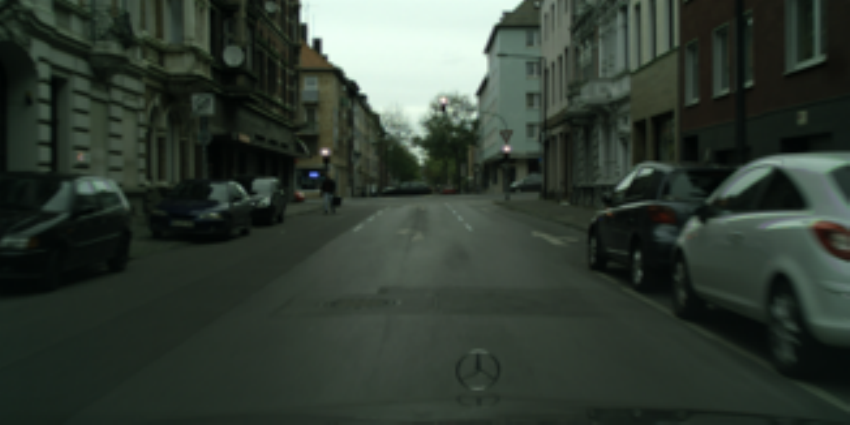}
    \\
    \small{(a) Domain transferred image}&
    \small{(b) Target image}
    \vspace{1mm}
    \\
    \includegraphics[height=\h\linewidth]{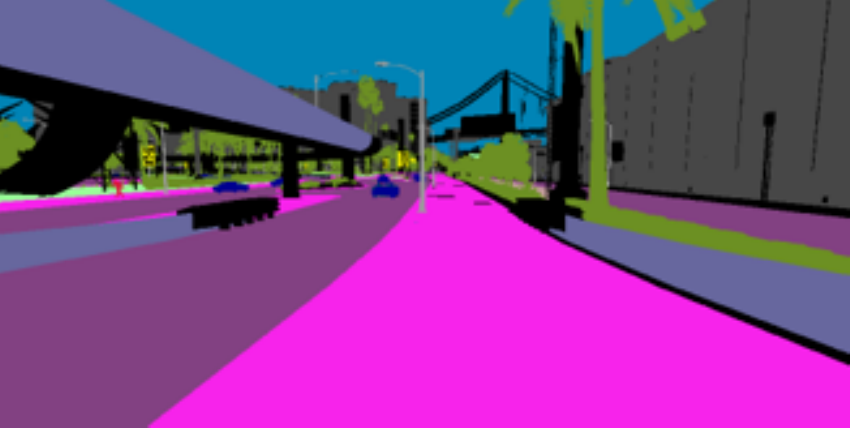} &
    \includegraphics[height=\h\linewidth]{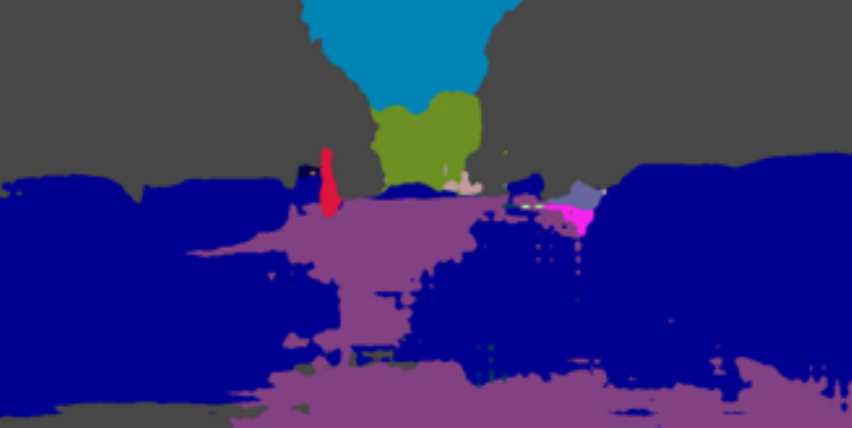}
    \\
    \small{(c) Ground-truth label of (a)}&
    \small{(d) Pseudo label of (b)}
    \vspace{1mm}
    \\
    \includegraphics[height=\h\linewidth]{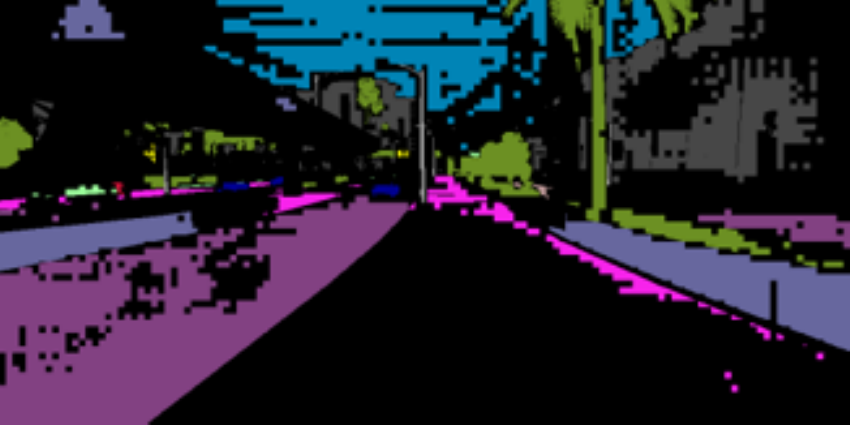} &
    \includegraphics[height=\h\linewidth]{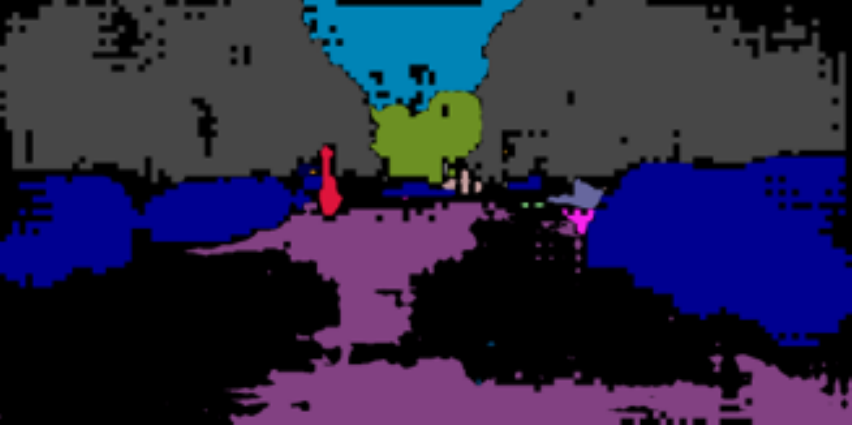}
    \\
    \small{(e) Selected region of (c)}&
    \small{(f) Selected region of (d)}
    \\
   
    \end{tabular}
	\caption{Examples of the regions with similar attributes but different labels (c) {\footnotesize(\textcolor{purple}{\textbf{purple}}: road, \textcolor{pink}{\textbf{pink}}: sidewalk)},
	and the noisy prediction (d). The black regions in (e) and (f) are regions filtered by BARS.}
	% a : similar appearnace between domain transferred image and target image
	\label{fig:BARS_example}
	\vspace{-3mm}
\end{figure}

\noindent\textbf{BARS} 
Although the visual attributes across domains are well-aligned, there are still some attribute ambiguities among the class labels in semantic segmentation.
The ambiguity is usually observed on the regions with similar attributes but different label, such as the sidewalks in GTA5 and the roads in Cityscapes as shown in \figref{fig:BARS_example}-(a),(c).
This confuses the model finding the accurate decision boundary.
Moreover, the predictions from target domains usually have noisy labels leading to inaccurate training of the task network, as shown in \figref{fig:BARS_example}-(b),(d).
To solve these issues, we propose a Bi-directional Adaptive Region Selection (BARS), which alleviates the confusion.
It adaptively selects the regions with consistent feature statistics as shown in~\figref{fig:BARS_example}-(e).
It can also select the pseudo label of the target images for our self-training scheme, as shown in~\figref{fig:BARS_example}-(f). 
We show that BARS allows the task network to perform robust training and achieve the improved performance.

To the best of our knowledge, our multi-target domain adaptation method is the first approach that directly adapts the task network to multiple target domains without pretrained STDA models in semantic segmentation. 
The extensive experiments show that the proposed method achieves state-of-the-art performance on semantic segmentation task.
At the end, we demonstrate the effectiveness of the proposed MTDT-Net and BARS.

\section{Related Work}
\label{sec:related}

\subsection{Domain Transfer} 
With the advent of generative adversarial networks (GANs)\cite{goodfellow2014generative}, the adversarial learning has shown promising results not only in photorealistic image synthesis\cite{radford2015unsupervised, arjovsky2017wasserstein, miyato2018spectral, mao2017least, karras2017progressive, karras2019style, brock2018large, karras2020analyzing} but also in domain transfer\cite{isola2017image, zhu2017unpaired, liu2017unsupervised, taigman2016unsupervised, huang2018multimodal, lee2018diverse, liu2019few, shen2019towards, hoffman2018cycada, jeong2021memory, lee2021dranet}.
The traditional domain transfer methods rely on adversarial learning\cite{isola2017image, zhu2017unpaired, liu2017unsupervised, taigman2016unsupervised} or the typical style transfer method\cite{gatys2016image, ulyanov2016instance, dumoulin2016learned, nam2018batch, huang2017arbitrary}.
Afterwards, the studies on feature disentanglement~\cite{huang2018multimodal, lee2018diverse, liu2019few, park2020swapping, lee2021dranet} present a model that can apply various styles by appropriately utilizing disentangled features that are separately encoded as content and style.
Recent works\cite{richter2021enhancing, zhu2020sean} have tackled more in-depth domain transfer problems.
Richter \etal\cite{richter2021enhancing} propose a rendering-aware denormalization (RAD) that constructs style tensors by using the abundant condition information from G-buffers, and show high fidelity domain transfer in a driving scene.
Zhu \etal\cite{zhu2020sean} propose a semantic region-wise domain transfer model by extracting a style vector for each semantic region.

\subsection{Unsupervised Domain Adaptation for Semantic Segmentation}
Traditional feature-level adaptation methods \cite{hoffman2016fcns, luo2019significance, luo2019taking, tsai2018learning, vu2019advent} aim to align the source and target distribution in feature space.
Most of them \cite{hoffman2016fcns, luo2019significance, luo2019taking} adopt adversarial learning with the intermediate features of the segmentation network, and the others \cite{tsai2018learning, vu2019advent} directly apply adversarial loss to output prediction.
Pixel-level adaptation methods\cite{liu2016coupled, bousmalis2017unsupervised, murez2018image, lee2021dranet} reduce the domain gap in the image-level by synthesizing target-styled images. 
Several works \cite{hoffman2018cycada, chen2019crdoco, chen2019learning} adopt both feature-level and pixel-level methods.
Another direction of UDA \cite{zou2018unsupervised, zheng2021rectifying, li2019bidirectional,ma2021coarse, zhang2021prototypical} is to take a self-supervised learning approach for dense prediction tasks, such as semantic segmentation.
Some works \cite{zou2018unsupervised, zheng2021rectifying, li2019bidirectional} obtain high confidence labels measured by the uncertainty of target prediction, and use them as pseudo ground-truth.
The others \cite{ma2021coarse, zhang2021prototypical} use proxy features by extracting the centroid of the intermediate features of each class to remove uncertain regions in the pseudo-label.

\subsection{Multi-Target Domain Adaptation}
Early studies on MTDA have tackled classification tasks using adaptive learning of a common model parameter dictionary~\cite{yu2018multi}, domain-invariant feature extraction~\cite{peng2019domain}, or knowledge distillation~\cite{nguyen2021unsupervised}.
Recently, using MTDA on more high-level vision tasks such as semantic segmentation \cite{isobe2021multi,saporta2021multi} has become an interesting and challenging research topic.
These works employ knowledge distillation to transfer the knowledge of domain specific teacher models to a domain-agnostic student model. 
For more robust adaptation, Isobe \etal \cite{isobe2021multi} enforce the weight regularization to the student network and Saporta \etal \cite{saporta2021multi} use a shared feature extractor that constructs a common feature space for all domains.
In this work, we present a more efficient and simpler method that handles multiple domains using a unified architecture without teacher networks or weight regularization.

% Yu \etal\cite{yu2018multi} present a model parameter adaptation to get target model parameters by sparse dictionary learning. 
% Peng \etal \cite{peng2019domain} propose an architecture to extract domain-invariant features. 
% Mutual information minimization\cite{belghazi2018mutual} was exploited to disentangle the feature to domain-invariant, domain-specific and class-irrelevant features.
% Domain-invariant and class-irrelevant features are constructed by the adversarial learning with the classification network and domain classifier.
% task로 나누지말고 이전 방법들과 최근 방법들로 나눠서 설명. segmentation 논문들도 쫌더 까는 형식으로, 이런방식으로 밖에 풀어내지 못했다. 
%The most recent works \cite{isobe2021multi, saporta2021multi} of MTDA address more high-level vision task such as semantic segmentation. 
%However, learning from domain specific teacher models that have different feature spaces is likely to induce student model to create complicated decision boundaries. 

% Method
\begin{figure*}[t] 
	\centering
    \includegraphics[width=0.99\linewidth]{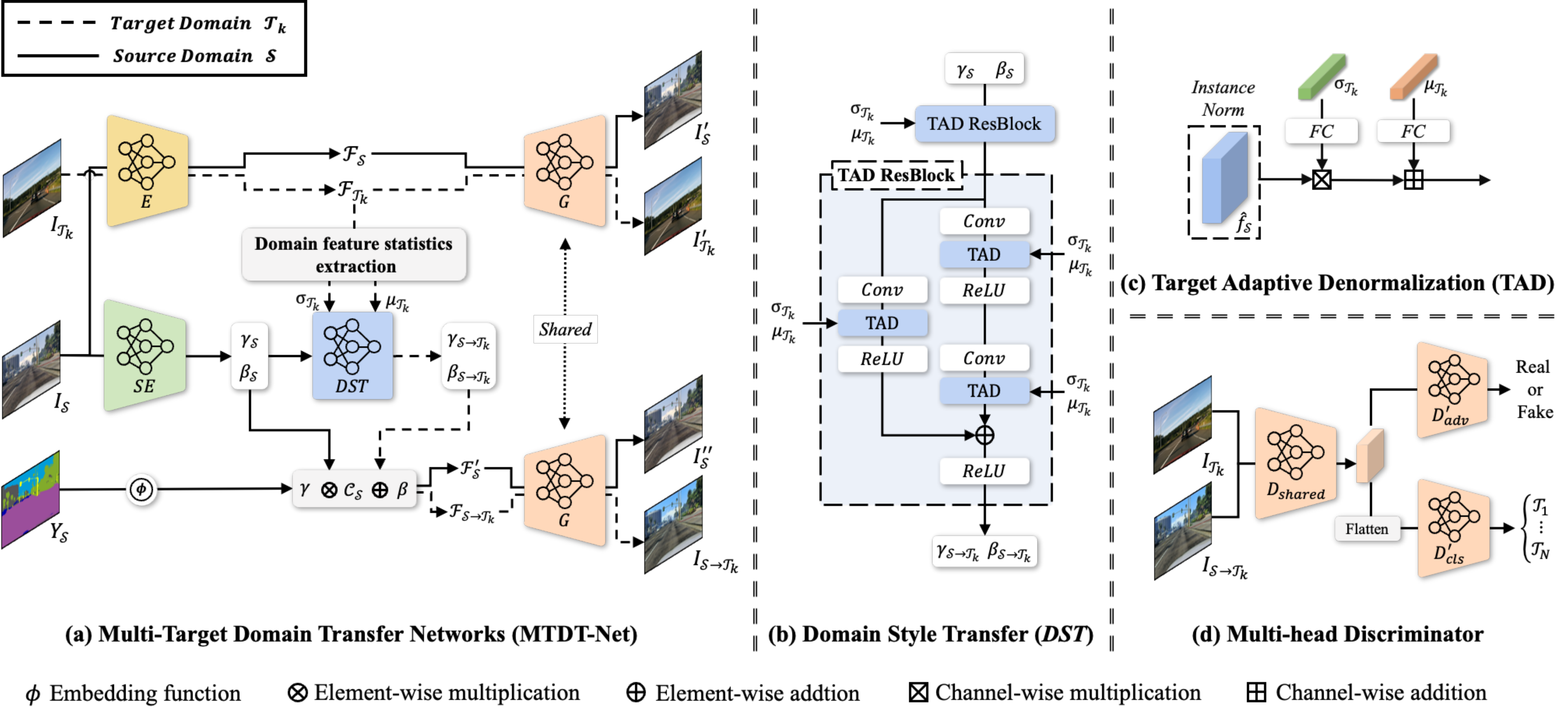}
	\vspace{-2mm}
    % 각 notation에 대한 설명. 네트워크, 이미지. 왼쪽, 가운데, 오른쪽 순서로.
	\caption{Overview of the proposed MTDT-Net.
	(a) MTDT-Net consists of an encoder $E$, a style encoder $SE$, a domain style transfer network $DST$ and a generator $G$. Given a source image, label map $I_\mathcal{S}, Y_\mathcal{S}$, and target images $I_{\mathcal{T}_k}$, MTDT-Net aims to produce domain transferred image $I_{\mathcal{S}\to \mathcal{T}_k}$.
	The other reconstructed images $I'_\mathcal{S}, I'_{\mathcal{T}_k}, I''_\mathcal{S}$ are auxiliary outputs generated only during the training process.
	(b) $DST$ consists of two TAD residual blocks (ResBlock). The TAD module is followed by each convolutional layer, given the channel-wise statistics of target domains $\mu_{\mathcal{T}_k}, \sigma_{\mathcal{T}_k}$.
	(c) TAD transfers the target domain with $\mu_{\mathcal{T}_k}, \sigma_{\mathcal{T}_k}$ by statistics modulation.
	(d) The multi-head discriminator predicts which domain the image is from, as well as determines whether the image is real or fake.
	Note that, for the sake of brevity, we illustrate a single target domain setting, but our model deals with multi-target domain adaptation.
	}
	\label{fig:overview}
	\vspace{-3mm}
\end{figure*}

%%%%%%%%% Method
\section{A Direct Adaptation Strategy (ADAS)}
\label{sec:method}

In this section, we describe our direct adaptation strategy for multi-target domain adaptive semantic segmentation.
We have a labeled source dataset $\mathcal{S}=\{I_\mathcal{S}, Y_\mathcal{S} \}$ and $N$ unlabeled target datasets $\mathcal{T}_k=\{I_{\mathcal{T}_k}\}, k\in \{1,...,N\}$, where $I$ and $Y$ are the image and the ground-truth label, respectively.
The goal of our approach is to directly adapt a segmentation network $T$ to multiple target domains without training STDA models.
Our strategy contains two sub-modules: a multi-target domain transfer network (MTDT-Net), and a bi-directional adaptive region selection (BARS).
We describe the details of MTDT-Net and BARS in \secref{sec:MTDT-Net} and \secref{sec:BARS}, respectively.
% For the first part, we design a Multi-Target Domain Transfer Network (MTDT-Net), which transfers the source image $I_S$ to N target domains $I_{S\to T_1}, ..., I_{S\to T_N}$, which is later used for the attribute alignment.
% For the domain transfer, we use the autoencoder structure to build the image feature space, and we consider the image feature composed of the scene structure and detail representation which we call the content and style in this paper.
% For the second part, we devise a Bi-directional Adaptive Region Selection (BARS) that filters out the pixel-wise outliers both of the domain transferred image and the target image.

\subsection{Multi-Target Domain Transfer Network (MTDT-Net)}
\label{sec:MTDT-Net}
The overall pipeline of MTDT-Net is illustrated in \figref{fig:overview}-(a).
The network consists of an encoder $E$, a generator $G$, a style extractor $SE$, a domain style transfer network $DST$.
To build an image feature space, we adopt a typical autoencoder structure with the encoder $E$ and the generator $G$.
Given the source and target images $I_\mathcal{S}, I_{\mathcal{T}_1}, ..., I_{\mathcal{T}_N}$, the encoder $E$ extracts the individual features $\mathcal{F}_\mathcal{S}, \mathcal{F}_{\mathcal{T}_1}, ... \mathcal{F}_{\mathcal{T}_N}$ that are later passed through the generator $G$ to reconstruct the original input images $I'_\mathcal{S}, I'_{\mathcal{T}_1}, ..., I'_{\mathcal{T}_N}$ as follows:
\begin{gather}
\begin{split}
\mathcal{F}_{\mathcal{S}} = E(I_\mathcal{S}), ~~ \mathcal{F}_{\mathcal{T}_k} = E(I_{\mathcal{T}_k}), \\
I'_\mathcal{S} = G(\mathcal{F}_\mathcal{S}), ~~ I'_{\mathcal{T}_k} = G(\mathcal{F}_{\mathcal{T}_k}).
\end{split}
\label{eq:direct_recon}
\end{gather}

We extract the style tensors $\gamma_\mathcal{S}$, $\beta_\mathcal{S}$ of the source image through the style encoder $SE$, and the content tensor $\mathcal{C}_\mathcal{S}$ from the segmentation label only in the source domain through an $1\times1$ convolutional layer $\phi(\cdot)$ as follows:
\begin{gather}
\begin{split}
\{\gamma_\mathcal{S}, \beta_\mathcal{S}\} = SE(I_\mathcal{S}),~\mathcal{C_S}=\phi(Y_\mathcal{S}). %\\
%I''_S = G(\mathcal{F}'_S).
\end{split}
\label{eq:indirect_recon1}
\end{gather}

We assume that the image features are composed of the scene structure and detail representation, which we call the content feature $\mathcal{C}_\mathcal{S}$ and style feature $\gamma_\mathcal{S}$, $\beta_\mathcal{S}$ as follows:
\begin{gather}
\begin{split}
I''_\mathcal{S} = G(\mathcal{F}'_\mathcal{S}),~\mathcal{F}'_\mathcal{S} = \gamma_\mathcal{S} \mathcal{C_S}  + \beta_\mathcal{S},
\end{split}
\label{eq:indirect_recon}
\end{gather}
where the source image feature $\mathcal{F}'_\mathcal{S}$ is passed through generator $G$ to obtain the reconstructed input image $I''_\mathcal{S}$.
The synthesized images $I'_\mathcal{S}, I'_{\mathcal{T}_k}, I''_\mathcal{S}$ are auxiliary outputs to be utilized for network training.
The goal of our network is to generate a domain transferred image $I_{\mathcal{S}\to{\mathcal{T}_k}}$ using the same generator $G$ as follows:
\begin{gather}
\begin{split}
I_{\mathcal{S}\to{\mathcal{T}_k}} = G(\mathcal{F}_{\mathcal{S}\to{\mathcal{T}_k}}),
~\mathcal{F}_{\mathcal{S}\to{\mathcal{T}_k}} = \gamma_{\mathcal{S}\to{\mathcal{T}_k}} \mathcal{C_S}  + \beta_{\mathcal{S}\to{\mathcal{T}_k}},
\end{split}
\label{eq:transfer}
\end{gather}
where $\mathcal{F}_{\mathcal{S}\to{\mathcal{T}_k}}$ is the domain transferred features, which is composed of the source content $\mathcal{C_S}$ and the $k$-th target domain style features $\gamma_{\mathcal{S}\to{\mathcal{T}_k}}, \beta_{\mathcal{S}\to{\mathcal{T}_k}}$.

To obtain the target domain style tensors, we design a domain style transfer network ($DST$) which transfers the source style tensors $\gamma_\mathcal{S}, \beta_\mathcal{S}$ to the target style tensors $\gamma_{\mathcal{S}\to \mathcal{T}_k}, \beta_{\mathcal{S}\to \mathcal{T}_k}$ as follows:
\begin{gather}
\begin{split}
\gamma_{\mathcal{S}\to{\mathcal{T}_k}}, \beta_{\mathcal{S}\to{\mathcal{T}_k}} = DST(\gamma_\mathcal{S},\beta_\mathcal{S}, \mu_{\mathcal{T}_k}, \sigma_{\mathcal{T}_k}),
\end{split}
\label{eq:DST}
\end{gather}
where the channel-wise mean $\mu_{\mathcal{T}_k}$ and variance $(\sigma_{\mathcal{T}_k})^2$ vectors encode the $k$-th target domain feature statistics computed by the cumulative moving average (CMA) algorithm and Welford's online algorithm \cite{welford1962note} described in \algref{alg:CMA}.
The $DST$ in \figref{fig:overview}-(b) consists of two TAD ResBlock built with a series of convolutional layer, our new Target-Adaptive Denormalization (TAD), and ReLU.
TAD is a conditional normalization module that modulates the normalized input with learned scale and bias similar to SPADE\cite{park2019semantic} and RAD\cite{richter2021enhancing} as shown in \figref{fig:overview}-(c).
We pass the standard deviation $\sigma_{\mathcal{T}_k}$ and the target mean $\mu_{\mathcal{T}_k}$ through each fully connected (FC) layer and use them as scale and bias as follows:
\begin{gather}
\begin{split}
\text{TAD}(\hat{f}_\mathcal{S}, \mu_{\mathcal{T}_k}, \sigma_{\mathcal{T}_k}) = FC(\sigma_{\mathcal{T}_k})\hat{f}_\mathcal{S} + FC(\mu_{\mathcal{T}_k}),
\end{split}
\label{eq:TAD}
\end{gather}
where $\hat{f}_\mathcal{S}$ is the instance-normalized \cite{ulyanov2016instance} input to TAD.
For adversarial learning with multiple target domains, we adopt a multi-head discriminator composed of an adversarial discriminator $D_{adv}=D'_{adv}\circ D_{shared}$ and a domain classifier $D_{cls}=D'_{cls}\circ D_{shared}$ as shown in \figref{fig:overview}-(d). 
%The two groups of networks $\mathcal{G}=\{ E, SE, DST, G, \phi \}$ and $\mathcal{D}=\{ D_{adv}, D_{cls} \}$ are trained with the following losses.

Each group of networks $\mathcal{G}=\{ E, SE, DST, G, \phi \}$ and $\mathcal{D}=\{ D_{adv}, D_{cls} \}$ is trained by minimizing the following losses, $\mathcal{L}^{\mathcal{G}}$ and $\mathcal{L}^{\mathcal{D}}$, respectively:
\begin{gather}
\begin{split}
    \mathcal{L}^{\mathcal{D}} = -\mathcal{L}_{adv} + \mathcal{L}_{cls}^{\mathcal{D}},\\
    \mathcal{L}^{\mathcal{G}} = \mathcal{L}_{rec} + \mathcal{L}_{per} + \mathcal{L}_{adv} + \mathcal{L}_{cls}^{\mathcal{G}}.
\end{split}
\label{eq:full_obj}
\end{gather}

\noindent\textbf{Reconstruction Loss} 
We impose L1 loss on the reconstructed images $I'_\mathcal{S}, I'_{\mathcal{T}_k}, I''_\mathcal{S}$ to build an image feature space: 
\begin{gather}
\begin{split}
\mathcal{L}_{rec} = \mathcal{L}_1(I_\mathcal{S}, I'_\mathcal{S}) + \mathcal{L}_1(I_\mathcal{S}, I''_\mathcal{S}) + \sum_{k=1}^N \mathcal{L}_1(I_{\mathcal{T}_k}, I'_{\mathcal{T}_k}).
\end{split}
\label{eq:rec}
\end{gather}

\noindent\textbf{Adversarial Loss}
We apply the patchGAN \cite{isola2017image} discriminator $D_{adv}$ to impose an adversarial loss on the domain transferred images and the corresponding target images:
\begin{gather}
\begin{split}
    \mathcal{L}_{adv}
    & = \sum_{k=1}^{N} \Big( \mathbb{E}_{I_{\mathcal{T}_k}} \left[ \log D_{adv}(I_{\mathcal{T}_k}) \right] \\
    &+ \mathbb{E}_{I_{\mathcal{S}\to {\mathcal{T}_k}}} \left[1 - \log D_{adv}(I_{\mathcal{S}\to {\mathcal{T}_k}}) \right] \Big). \\
\end{split}
\label{eq:adv}
\end{gather}

\begin{algorithm}
\caption{Domain feature statistics extraction}
\label{alg:CMA}
\begin{algorithmic}[1]
 \renewcommand{\algorithmicrequire}{\textbf{Input: }}
 \renewcommand{\algorithmicensure}{\textbf{Update: }}
 \REQUIRE $\mathcal{F}_{\mathcal{T}_k} \in \mathbb{R}^{H\times W\times C}, k\in \{1,...,N \}$
 \ENSURE  $\mu_{\mathcal{T}_k}, (\sigma_{\mathcal{T}_k})^2 \in \mathbb{R}^{C}$
 \\ \% \textit{1. Initialization}
  \FOR {$k=1 ~to~ N$} 
  \STATE $M_{\mathcal{T}_k} = 0$, $S_{\mathcal{T}_k} = 0 $ \textit{ \hfill//$M_{ \mathcal{T}_k}, S_{\mathcal{T}_k} \in \mathbb{R}^{H\times W\times C}$}
  \ENDFOR
 \\ \% \textit{2. Online update
  \hfill //$N_{update}$ is \# of update iterations}
  \FOR {$n = 0 ~to~ {N_{update}}$}
%   \WHILE {Training}
  \FOR {$k=1 ~to~ N$}
  \STATE $\mu_{n}^{\mathcal{T}_k} \xleftarrow{} \frac{1}{HW} \sum_{i=0}^{H-1}\sum_{j=0}^{W-1}{M_{\mathcal{T}_k}{(i,j)}}$
  \STATE $M_{\mathcal{T}_k} \xleftarrow{} M_{\mathcal{T}_k} + ({{\mathcal{F}_{\mathcal{T}_k} - M_{\mathcal{T}_k}})/({n+1})}$
  \STATE $\mu_{n+1}^{\mathcal{T}_k} \xleftarrow{} \frac{1}{HW} \sum_{i=0}^{H-1}\sum_{j=0}^{W-1}{M_{\mathcal{T}_k}(i,j)}$
  \STATE ${\Tilde{\mu}}_{n}^{\mathcal{T}_k}, {\Tilde{\mu}}_{n+1}^{\mathcal{T}_k}$ $\leftarrow{}$ expand $\mu_{n}^{\mathcal{T}_k},\mu_{n+1}^{\mathcal{T}_k}$ to $\mathbb{R}^{H\times W\times C}$ 
  \IF {$n = 0$}
  \STATE $S_{\mathcal{T}_k} \xleftarrow{} (\mathcal{F}_{\mathcal{T}_k}-{\Tilde{\mu}}_{n+1}^{\mathcal{T}_k})^2$
  \ELSE
  \STATE $S_{\mathcal{T}_k} \xleftarrow{} S_{\mathcal{T}_k} + (\mathcal{F}_{\mathcal{T}_k}-{\Tilde{\mu}}_{n}^{\mathcal{T}_k})(\mathcal{F}_{\mathcal{T}_k}-{\Tilde{\mu}}_{n+1}^{\mathcal{T}_k})$
%   \ENDIF
%   \IF{$n>0$}
  \STATE
  $\mu_{\mathcal{T}_k} \xleftarrow{} \mu^{\mathcal{T}_k}_{n+1}$
  \STATE
  $(\sigma_{\mathcal{T}_k})^2 \xleftarrow{} \frac{1}{nHW} \sum_{i=0}^{H-1}\sum_{j=0}^{W-1}{S_{\mathcal{T}_k}(i,j)}$
  \ENDIF
  \ENDFOR
  \ENDFOR
% \RETURN $\mu_{n+1}^{\mathcal{T}_k}, (\sigma^{\mathcal{T}_k})^2$
\end{algorithmic}
\end{algorithm}

\noindent\textbf{Domain Classification Loss}
We build the domain classifier $D_{cls}$ to classify the domain of the input images.
We impose the cross-entropy loss with the target images for $\mathcal{D}$ and with the domain transferred images for $\mathcal{G}$:
\begin{gather}
\begin{split}
    \mathcal{L}_{cls}^{\mathcal{D}}
    & = -\sum_{k=1}^{N} t_k\log D_{cls}(I_{\mathcal{T}_k}), \\
    \mathcal{L}_{cls}^{\mathcal{G}}
    & = -\sum_{k=1}^{N} t_k\log D_{cls}(I_{\mathcal{S} \to {\mathcal{T}_k}}),
\end{split}
\label{eq:cls}
\end{gather}
where $t_k \in \mathbb{R}^N$ is the one-hot encoded class label of the target domain $\mathcal{T}_k$.

\noindent\textbf{Perceptual Loss}
We impose a perceptual loss \cite{johnson2016perceptual} widely used for domain transfer as well as style transfer \cite{dumoulin2016learned, huang2017arbitrary}:
\begin{gather}
\begin{split}
    \mathcal{L}_{per} = \sum_{k=1}^{N} \sum_{l \in L} || P_l(I_{\mathcal{S}}) - P_l(I_{\mathcal{S} \to {\mathcal{T}_k}}) ||_2^2,
\end{split}
\label{eq:per}
\end{gather}
where the set of layers $L$ is the subset of the perceptual network $P$.

% \noindent\textbf{Full Objective}
% We train each group of networks $\mathcal{D}$, $\mathcal{G}$ by minimizing the following losses, $\mathcal{L}_{\mathcal{D}}$ and $\mathcal{L}_{\mathcal{G}}$, respectively:
% \begin{gather}
% \begin{split}
%     \mathcal{L}_{\mathcal{D}} = -\mathcal{L}_{adv} + \mathcal{L}_{cls}^{\mathcal{D}},\\
%     \mathcal{L}_{\mathcal{G}} = \mathcal{L}_{rec} + \mathcal{L}_{per} + \mathcal{L}_{adv} + \mathcal{L}_{cls}^{\mathcal{G}}.
% \end{split}
% \label{eq:full_obj}
% \end{gather}

\begin{figure}[t] 
	\centering
	\begin{tabular}{c}
    \includegraphics[width=0.99\linewidth]{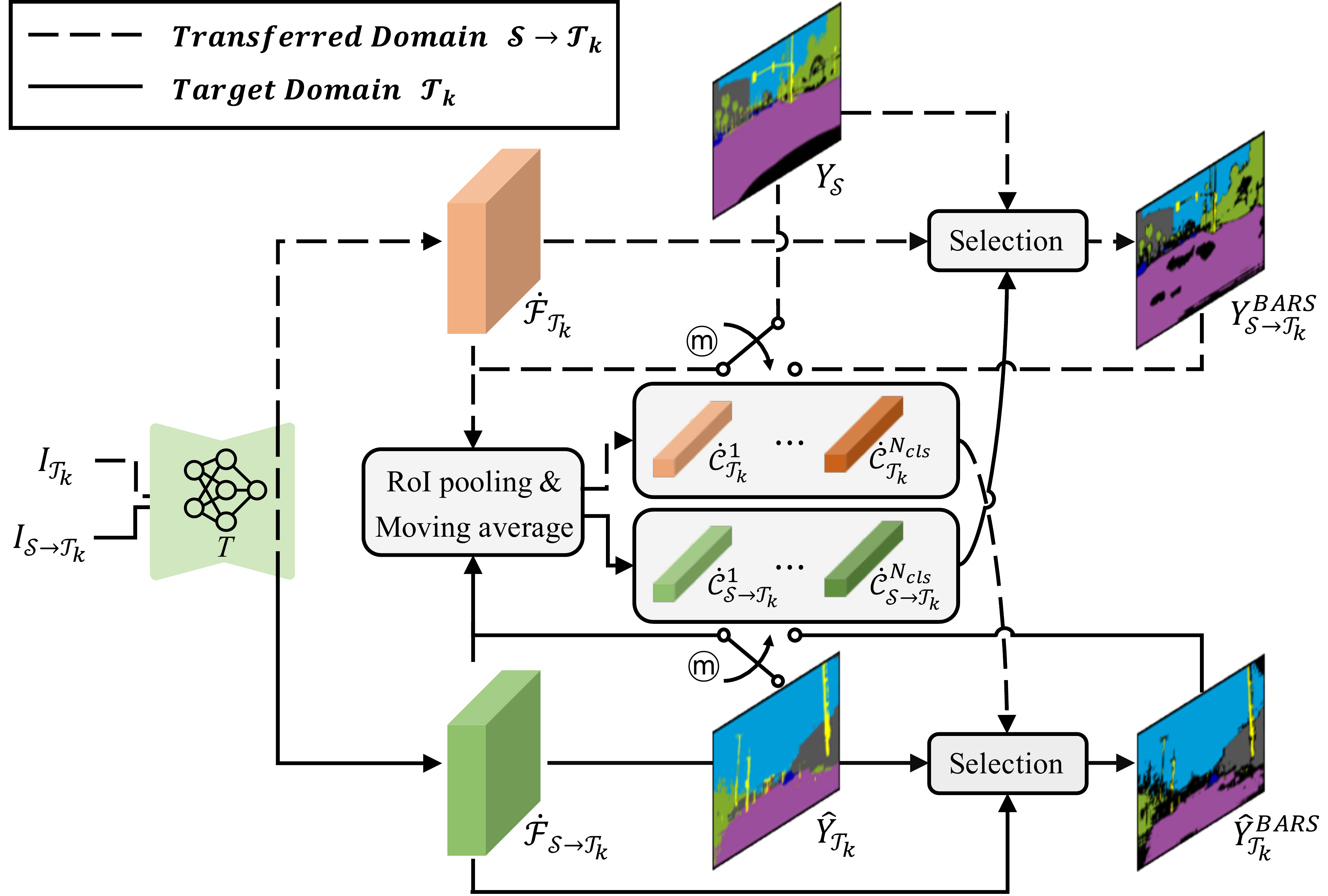} 
    \end{tabular}
	\caption{Overview of BARS.
	For each class $c\in \{1,...,N_{cls} \}$,
	BARS extracts the centroids $\dot{\mathcal{C}}^{c}_{\mathcal{S}\to{\mathcal{T}_k}}, \dot{\mathcal{C}}^{c}_{\mathcal{T}_k}$ from the intermediate features $\dot{\mathcal{F}}_{\mathcal{S}\to{\mathcal{T}_k}}, \dot{\mathcal{F}}_{\mathcal{T}_k} $ of the segmentation network $T$ with RoI pooling and update them with CMA algorithm.
	Then, BARS measures the similarity of two cases, ``$\dot{\mathcal{F}}_{\mathcal{S}\to{\mathcal{T}_k}}\leftrightarrow \dot{\mathcal{C}}_{\mathcal{T}_k}^c$" and ``$\dot{\mathcal{F}}_{\mathcal{T}_k}\leftrightarrow \dot{\mathcal{C}}_{\mathcal{S}\to{\mathcal{T}_k}}^c$", and selects the adaptive region.
	\textcircled{m} is a switch that selects the labels for centroid update in \equref{eq:centroid}, either $Y_\mathcal{S}$, $\hat{Y}_{\mathcal{T}_k}$ for the first $m$ iterations or $Y^{BARS}_{\mathcal{S}\to{\mathcal{T}_k}}$, $\hat{Y}^{BARS}_{\mathcal{T}_k}$ after the $m$ iterations. We set m as 300 iterations for our experiments.
	}
	%BARS는 segmentation network의 intermediate feature $\mathcal{F}_{S}^T$, $\mathcal{F}_{S\to{\mathcal{T}_k}}^T$를 이용해 region-wise pooling과 CMA를 이용하여 Centroid $\mathcal{C}$를 추출해낸다. 그 후 가장 intermediate feature와 centroid를 이용하여 학습에 도움이 될 수 있는 region들을 selection한다.}
	\label{fig:BARS}
	\vspace{-3mm}
\end{figure}

\begin{table*}[t]
\newcommand\w{1.0cm}
    \centering
    \normalsize{
    \begin{tabular}{m{1.6cm}|
    l@{\hspace{1mm}}
    |
    >{\centering\arraybackslash}m{1cm}@{\hspace{1mm}}
    |
    >{\centering\arraybackslash}m{\w}@{\hspace{1mm}}
    >{\centering\arraybackslash}m{\w}@{\hspace{1mm}}
    >{\centering\arraybackslash}m{\w}@{\hspace{1mm}}
    >{\centering\arraybackslash}m{\w}@{\hspace{1mm}}
    >{\centering\arraybackslash}m{\w}@{\hspace{1mm}}
    >{\centering\arraybackslash}m{\w}@{\hspace{1mm}}
    >{\centering\arraybackslash}m{\w}@{\hspace{1mm}}|
    >{\centering\arraybackslash}m{\w}@{\hspace{1mm}}|
    >{\centering\arraybackslash}m{\w}@{\hspace{1mm}}}
    \toprule
    & Method & Target & flat & constr. & object & nature & sky & human & vehicle & mIoU & Avg.
    \\
    \hline
     \multirow{6}{*}{G $\to$C, I} & \multirow{2}{*}{ADVENT \cite{vu2019advent}} & C &
     93.9 & 80.2 & 26.2 & 79.0 & 80.5 & 52.5 & 78.0 & 70.0 & \multirow{2}{*}{67.4}
     \\
     & & I &
     \textbf{91.8} & 54.5 & 14.4 & 76.8 & 90.3 & 47.5 & 78.3 & 64.8 &
     \\
     \cline{2-12}
     & \multirow{2}{*}{MTKT\cite{saporta2021multi}} & C &
     94.5 & 82.0 & 23.7 & 80.1 & \textbf{84.0} & 51.0 & 77.6 & 70.4 & \multirow{2}{*}{68.2}
     \\
     & & I &
     91.4 & 56.6 & 13.2 & \textbf{77.3} & \textbf{91.4} & 51.4 & \textbf{79.9} & 65.9 &
     \\
     \cline{2-12}
     & \multirow{2}{*}{Ours} & C &
     \textbf{95.1} & \textbf{82.6} & \textbf{39.8} & \textbf{84.6} & 81.2 & \textbf{63.6} & \textbf{80.7} & \textbf{75.4} & \multirow{2}{*}{\textbf{71.2}}
     \\
     & & I &
     90.5 & \textbf{63.0} & \textbf{22.2} & 73.7 & 87.9 & \textbf{54.3} & 76.9 & \textbf{66.9} &
     \\
     
     \clineB{1-12}{2.5}
     \multirow{6}{*}{G $\to$C, M} & \multirow{2}{*}{ADVENT\cite{vu2019advent}} & C &
     93.1 & 80.5 & 24.0 & 77.9 & 81.0 & 52.5 & 75.0 & 69.1 & \multirow{2}{*}{68.9}
     \\
     & & M &
     90.0 & 71.3 & 31.1 & 73.0 & 92.6 & 46.6 & 76.6 & 68.7 &
     \\
     \cline{2-12}
     & \multirow{2}{*}{MTKT\cite{saporta2021multi}} & C &
     95.0 & 81.6 & 23.6 & 80.1 & 83.6 & 53.7 & 79.8 & 71.1 & \multirow{2}{*}{70.9}
     \\
     & & M &
     \textbf{90.6} & 73.3 & 31.0 & 75.3 & \textbf{94.5} & 52.2 & \textbf{79.8} & 70.8 &
     \\
     \cline{2-12}
     & \multirow{2}{*}{Ours} & C &
     \textbf{96.4} & \textbf{83.5} & \textbf{35.1} & \textbf{83.8} & \textbf{84.9} & \textbf{62.3} & \textbf{81.3} & \textbf{75.3} & \multirow{2}{*}{\textbf{73.9}}
     \\
     & & M &
     88.6 & \textbf{73.7} & \textbf{41.0} & \textbf{75.4} & 93.4 & \textbf{58.5} & 77.2 & \textbf{72.6} &
     \\
     
     \clineB{1-12}{2.5}
     \multirow{9}{*}{G $\to$C, I, M} & \multirow{3}{*}{ADVENT\cite{vu2019advent}} & C &
     93.6 & 80.6 & 26.4 & 78.1 & 81.5 & 51.9 & 76.4 & 69.8 & \multirow{3}{*}{67.8}
     \\
     & & I &
     \textbf{92.0} & 54.6 & 15.7 & 77.2 & 90.5 & 50.8 & 78.6 & 65.6 &\\
     & & M &
     89.2 & 72.4 & 32.4 & 73.0 & 92.7 & 41.6 & 74.9 & 68.0 &
     \\
     \cline{2-12}
     & \multirow{3}{*}{MTKT\cite{saporta2021multi}} & C &
     94.6 & 80.7 & 23.8 & 79.0 & 84.5 & 51.0 & 79.2 & 70.4 & \multirow{3}{*}{69.1}
     \\
     & & I &
     91.7 & \textbf{55.6} & 14.5 & 78.0 & \textbf{92.6} & 49.8 & \textbf{79.4} & 65.9 &
     \\
     & & M &
     \textbf{90.5} & \textbf{73.7} & 32.5 & 75.5 & \textbf{94.3} & 51.2 & \textbf{80.2} & 71.1 &
     \\
     \cline{2-12}
     & \multirow{3}{*}{Ours} & C &
     \textbf{95.8} & \textbf{82.4} & \textbf{38.3} & \textbf{82.4} & \textbf{85.0} & \textbf{60.5} & \textbf{80.2} & \textbf{74.9} & \multirow{3}{*}{\textbf{71.3}}
     \\
     & & I &
     89.9 & 52.7 & \textbf{25.0} & \textbf{78.1} & 92.1 & \textbf{51.0} & 77.9 & \textbf{66.7} &
     \\
     & & M &
     89.2 & 71.5 & \textbf{45.2} & \textbf{75.8} & 92.3 & \textbf{56.1} & 75.4 & \textbf{72.2} &
     \\
     \bottomrule
     
    \end{tabular}
    }
    \caption{Quantitative comparison between our method and state-of-the-art methods on GTA5 (G) to Cityscapes (C), IDD (I), and Mapilary (M) with 7 classes setting. \textbf{Bold}: Best score among all the methods.}
\label{tab:7classes}
\end{table*}

\subsection{Bi-directional Adaptive Region Selection (BARS)}
\label{sec:BARS}

The key idea of BARS is to select the pixels where the feature statistics are consistent, then train a task network $T$ by imposing loss on the selected region as illustrated in~\figref{fig:BARS}.
We apply it in both the domain transferred image and the target image.
We first extract each centroid feature $\dot{\mathcal{C}}$ of class $c$ as follows:
\begin{gather}
\begin{split}
\dot{\mathcal{C}}_{\mathcal{S}\to \mathcal{T}_k}^{c}
= \frac{1}{N_c}\sum_{i} \sum_{j} 
{\mathbbm{1}(Y_\mathcal{S}(i, j)=c)} \dot{\mathcal{F}}_{\mathcal{S}\to \mathcal{T}_k}(i, j),\\
\dot{\mathcal{C}}_{\mathcal{T}_k}^{c} = \frac{1}{N_c}\sum_{i} \sum_{j} 
{\mathbbm{1}(\hat{Y}_{\mathcal{T}_k}(i, j)=c)} \dot{\mathcal{F}}_{\mathcal{T}_k}(i, j),
\end{split}
\label{eq:centroid}
\end{gather}
where $\mathbbm{1}$ is an indicator function, $N_c$ is the number of pixels of class $c$, and $i, j$ are the indices of the spatial coordinates.
The feature map $\dot{\mathcal{F}}$ is from the second last layer of the task network $T$.
To extract the centroids, we use the ground-truth label $Y_\mathcal{S}$ of the domain transferred image and the pseudo label $\hat{Y}_{\mathcal{T}_k}$ of the target image.
For the online learning with the centroids, we also apply the CMA algorithm in \algref{alg:CMA} to the above centroids.
Then, we design the selection mechanism using the following two assumptions:
\begin{itemize}
% \item The region with the feature far from the target centroid feature $\mathcal{C}^{\mathcal{T}_k}$ in the $\mathcal{F}_{\mathcal{S}\to \mathcal{T}_k}^{T}$ would not be \textcolor{red}{helpful} for the adaptation.
\item The region with features $\dot{\mathcal{F}}_{\mathcal{S}\to \mathcal{T}_k}$ far from the target centroid $\dot{\mathcal{C}}_{\mathcal{T}_k}$ would disturb the adaptation process.
\item The region with target features $\dot{\mathcal{F}}_{\mathcal{T}_k}$ far from the centroids $\dot{\mathcal{C}}_{S\to \mathcal{T}_k}$ is likely to be a noisy prediction region.
\end{itemize}
% Based on these assumptions, we compute the distance $d$ between $\dot{\mathcal{F}}_{\mathcal{S}\to \mathcal{T}_k}$ and $\dot{\mathcal{C}}_{\mathcal{T}_k}$, and between $\dot{\mathcal{F}}_{\mathcal{T}_k}$ and $\dot{\mathcal{C}}_{\mathcal{S}\to \mathcal{T}_k}$:
% \begin{gather}
% \begin{split}
% d^{c}_{\mathcal{S}\to \mathcal{T}_k} (i, j) = || \dot{\mathcal{F}}_{\mathcal{S}\to \mathcal{T}_k}(i, j) - \dot{\mathcal{C}}^{c}_{\mathcal{T}_k} ||_2,\\
% d^{c}_{\mathcal{T}_k} (i, j) = || \dot{\mathcal{F}}_{\mathcal{T}_k}(i, j) - \dot{\mathcal{C}}_{\mathcal{S}\to \mathcal{T}_k}^c ||_2.
% \end{split}
% \end{gather}
Based on these assumptions, we find the nearest class $\dot{c}$ for each pixel in the feature map using the L2 distance between features on each pixels and centroid features as follows:
\begin{gather}
\begin{split}
\dot{c}_{\mathcal{S}\to \mathcal{T}_k}(i, j) = \argmin_c || \dot{\mathcal{F}}_{\mathcal{S}\to \mathcal{T}_k}(i, j) - \dot{\mathcal{C}}^{c}_{\mathcal{T}_k} ||_2, \\
\dot{c}_{\mathcal{T}_k}(i, j) = \argmin_c || \dot{\mathcal{F}}_{\mathcal{T}_k}(i, j) - \dot{\mathcal{C}}_{\mathcal{S}\to \mathcal{T}_k}^c ||_2.
\end{split}
\end{gather}
We obtain the filtered labels $Y_{\mathcal{S}\to{\mathcal{T}_k}}^{BARS}$, $\hat{Y}_{\mathcal{T}_k}^{BARS}$ using the nearest class $\dot{c}$:
\begin{gather}
\begin{split}
Y_{\mathcal{S}\to{\mathcal{T}_k}}^{BARS}(i, j) = 
\begin{cases}
Y_\mathcal{S}(i, j) & \quad \text{if}~\dot{c}_{\mathcal{S}\to \mathcal{T}_k}(i,j)=Y_\mathcal{S}(i, j),\\
\emptyset & \quad \text{otherwise}\\
\end{cases},
\\
\hat{Y}_{\mathcal{T}_k}^{BARS}(i, j) = 
\begin{cases}
\hat{Y}_{\mathcal{T}_k}(i, j) & \quad \text{if}~\dot{c}_{\mathcal{T}_k}(i,j)=\hat{Y}_{\mathcal{T}_k}(i, j),\\
\emptyset & \quad \text{otherwise}\\
\end{cases}.
\end{split}
\end{gather}
Finally, we train the task network $T$ with the labels using a typical cross-entropy loss $\mathcal{L}_{Task}$:
\setlength\abovedisplayskip{10pt plus 2pt minus 10pt}
\begin{gather}
\begin{split}
\min_{T} \left( \mathcal{L}_{Task} (I_{\mathcal{S} \to \mathcal{T}_k}, 
Y_{\mathcal{S}\to{\mathcal{T}_k}}^{BARS}) + \mathcal{L}_{Task} (I_{\mathcal{T}_k}, \hat{Y}_{\mathcal{T}_k}^{BARS}) \right).
% \\
% Y_{\mathcal{S}\to{\mathcal{T}_k}}^{BARS}(i, j) = 
% \begin{cases}
% Y_\mathcal{S}(i, j) & \quad \text{if}~c'_{\mathcal{S}\to \mathcal{T}_k}=Y_\mathcal{S}(i, j),\\
% \emptyset & \quad \text{otherwise}\\
% \end{cases},
% \\
% \hat{Y}_{\mathcal{T}_k}^{BARS}(i, j) = 
% \begin{cases}
% \hat{Y}_{\mathcal{T}_k}(i, j) & \quad \text{if}~c'_{\mathcal{T}_k}=\hat{Y}_{\mathcal{T}_k}(i, j),\\
% \emptyset & \quad \text{otherwise}\\
% \end{cases},
\end{split}
\end{gather}

% Experiments

\begin{figure*}
    \newcommand\w{2.3cm}
    \centering
    \begin{tabular}{
    >{\centering\arraybackslash}m{0.2cm}@{\hspace{3mm}}
    >{\centering\arraybackslash}m{\w}@{\hspace{1mm}}
    >{\centering\arraybackslash}m{\w}@{\hspace{1mm}}
    >{\centering\arraybackslash}m{\w}@{\hspace{1mm}}
    >{\centering\arraybackslash}m{\w}@{\hspace{1mm}}
    >{\centering\arraybackslash}m{\w}@{\hspace{1mm}}
    >{\centering\arraybackslash}m{\w}@{\hspace{1mm}}
    >{\centering\arraybackslash}m{\w}@{\hspace{1mm}}
    }
    
    % & Image & \multicolumn{2}{c}{GT} & \multicolumn{2}{c}{Source only} & \multicolumn{2}{c}{Ours}  
    & & \multicolumn{3}{c}{7 classes} & \multicolumn{3}{c}{19 classes}
    \\
    \cmidrule(lr){3-5} \cmidrule(lr){6-8}
    & Image & GT & Source only & Ours & GT & Source only & Ours  
    \\
    \multirow{4}{*}{C} 
    & \includegraphics[width=\w]{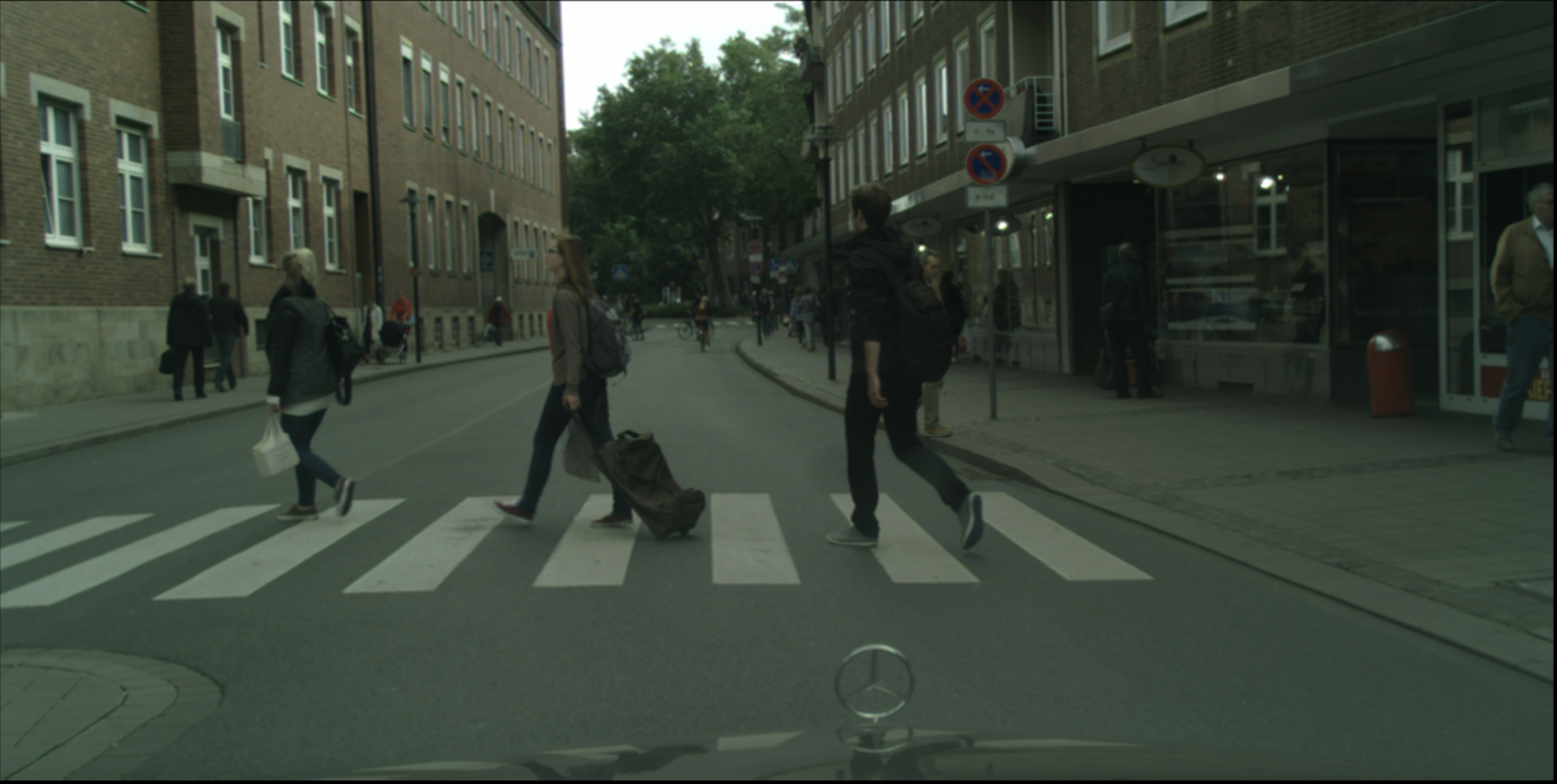}
    & \includegraphics[width=\w]{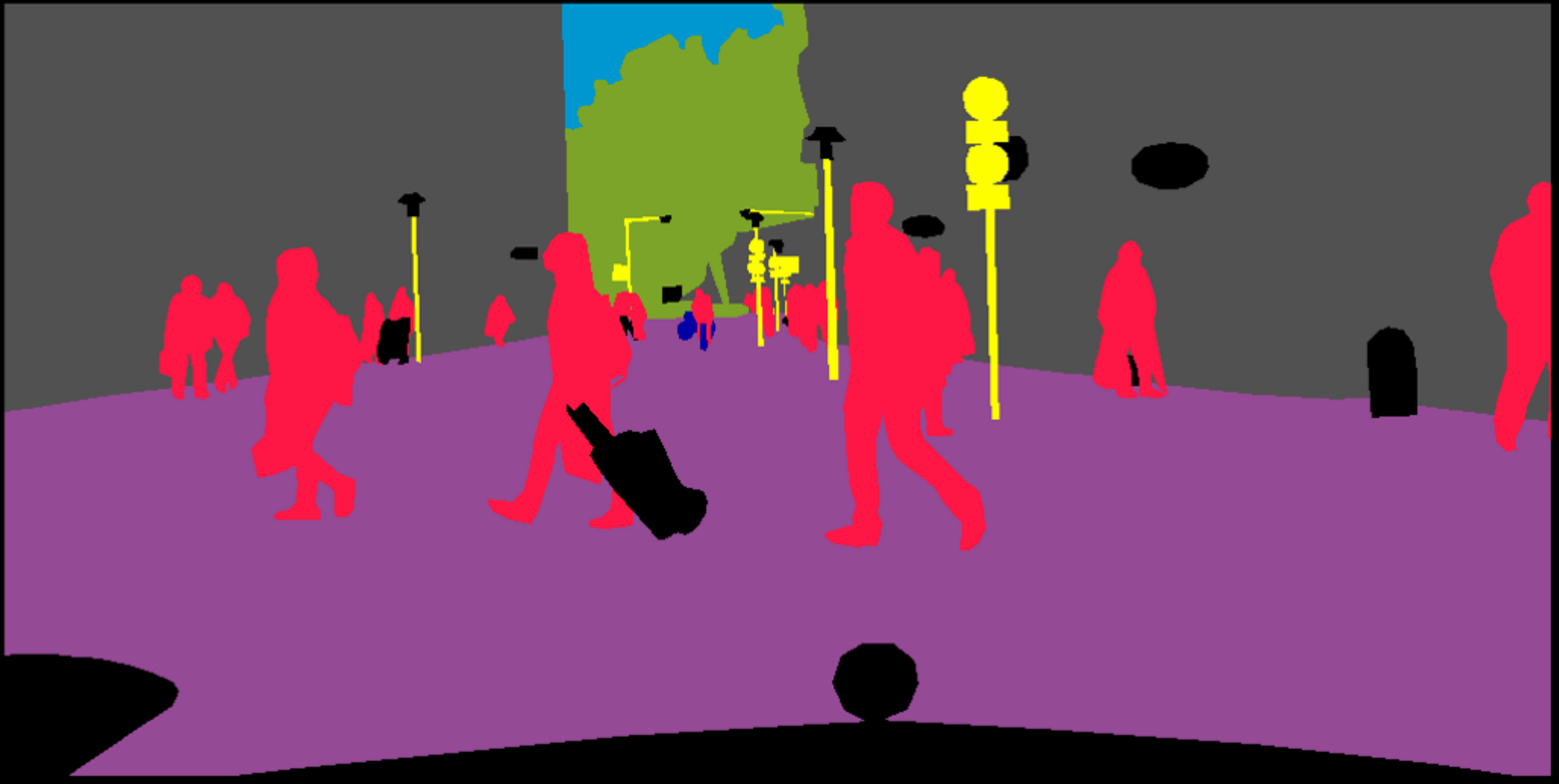}
    & \includegraphics[width=\w]{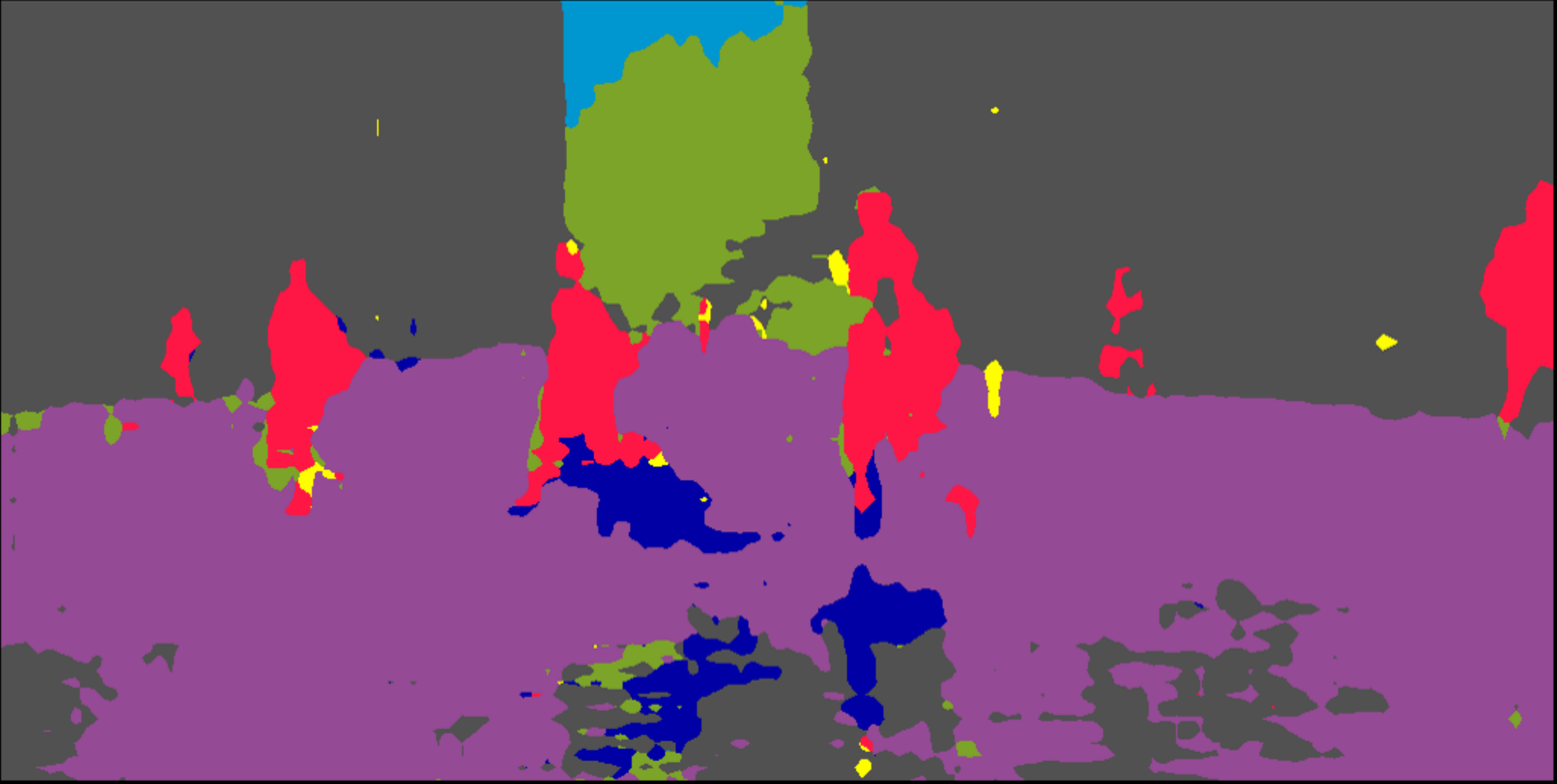}
    & \includegraphics[width=\w]{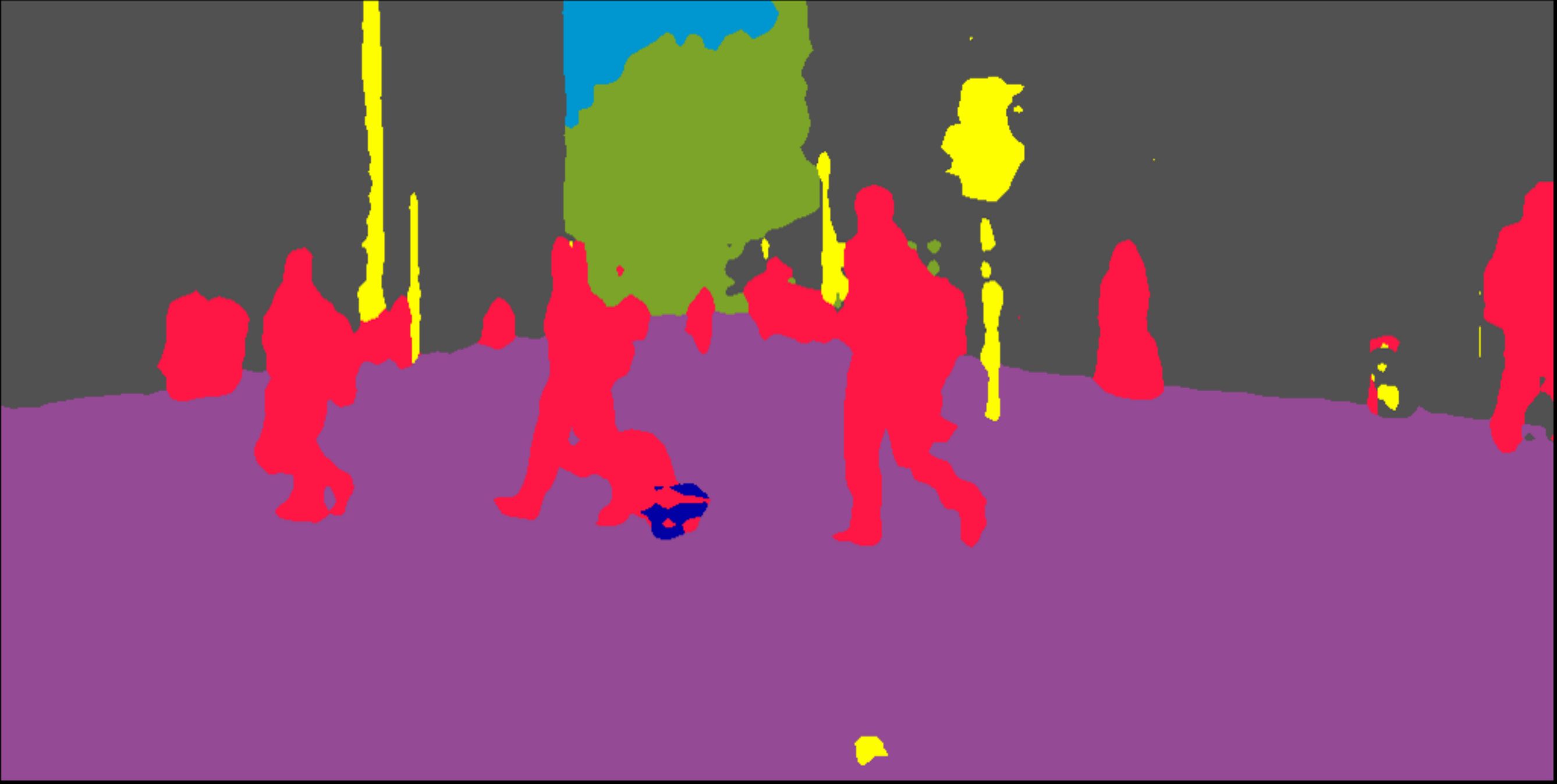}
    & \includegraphics[width=\w]{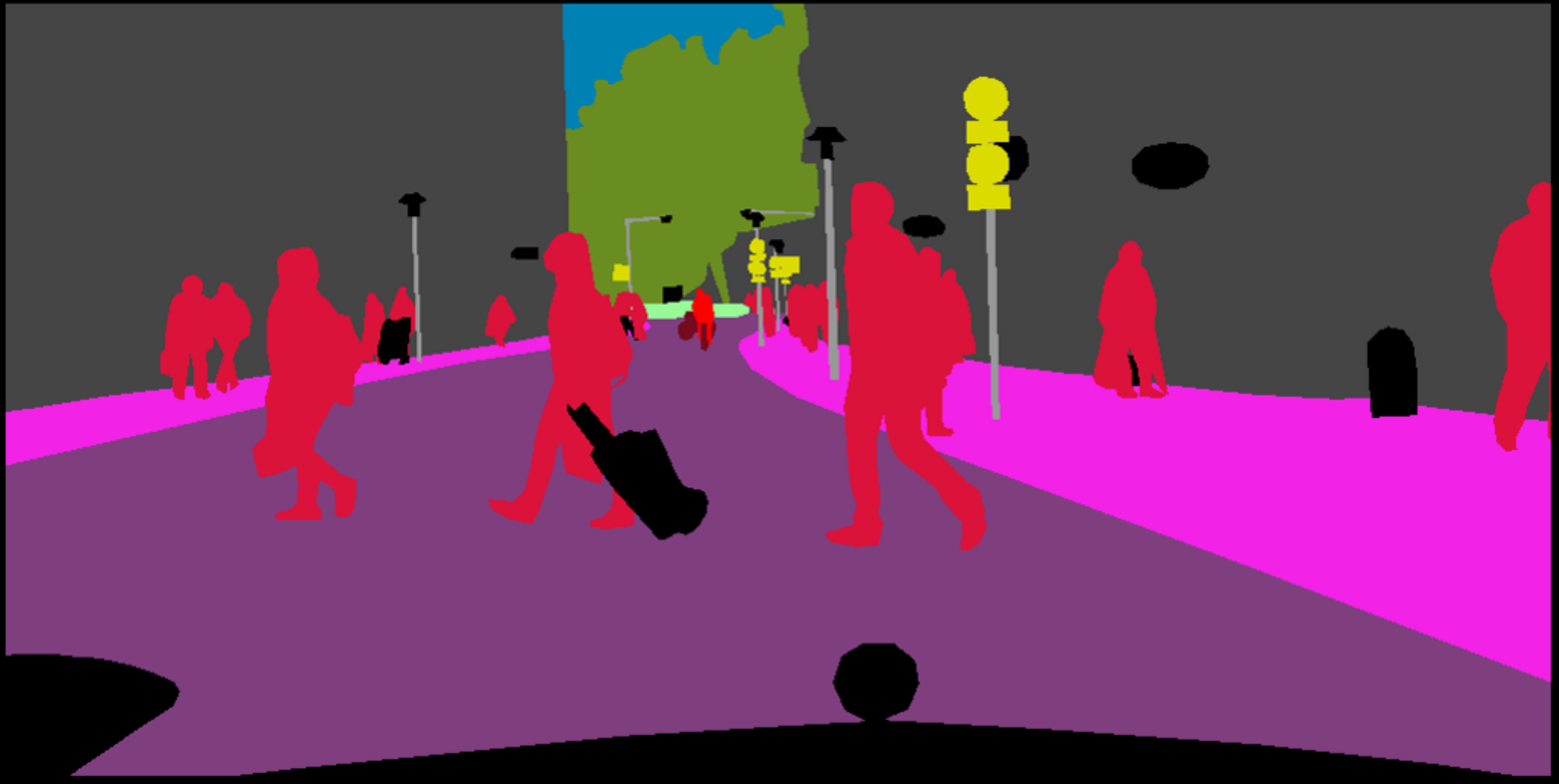}
    & \includegraphics[width=\w]{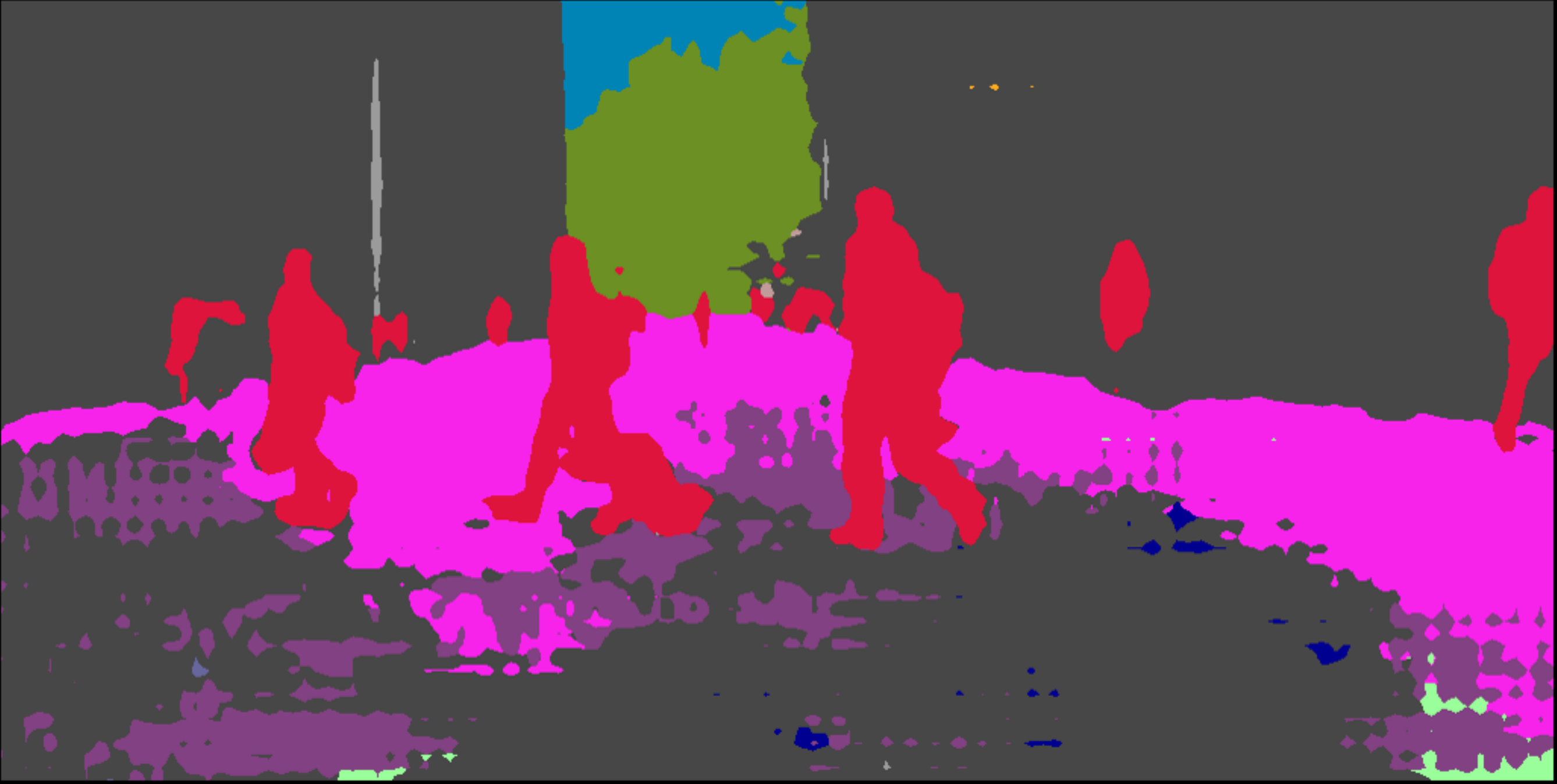}
    & \includegraphics[width=\w]{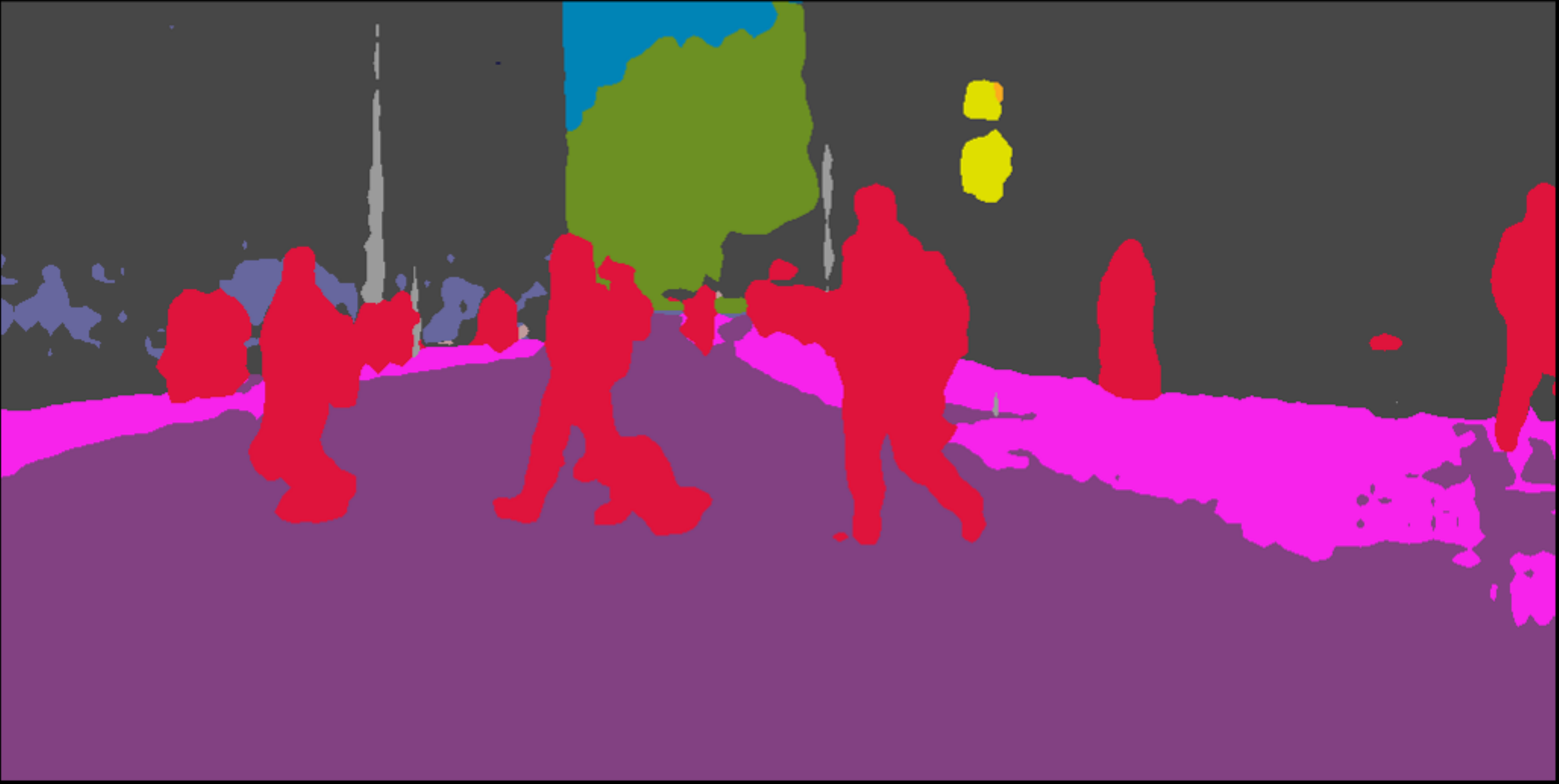}
    \\
    
    & \includegraphics[width=\w]{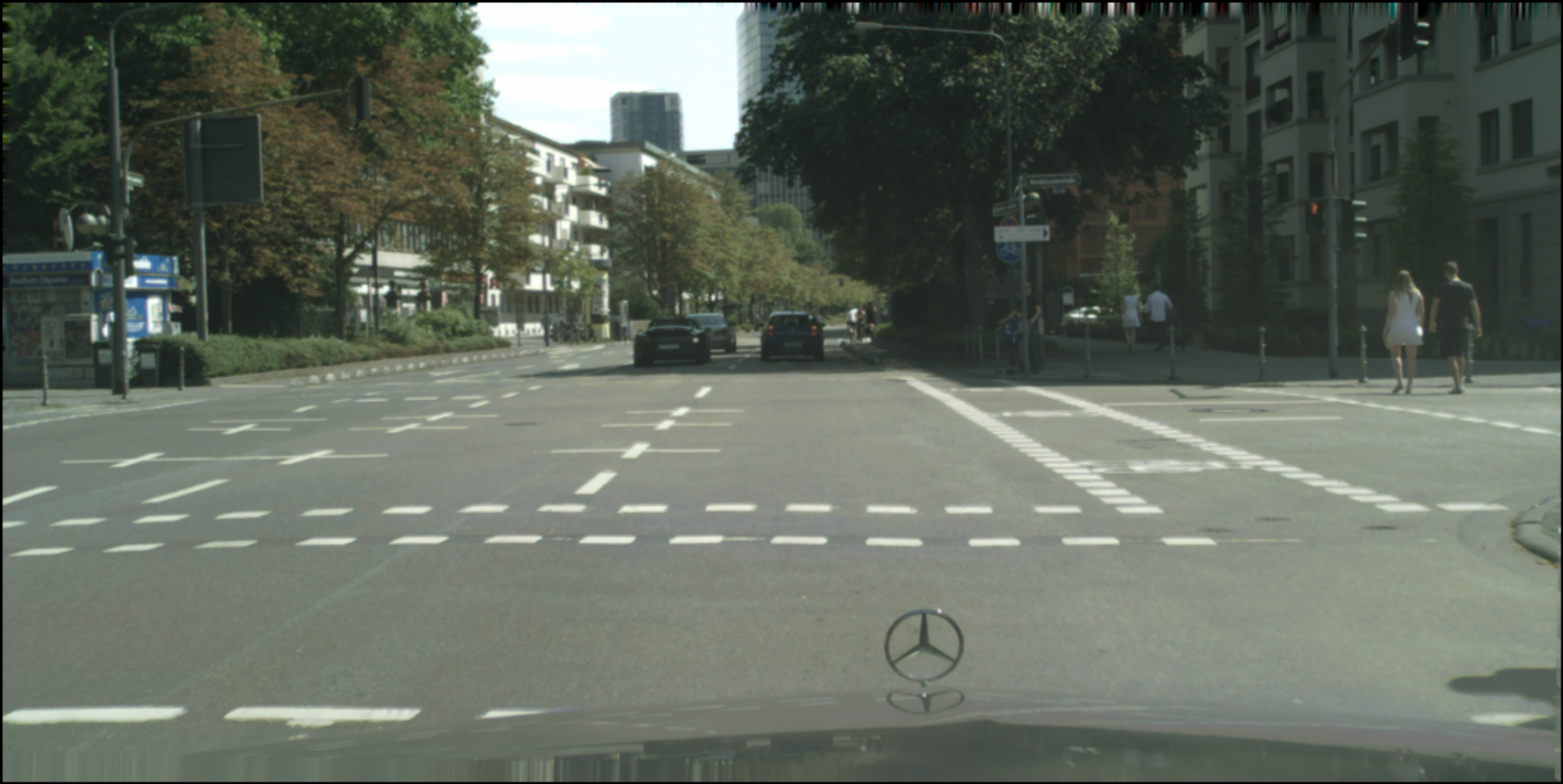}
    & \includegraphics[width=\w]{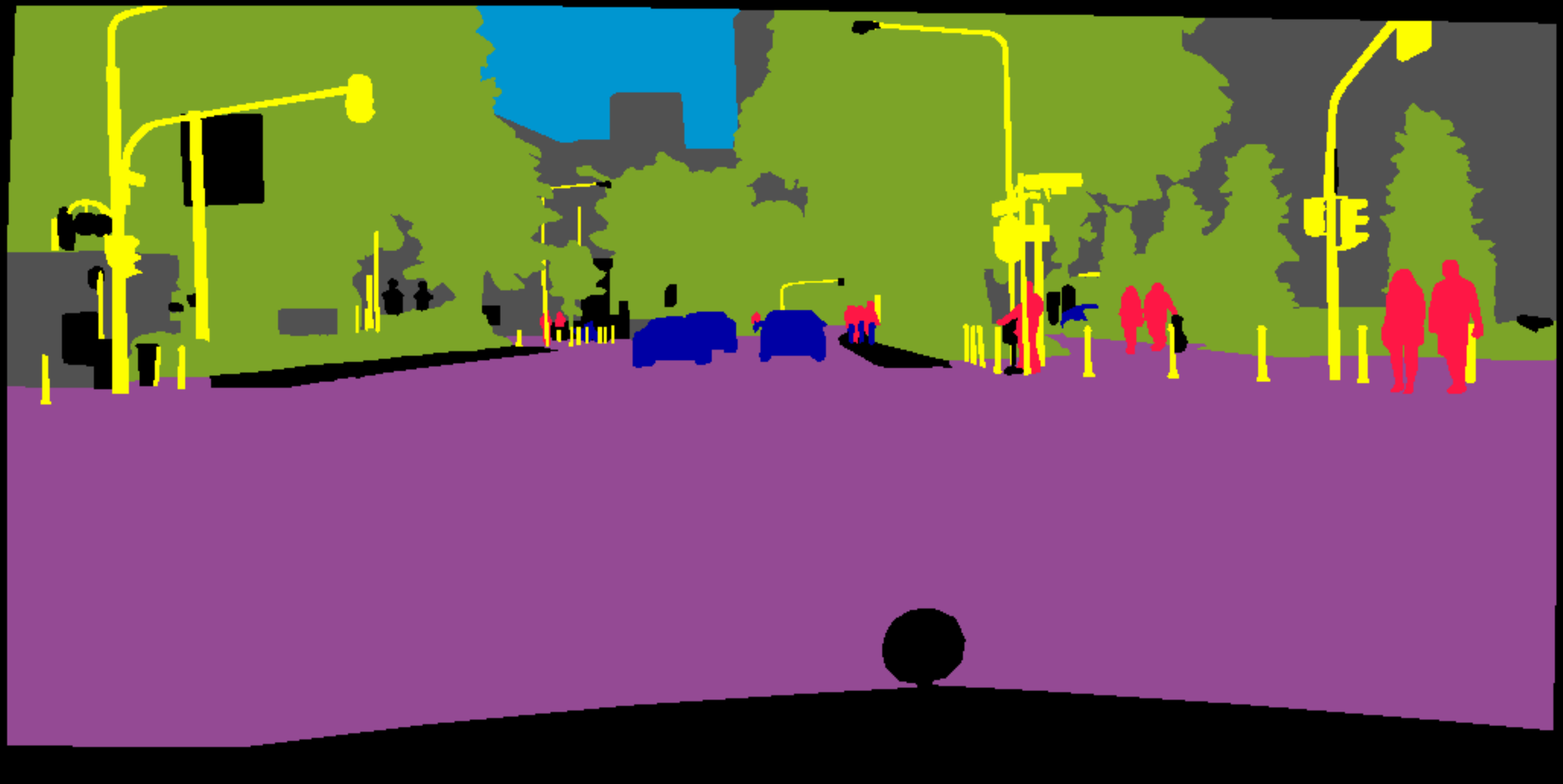}
    & \includegraphics[width=\w]{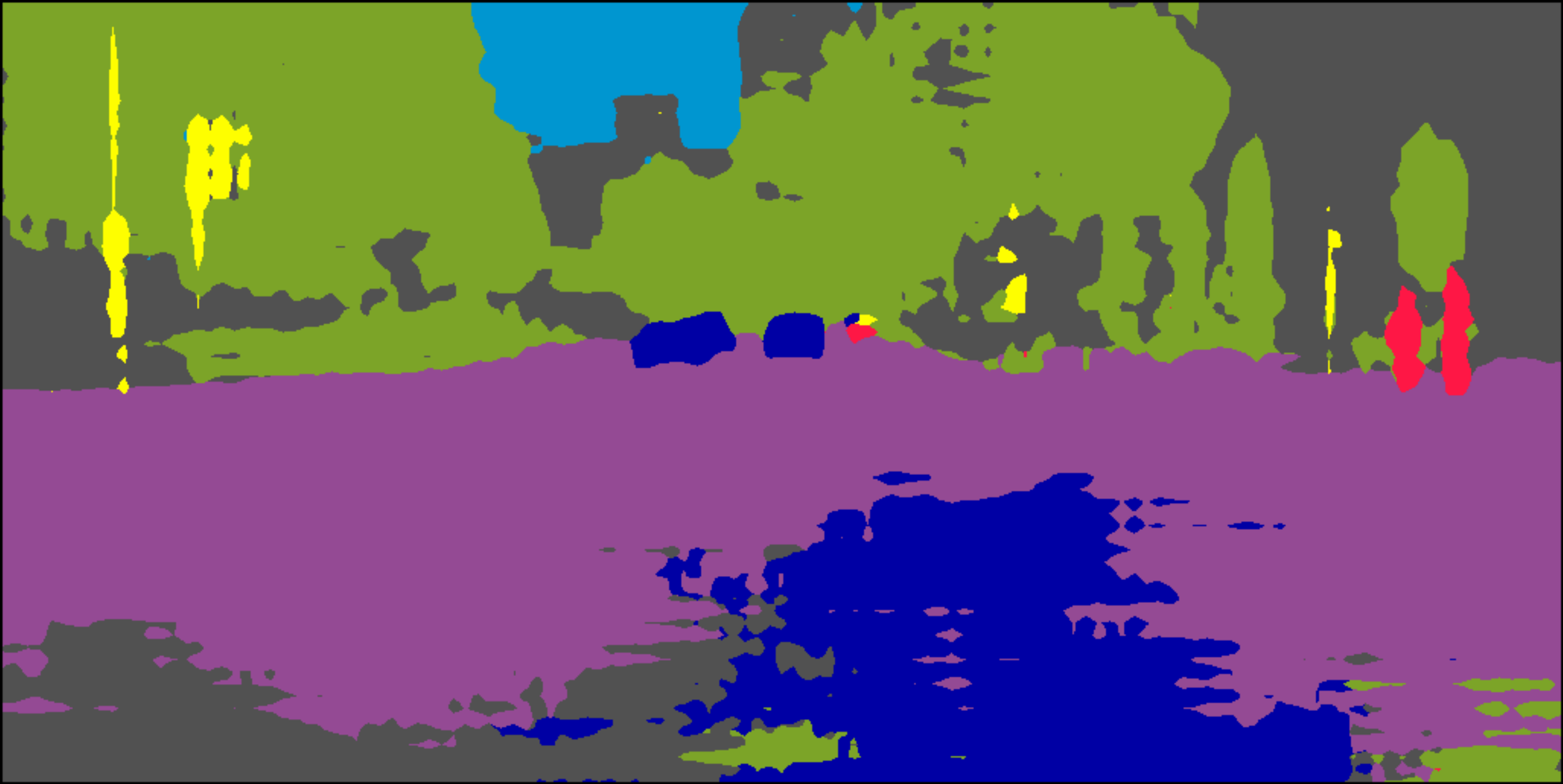}
    & \includegraphics[width=\w]{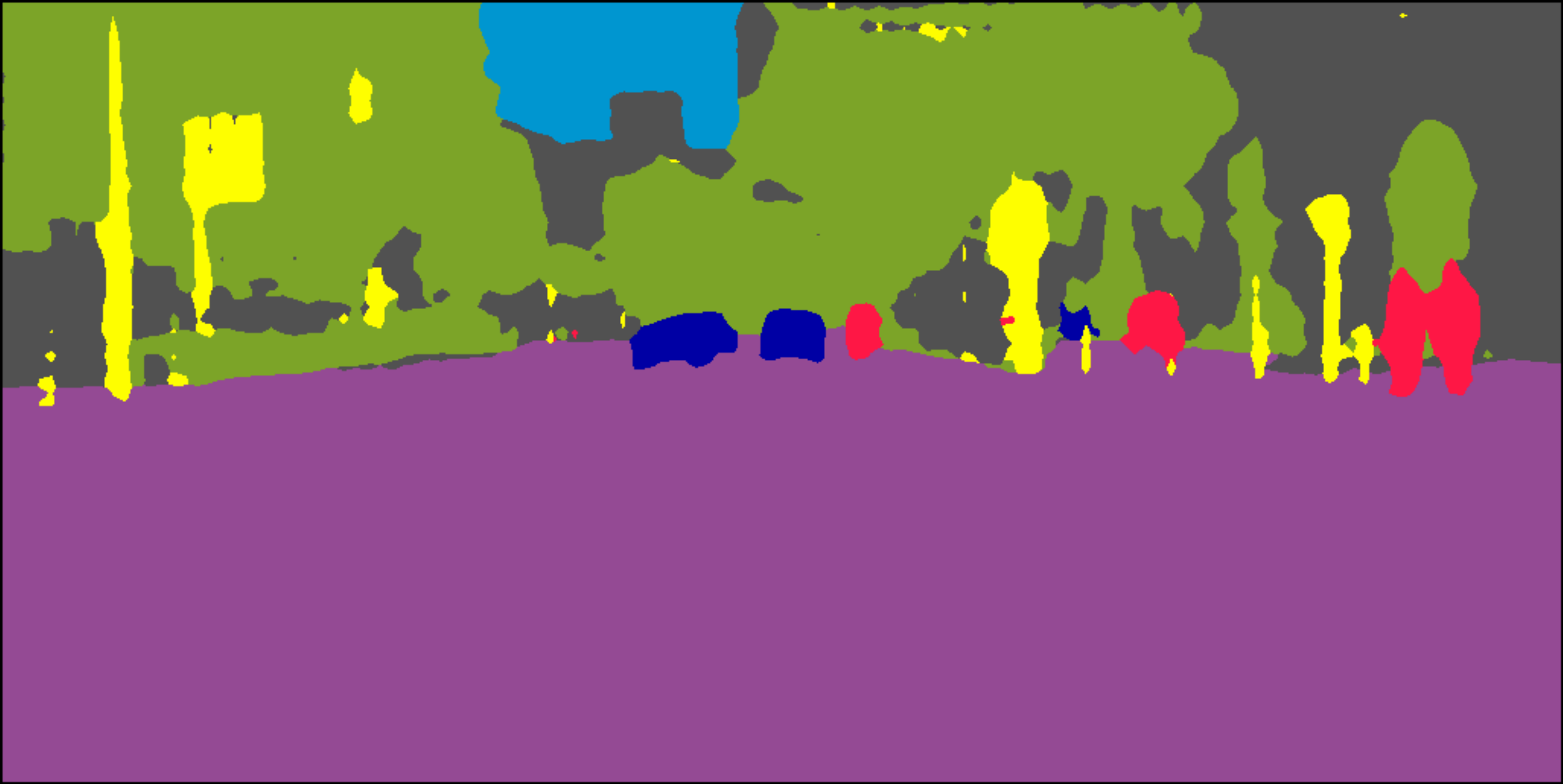}
    & \includegraphics[width=\w]{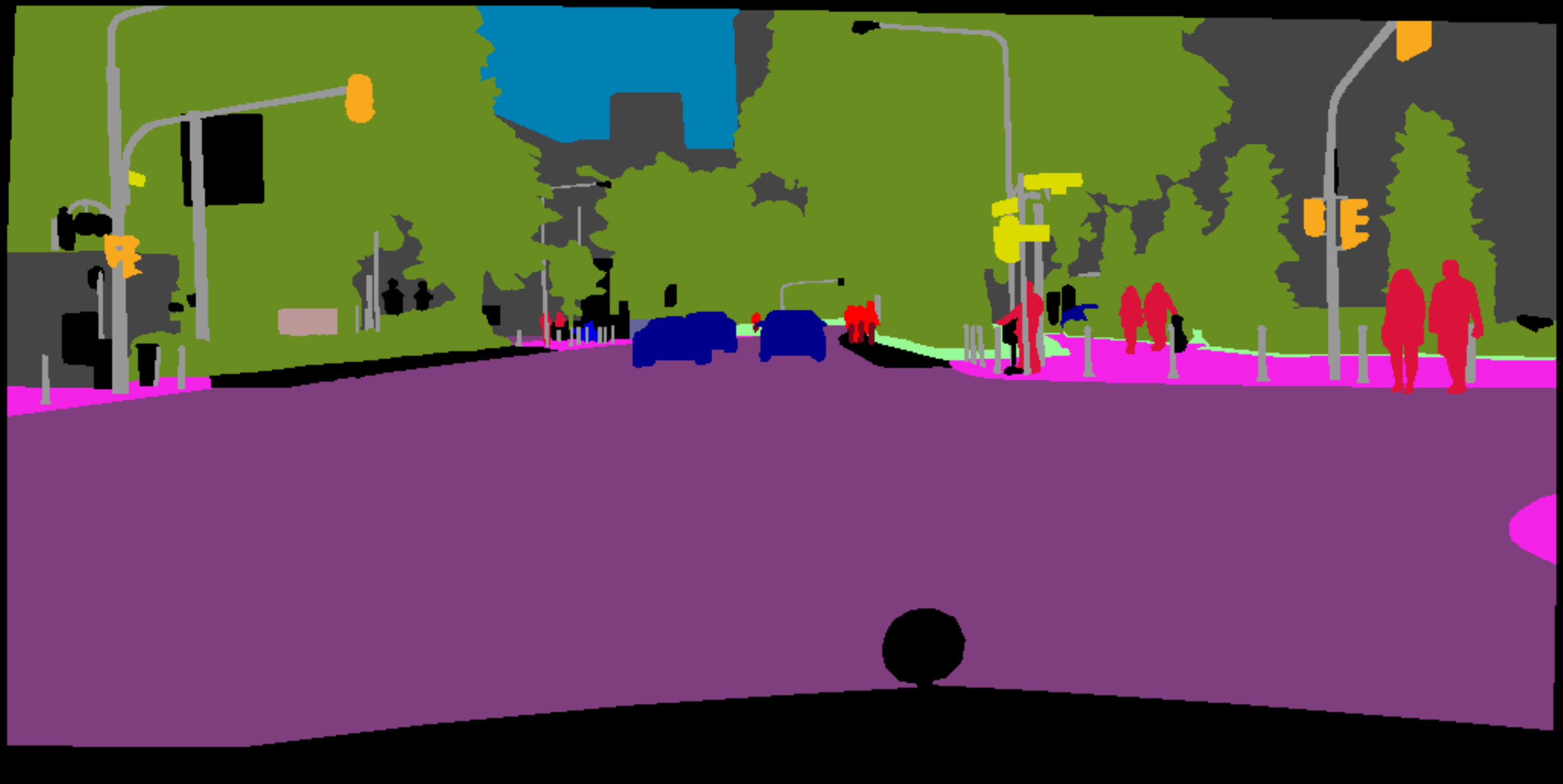}
    & \includegraphics[width=\w]{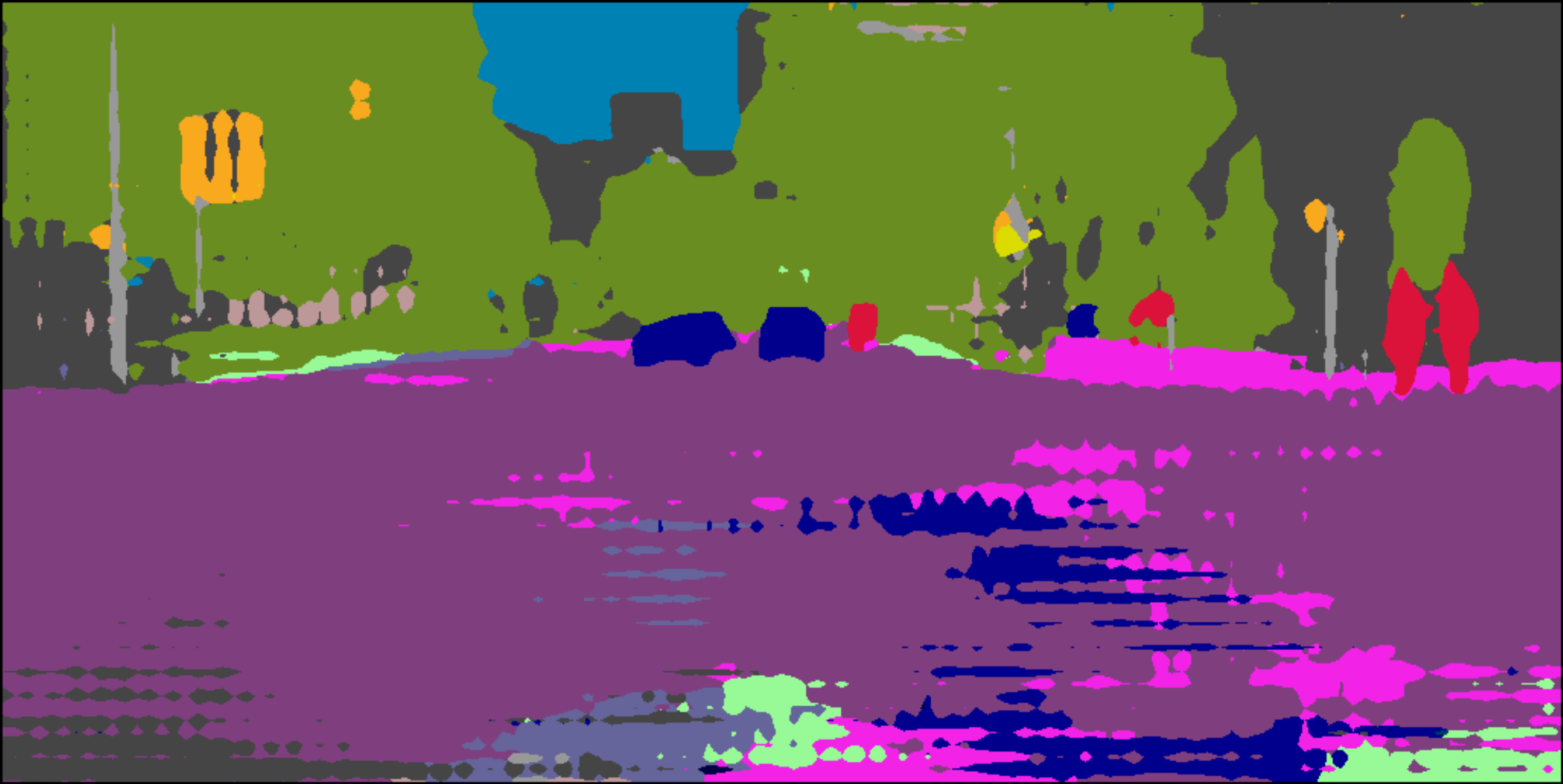}
    & \includegraphics[width=\w]{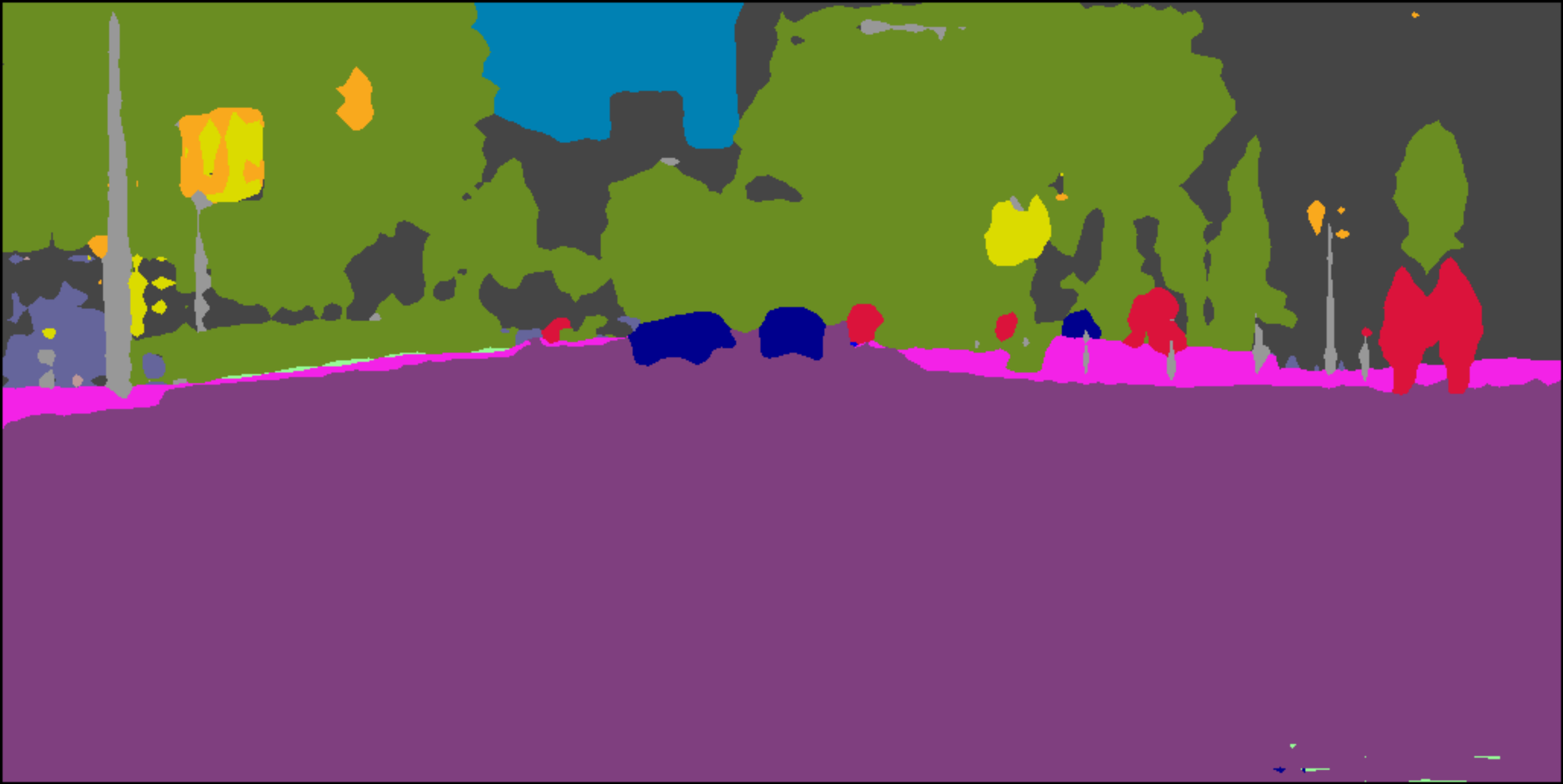}
    \\
    \multirow{4}{*}{I}
    & \includegraphics[width=\w]{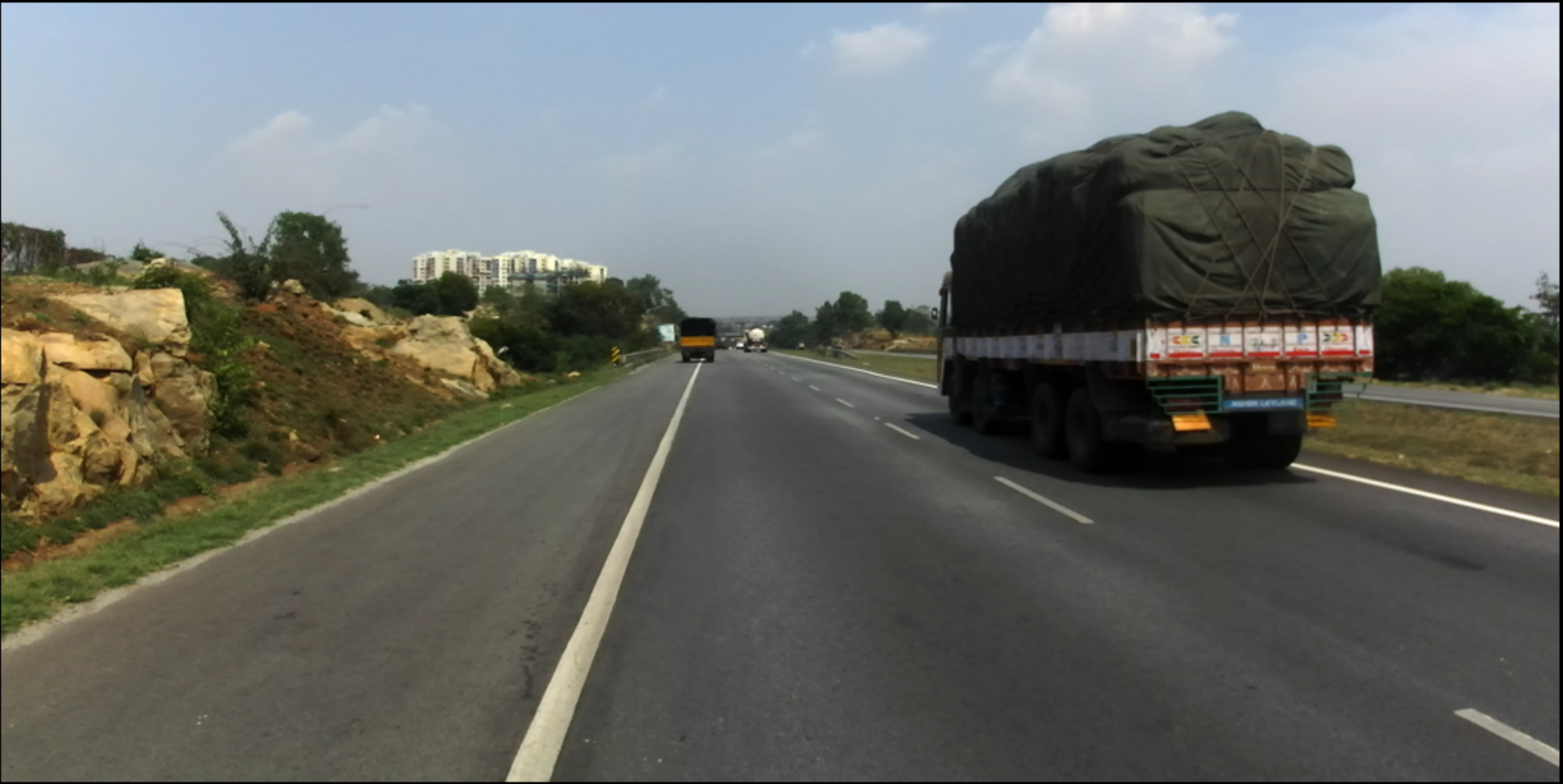}
    & \includegraphics[width=\w]{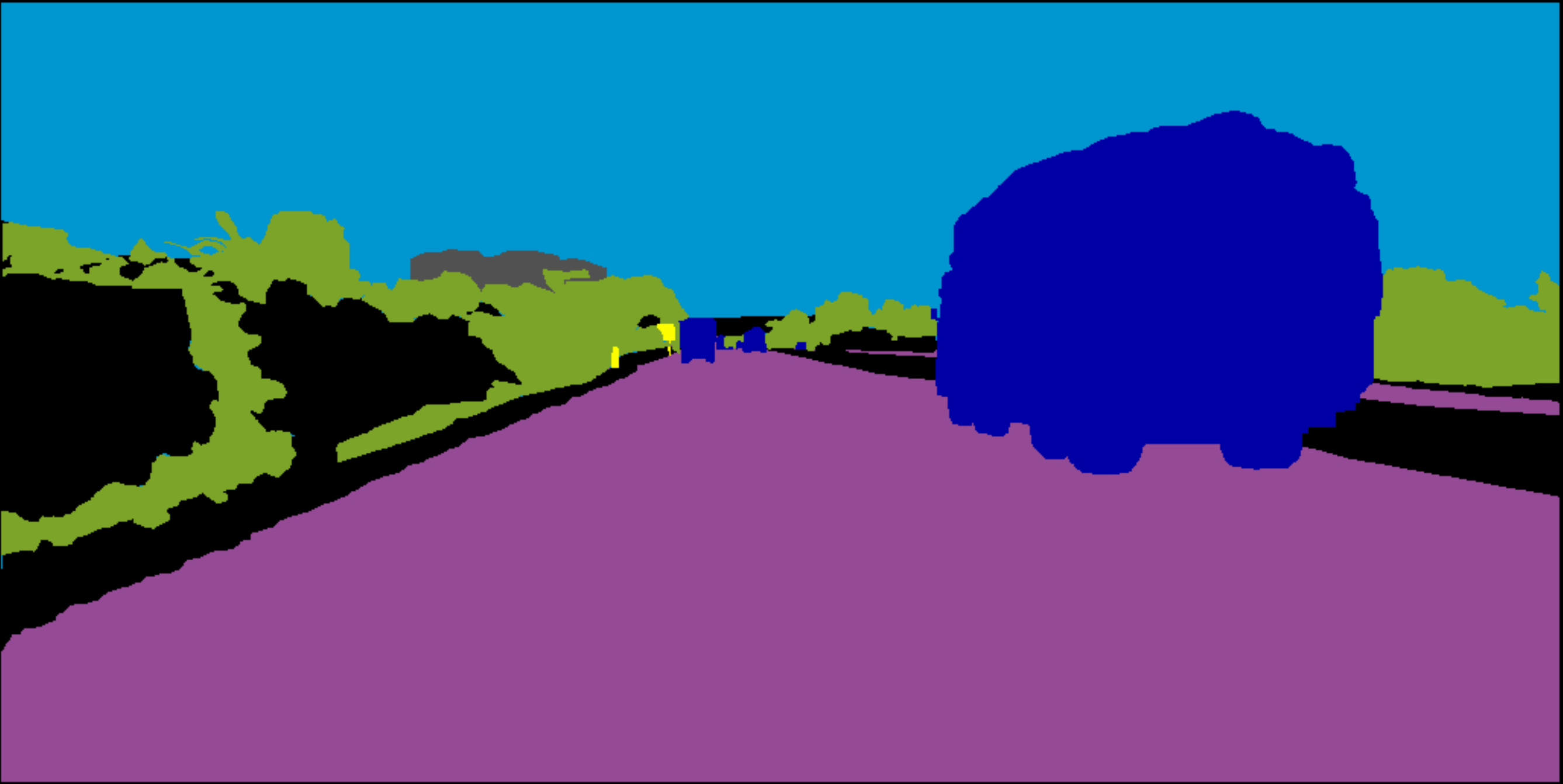}
    & \includegraphics[width=\w]{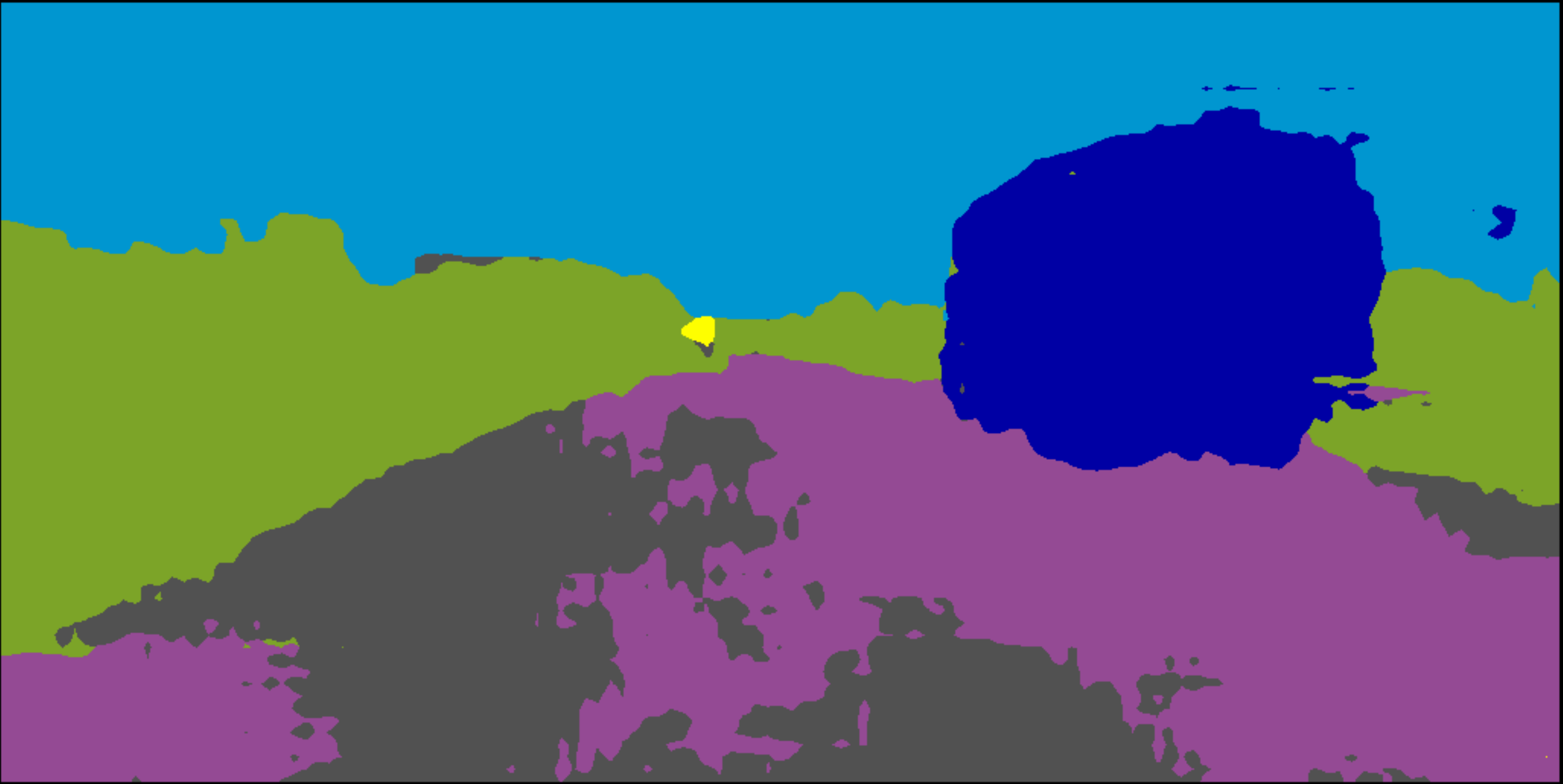}
    & \includegraphics[width=\w]{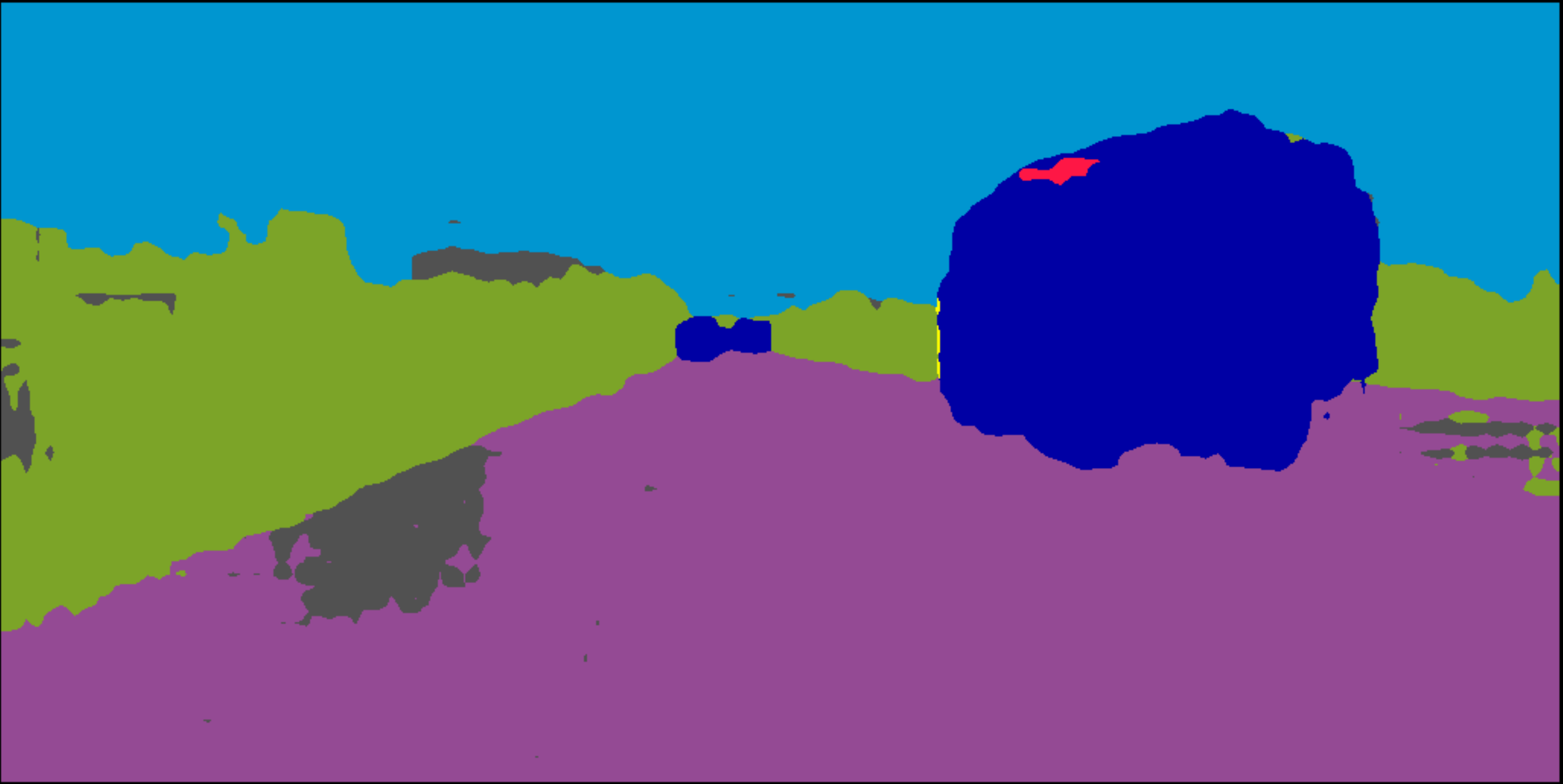}
    & \includegraphics[width=\w]{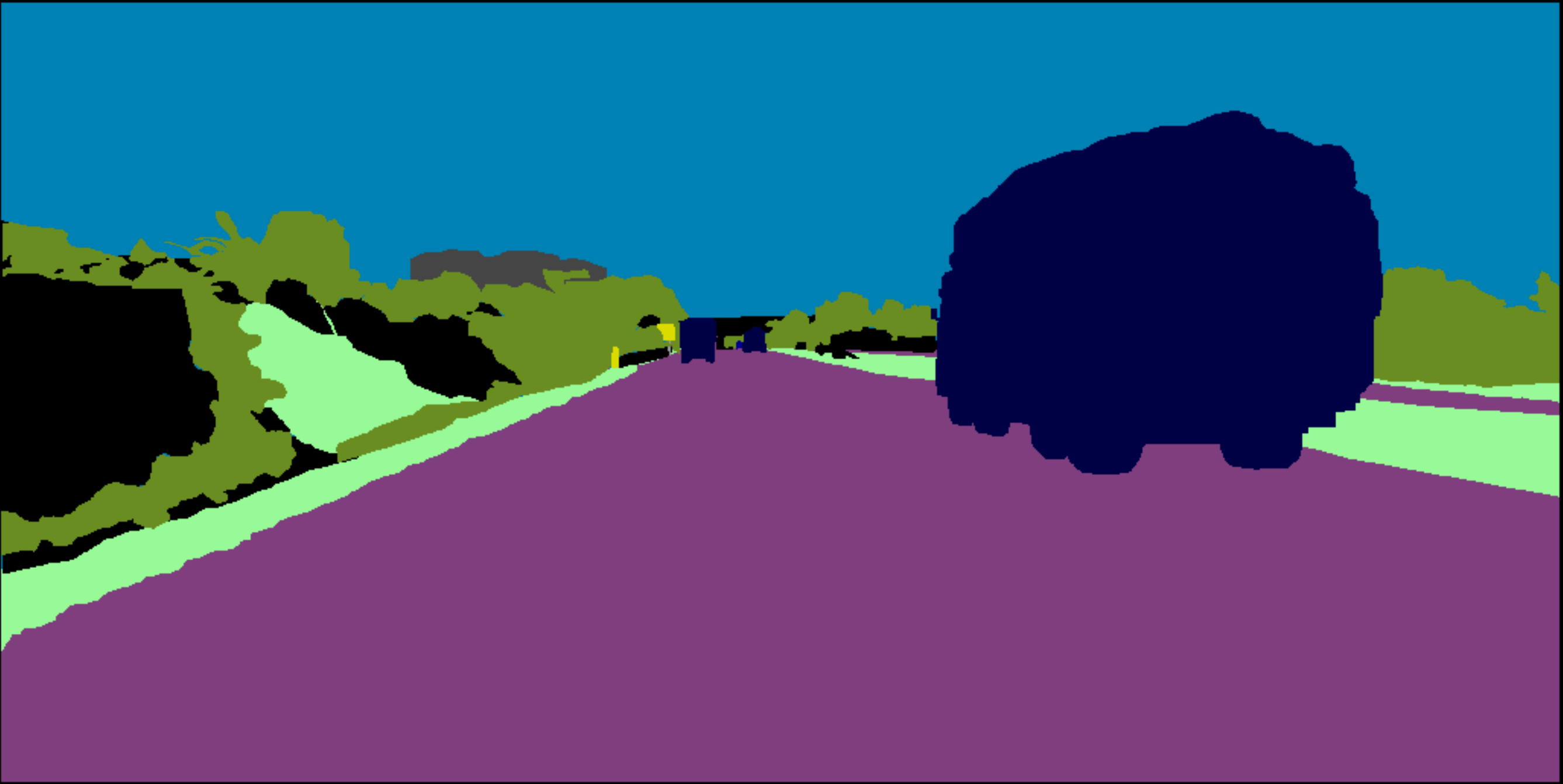}
    & \includegraphics[width=\w]{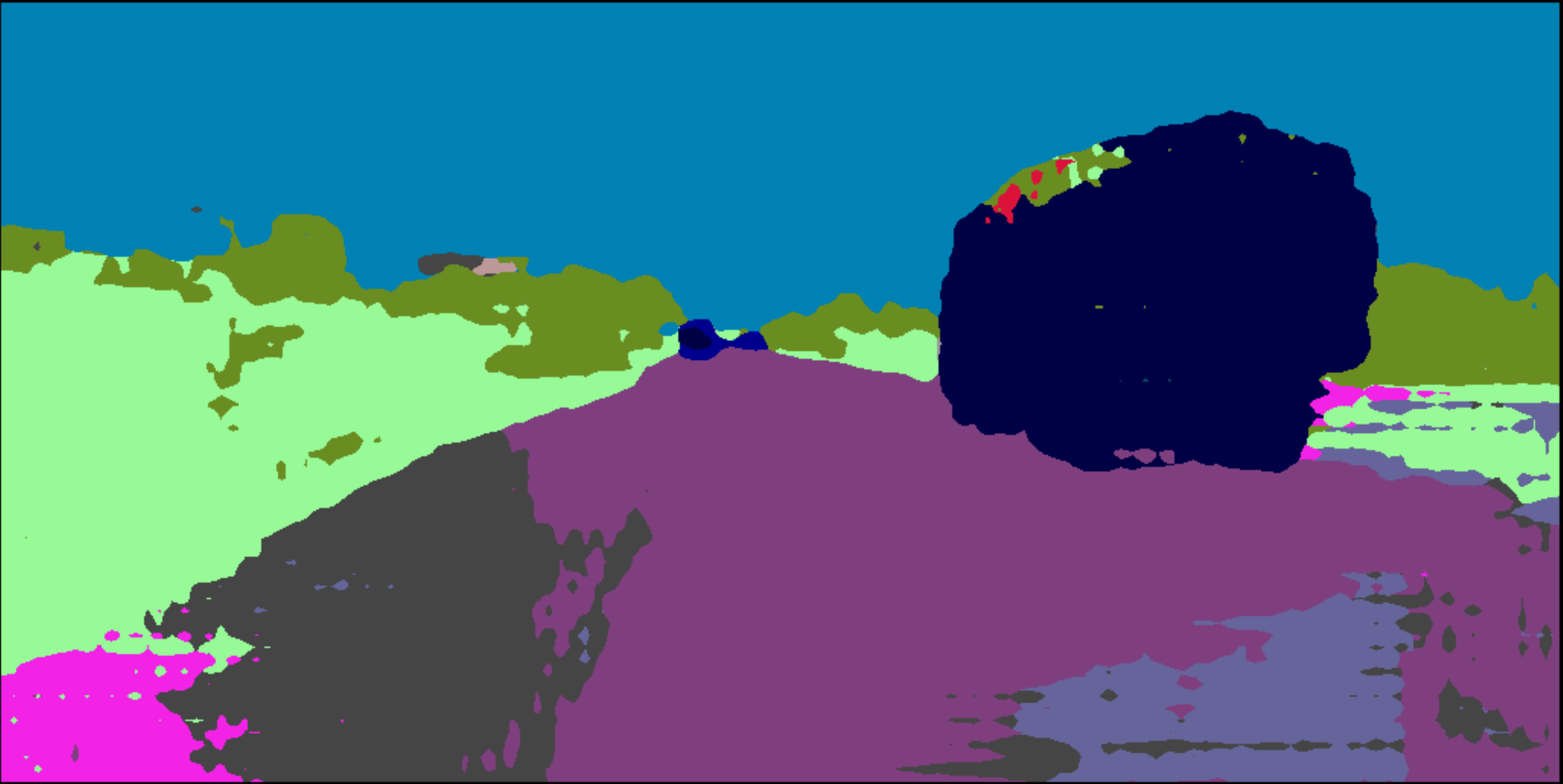}
    & \includegraphics[width=\w]{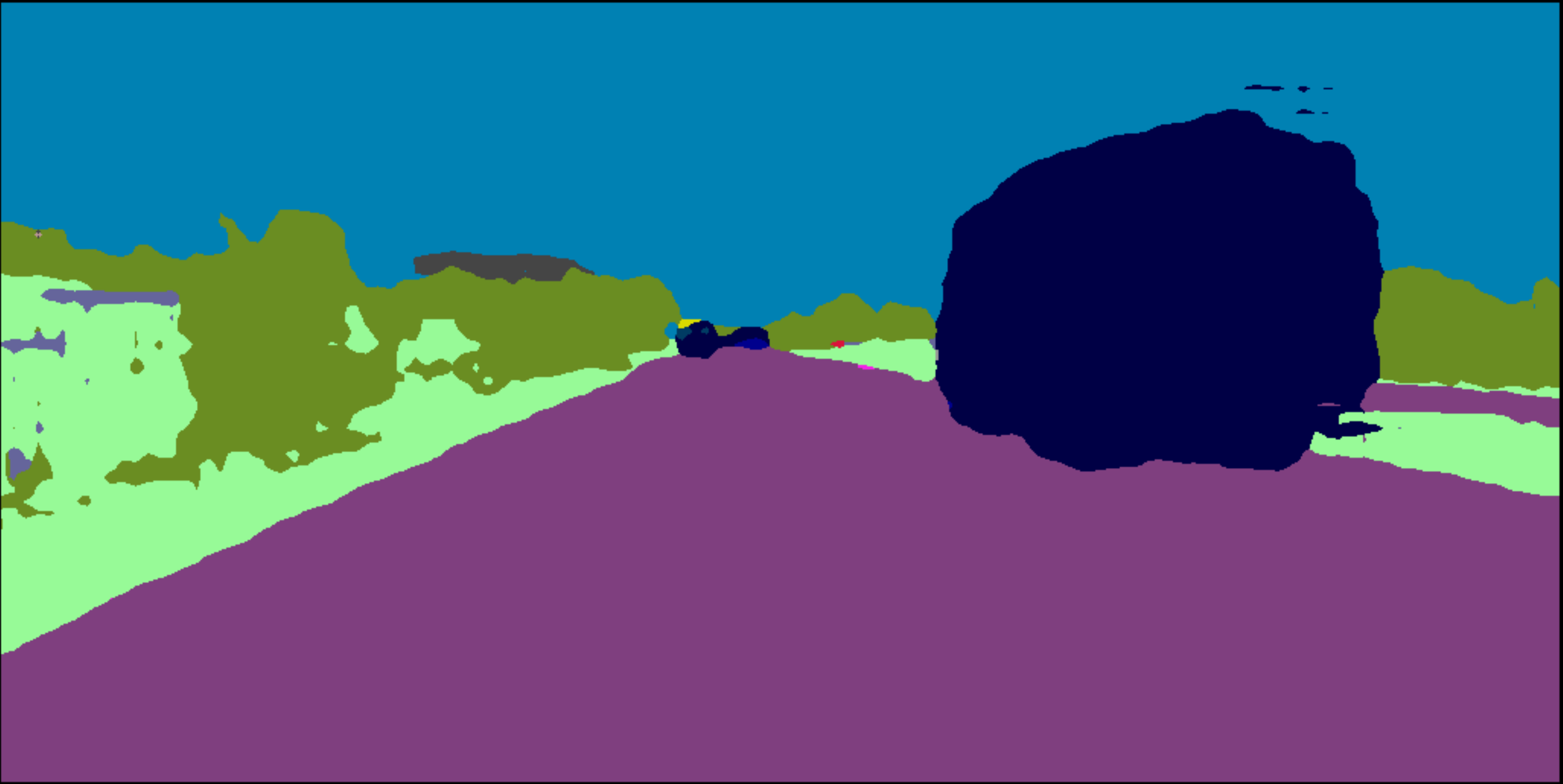}
    \\
    
    & \includegraphics[width=\w]{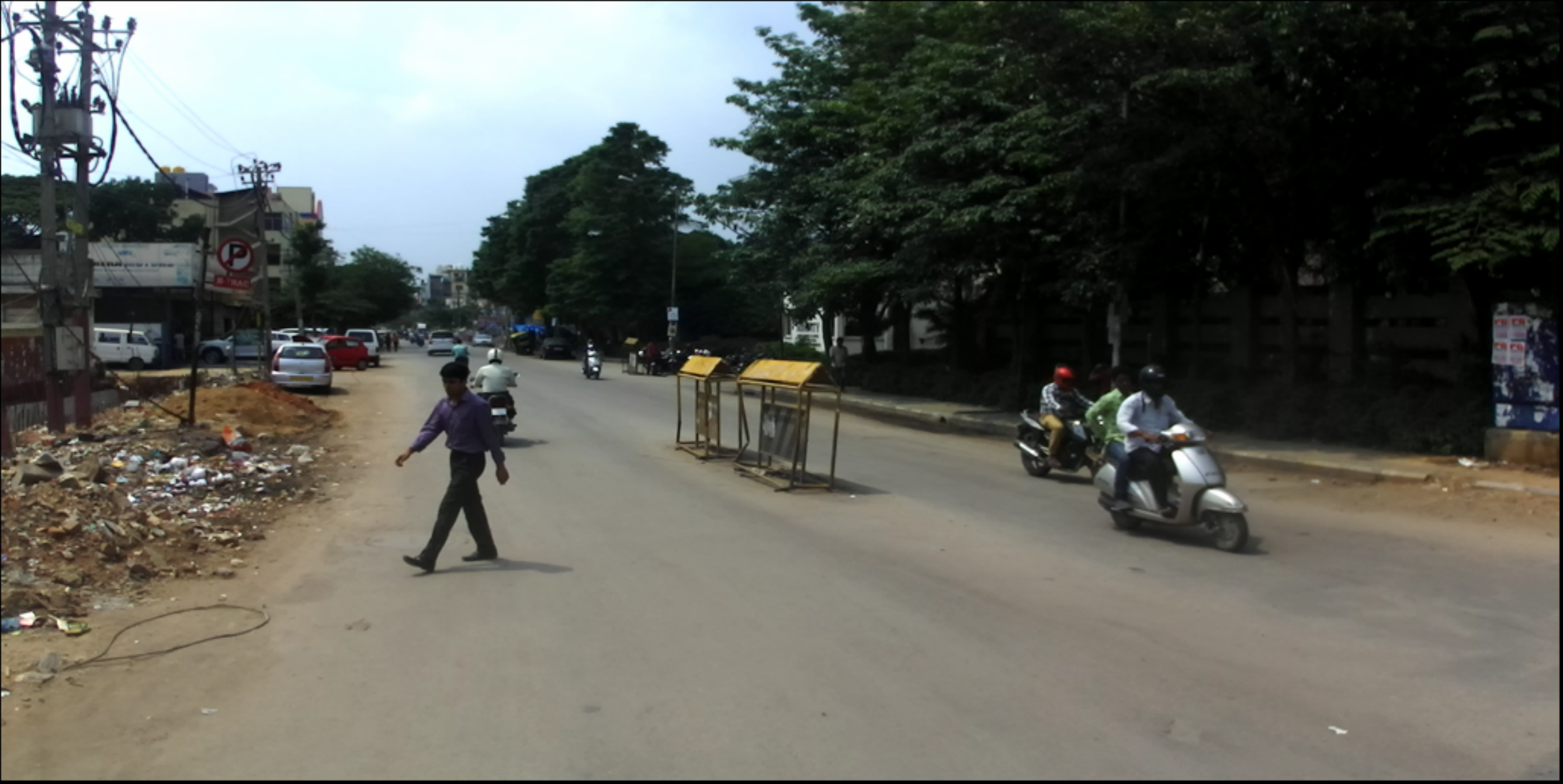}
    & \includegraphics[width=\w]{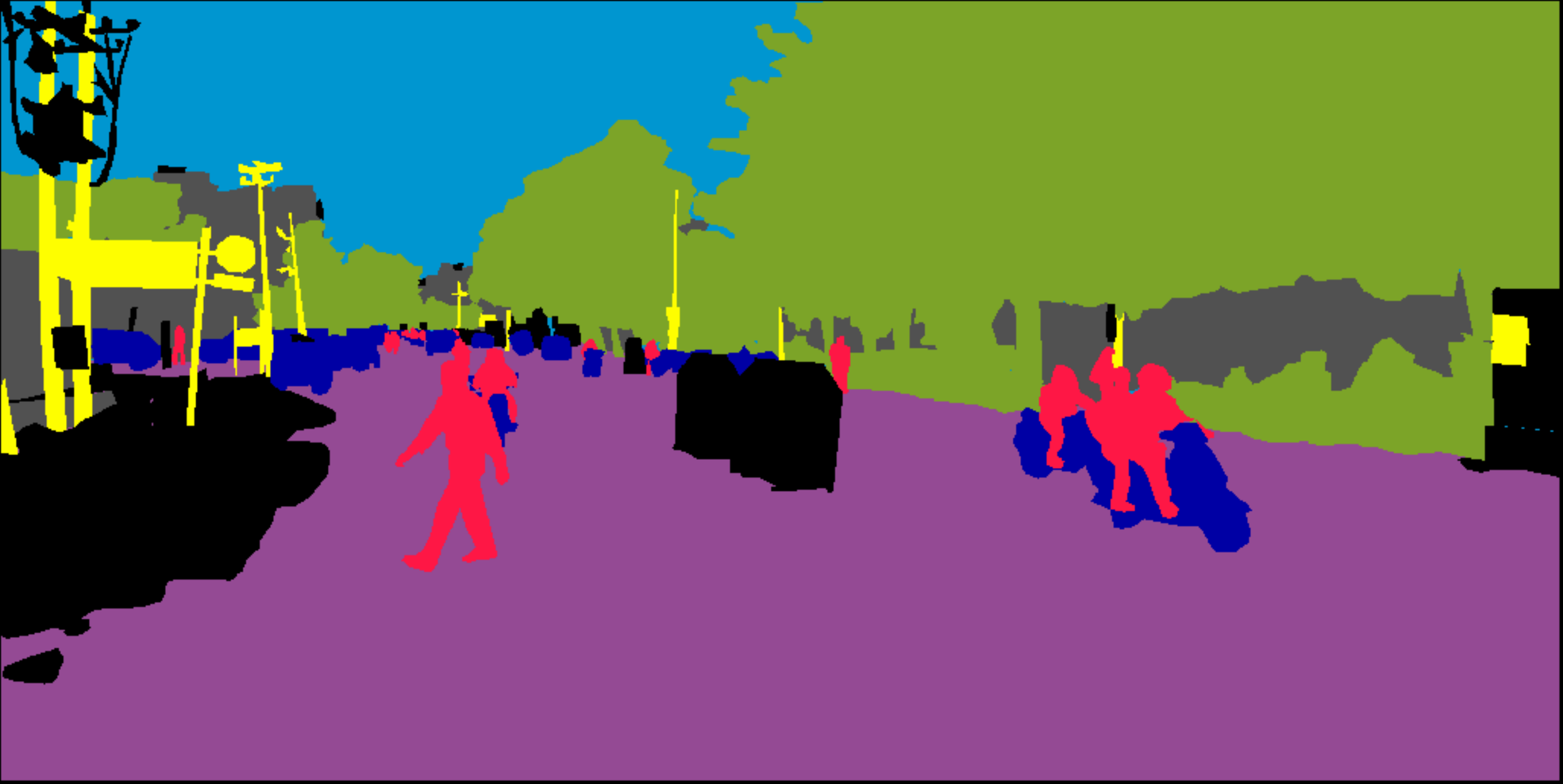}
    & \includegraphics[width=\w]{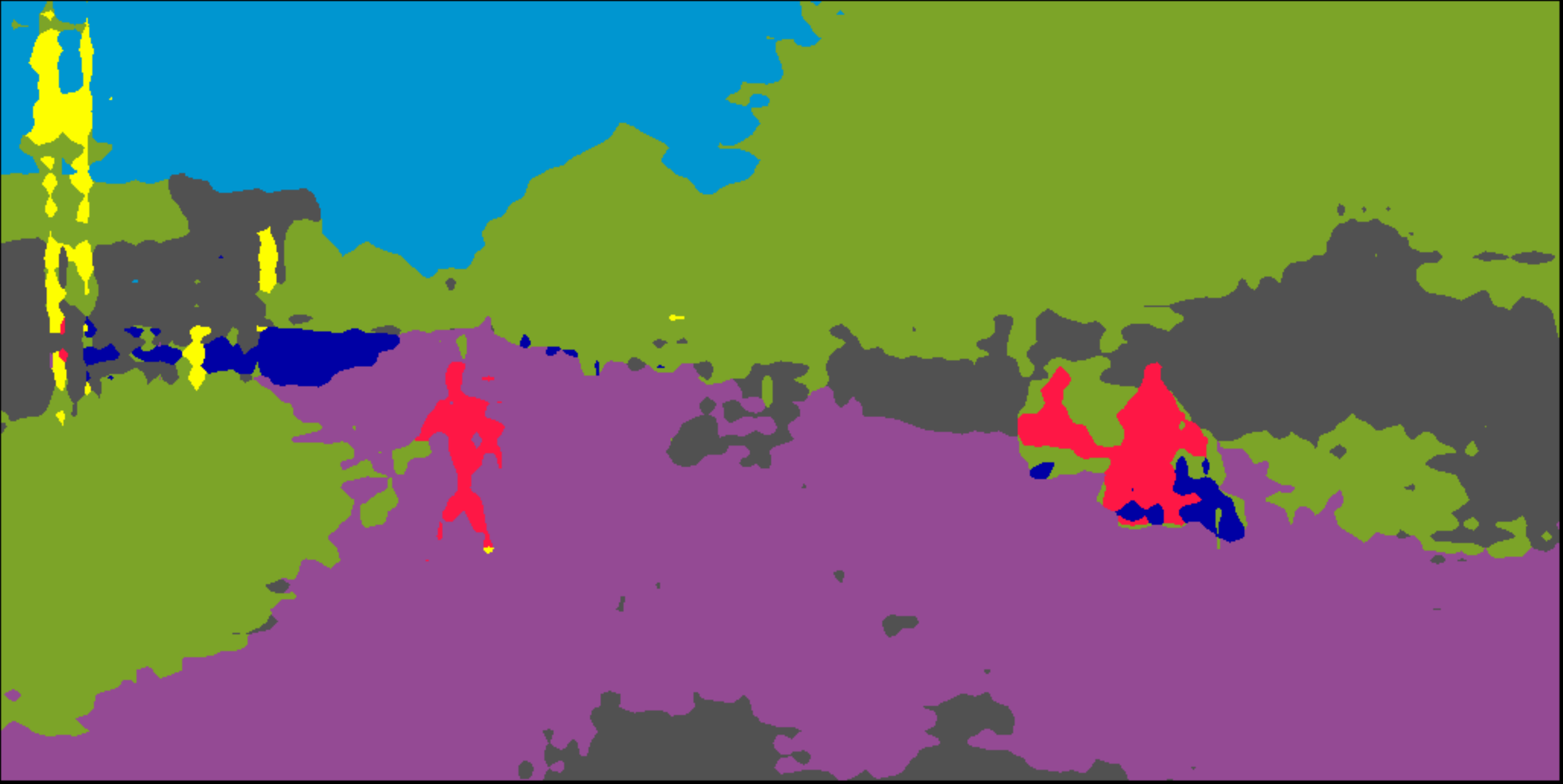}
    & \includegraphics[width=\w]{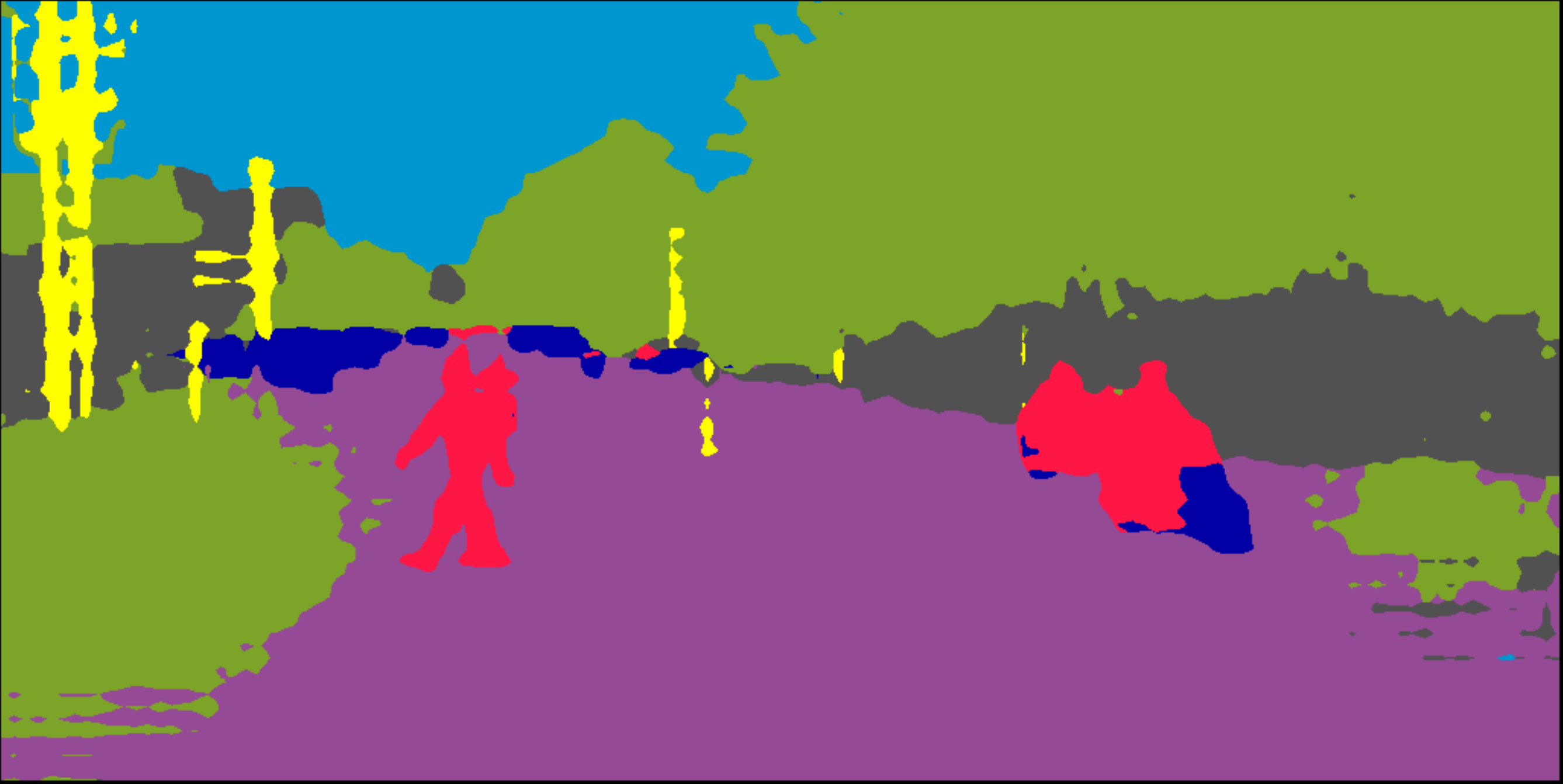}
    & \includegraphics[width=\w]{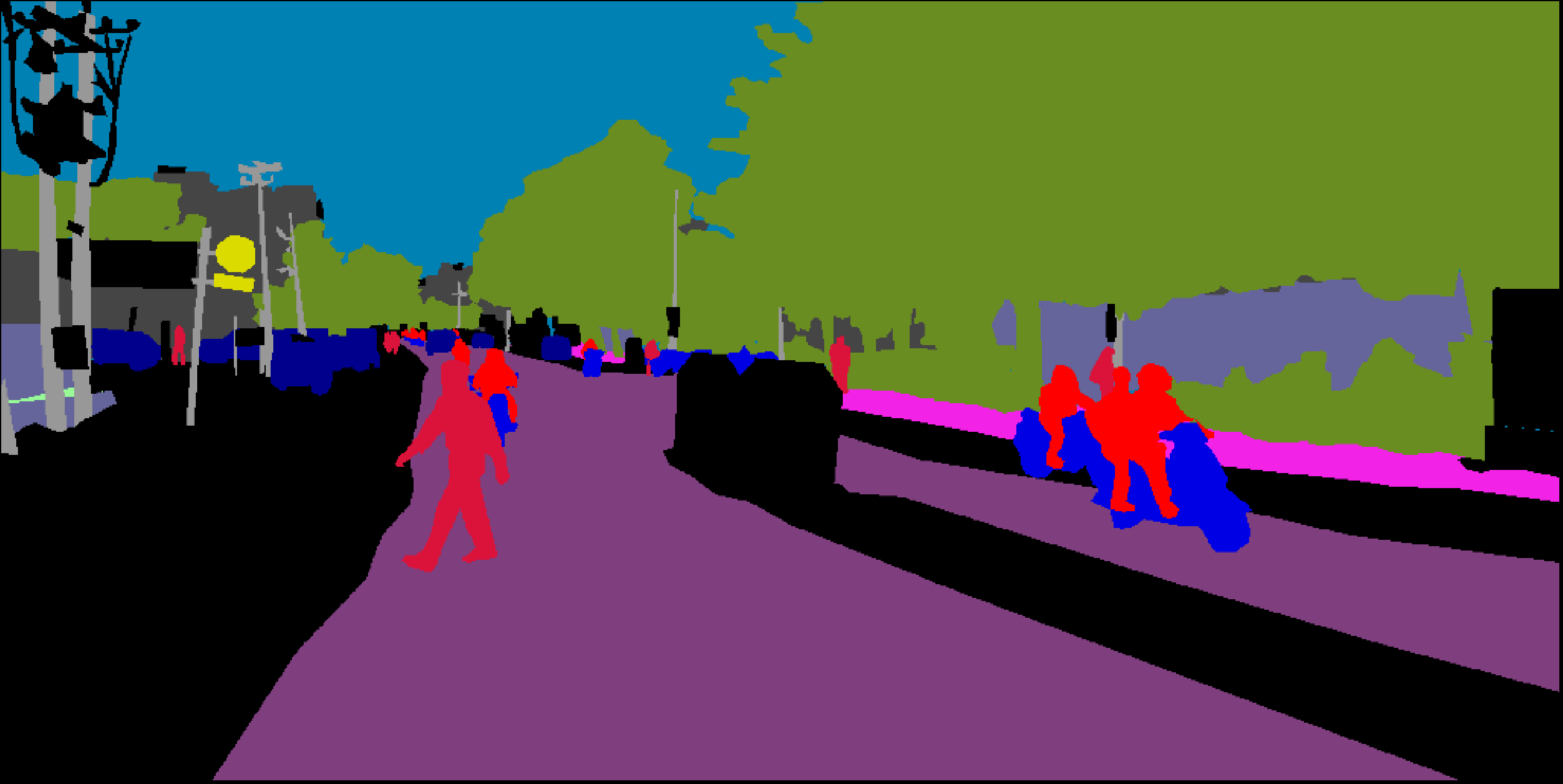}
    & \includegraphics[width=\w]{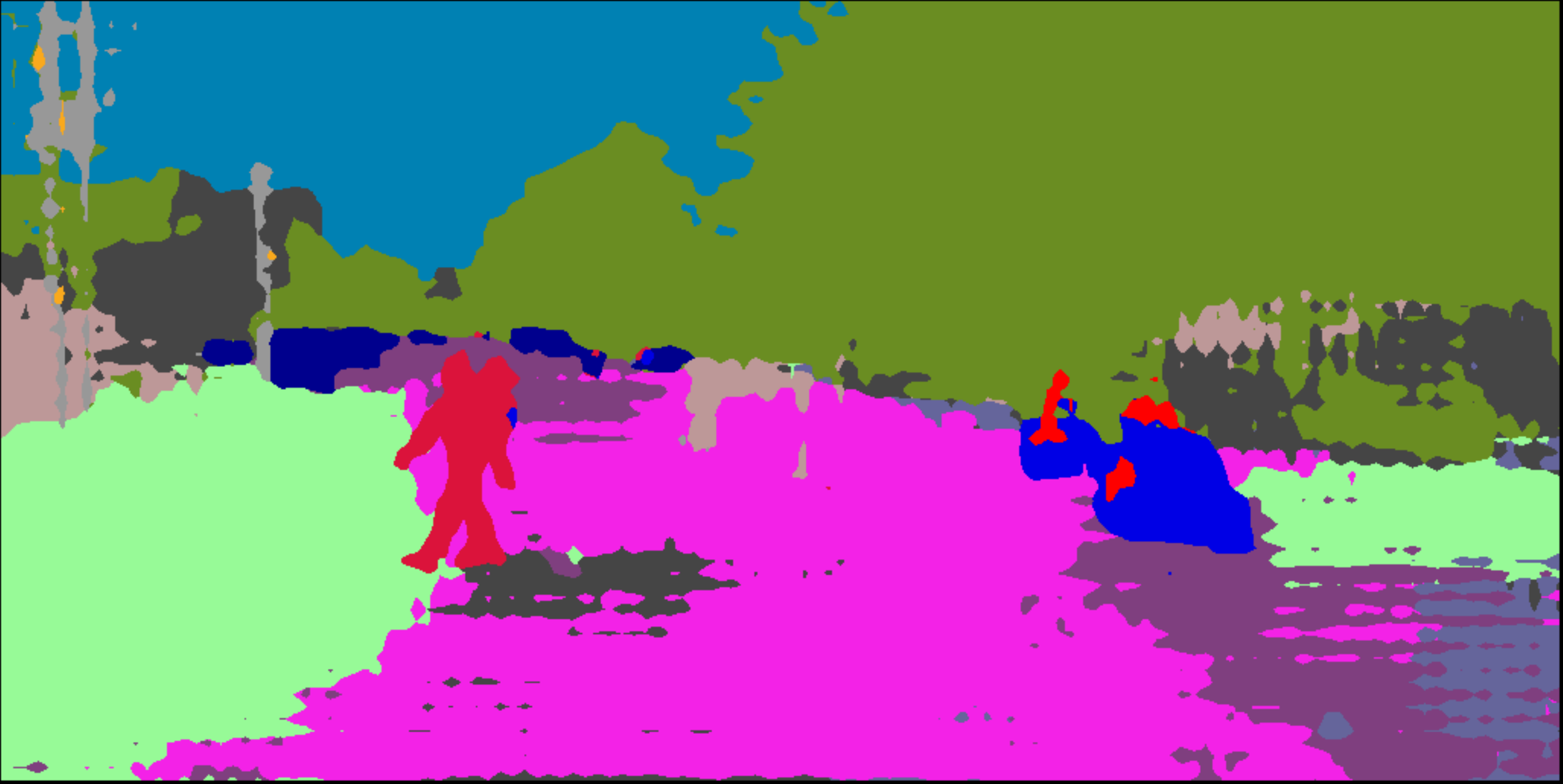}
    & \includegraphics[width=\w]{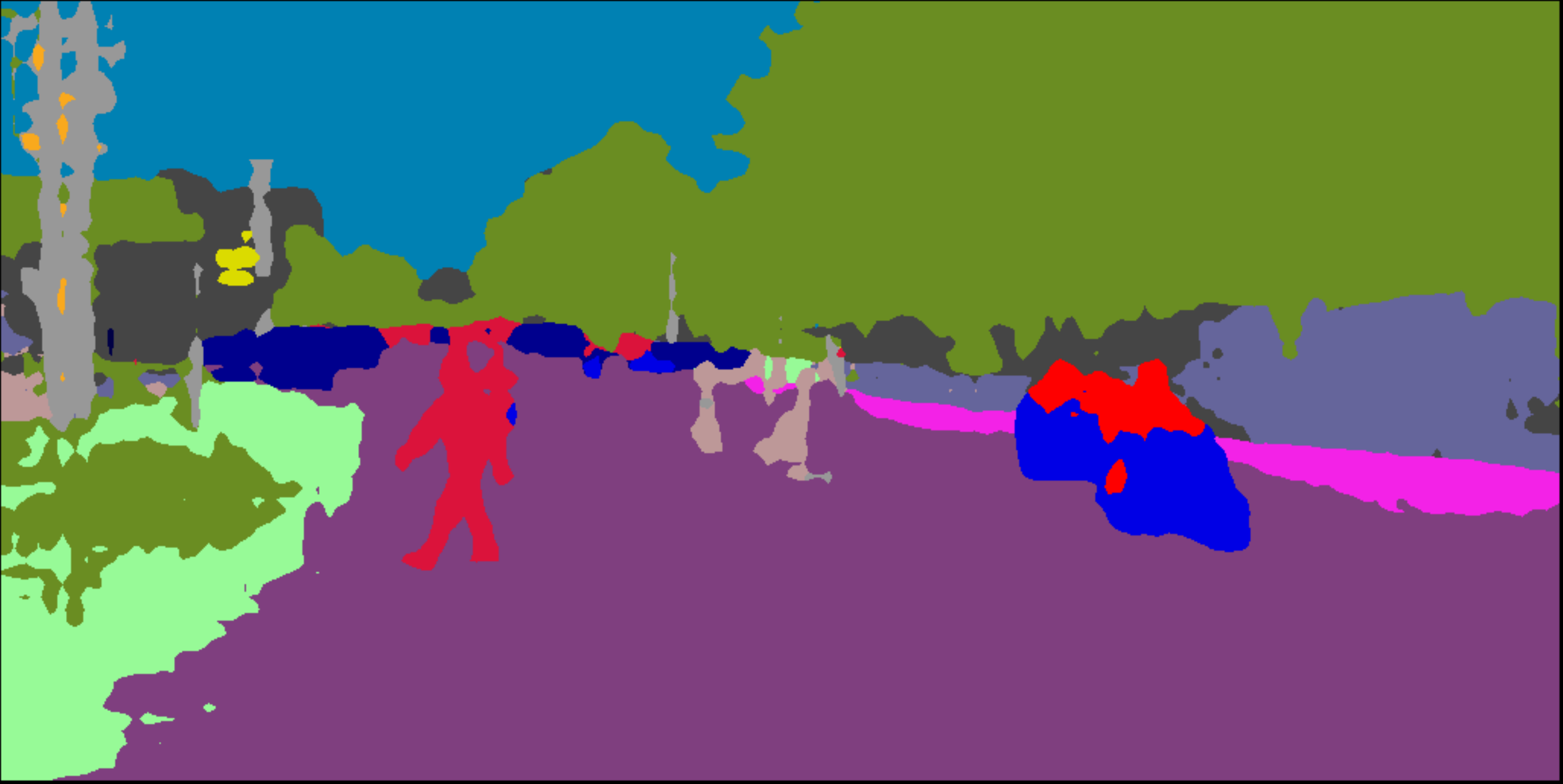}
    \\
    \multirow{4}{*}{M} 
    & \includegraphics[width=\w]{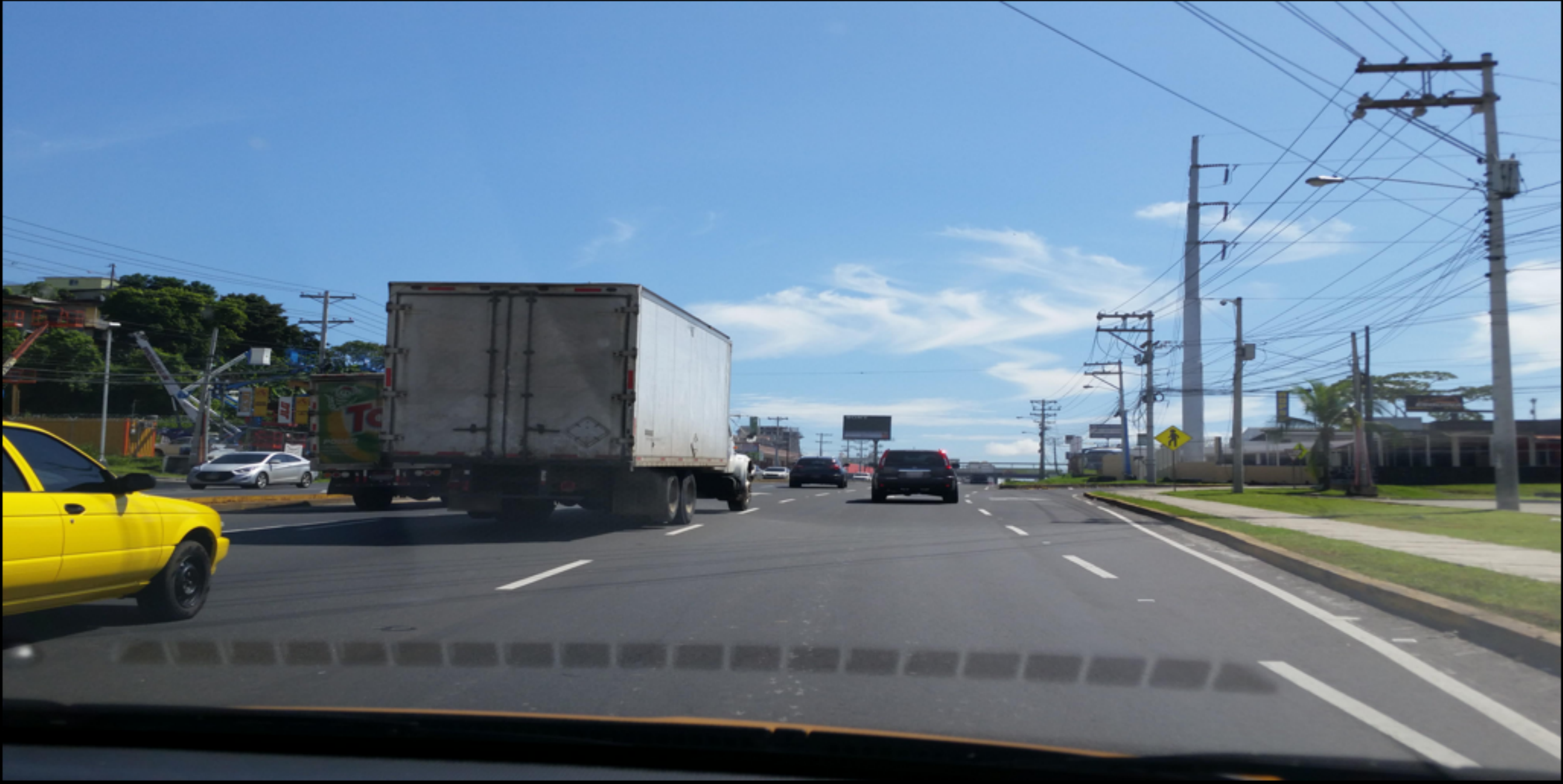} 
    & \includegraphics[width=\w]{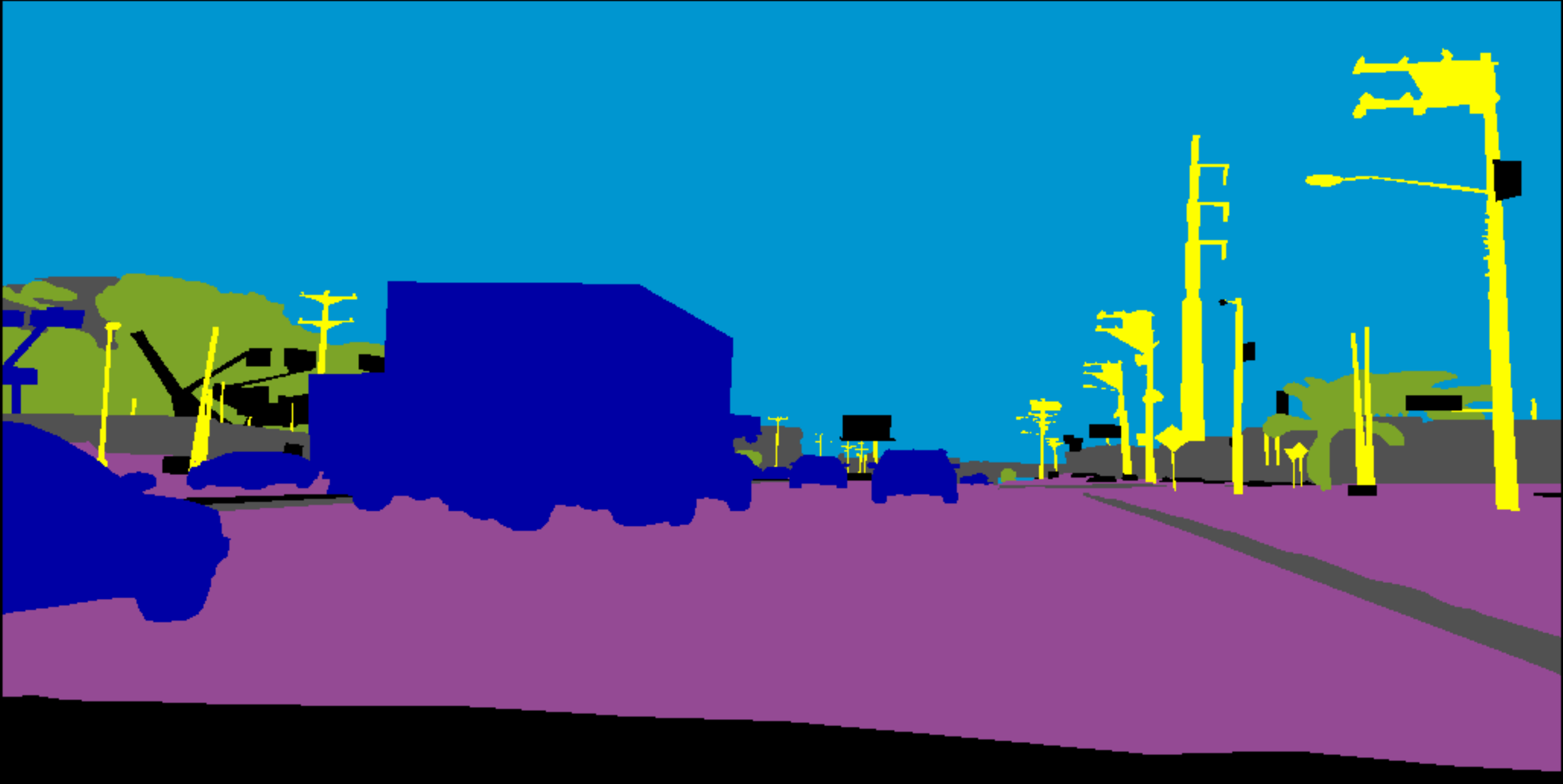}
    & \includegraphics[width=\w]{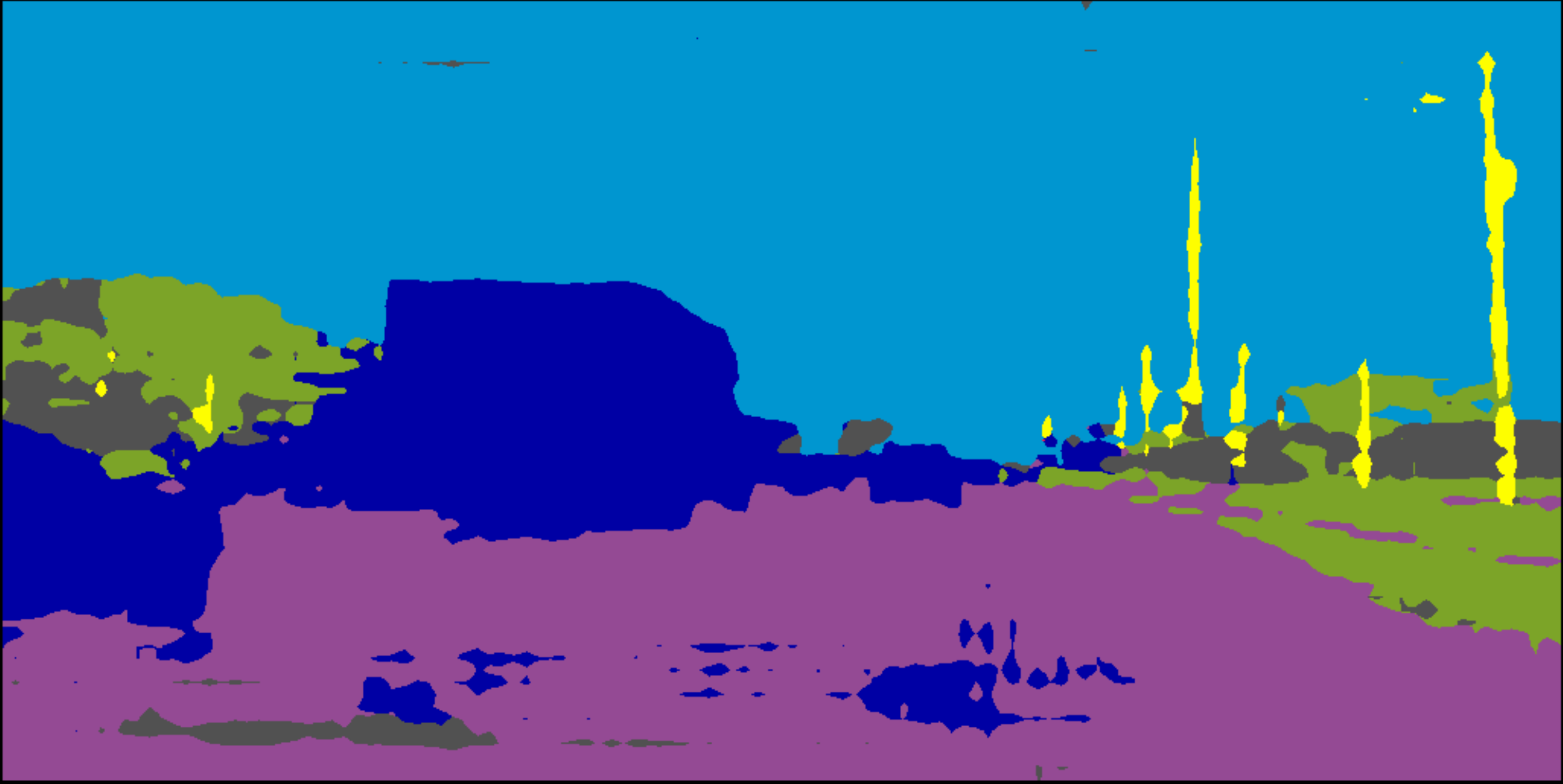}
    & \includegraphics[width=\w]{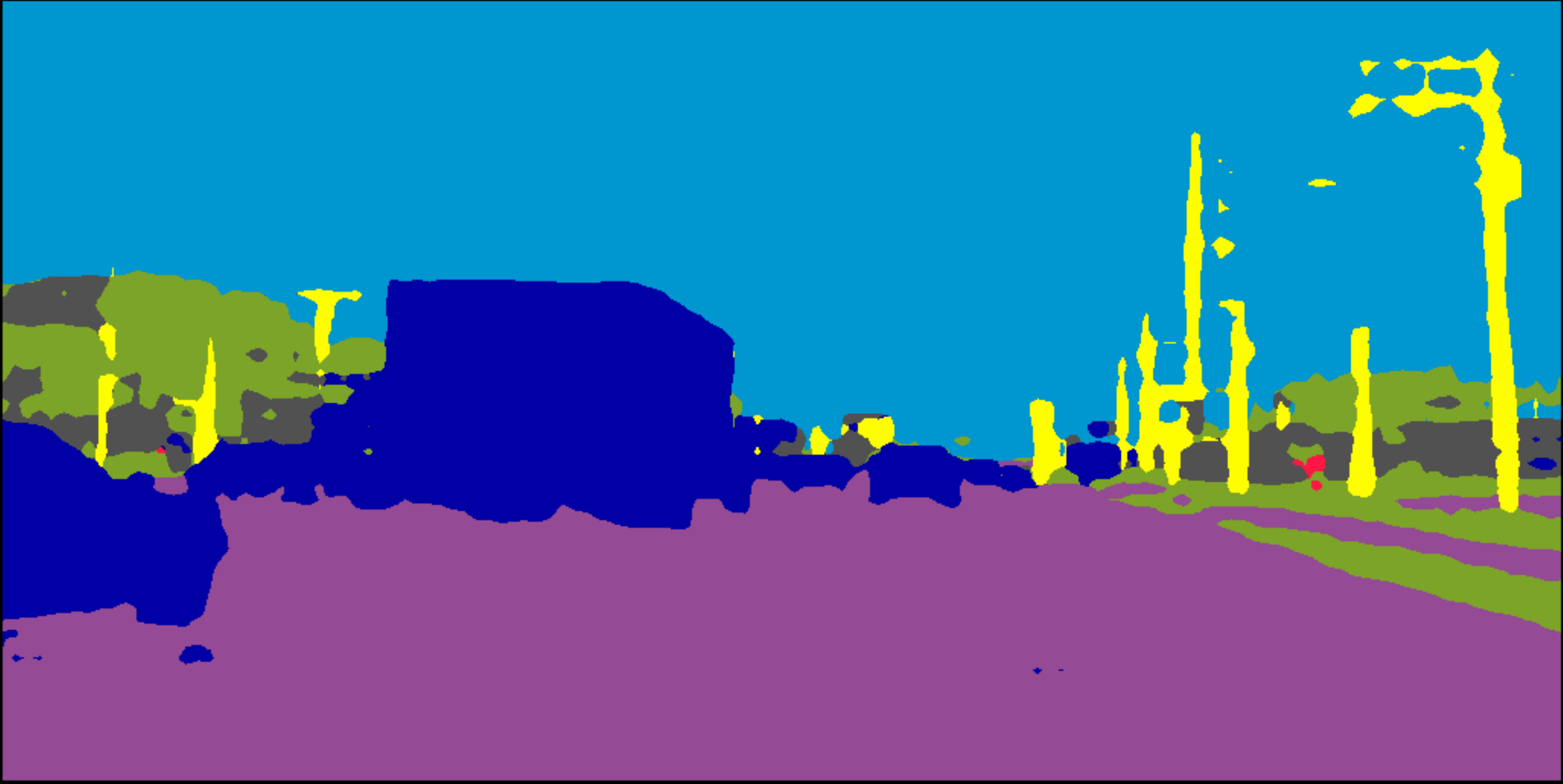}
    & \includegraphics[width=\w]{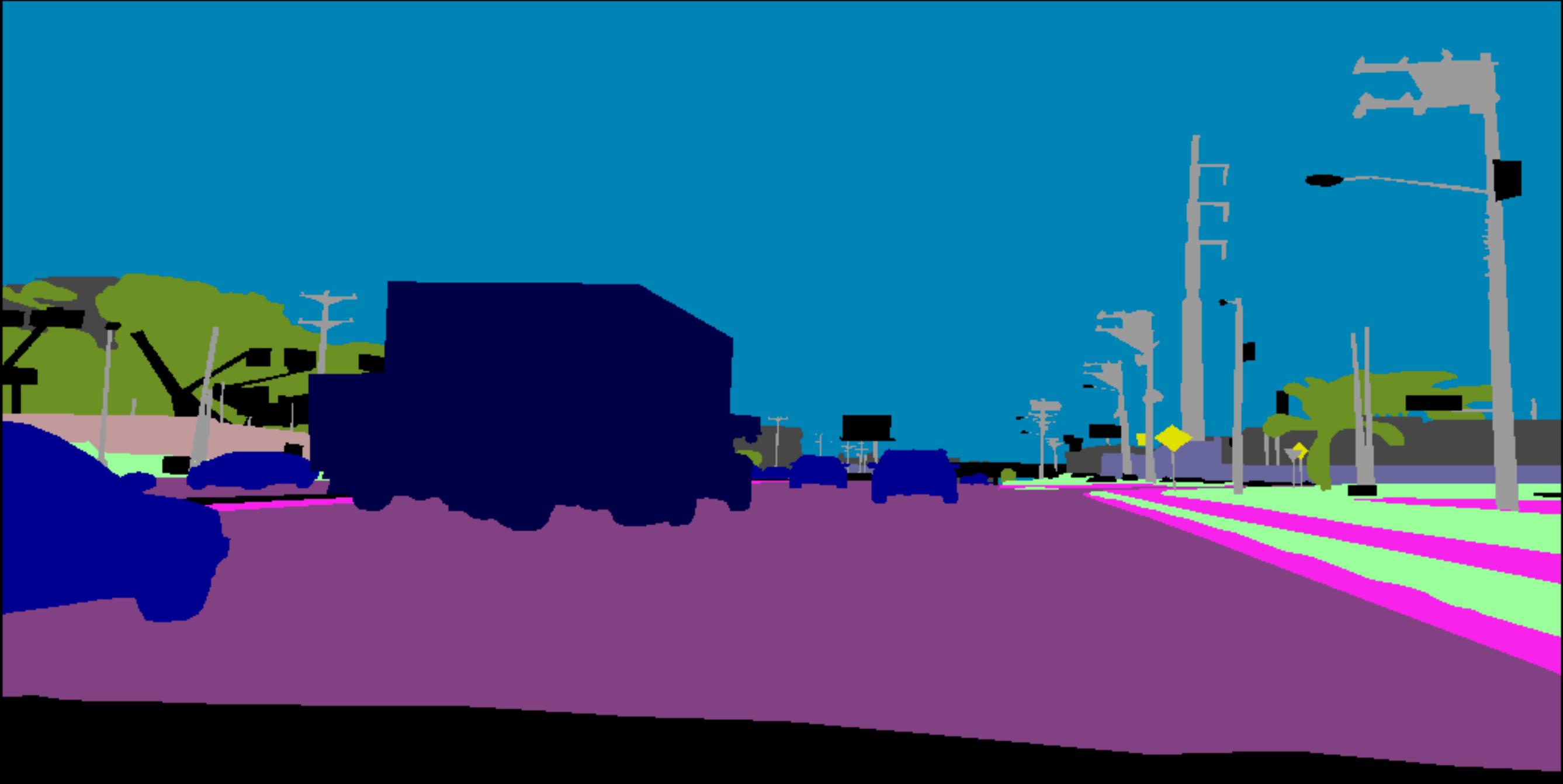}
    & \includegraphics[width=\w]{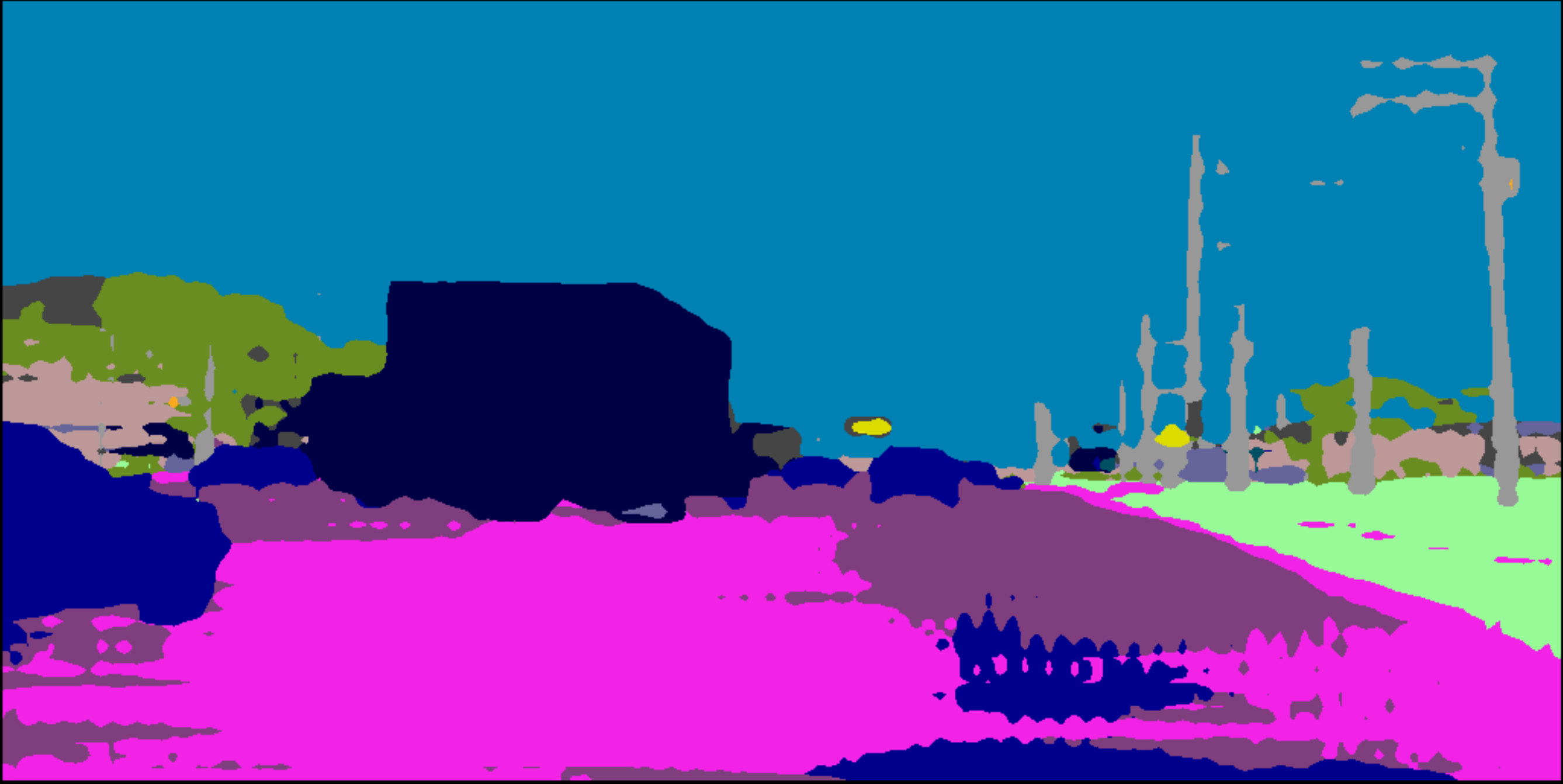}
    & \includegraphics[width=\w]{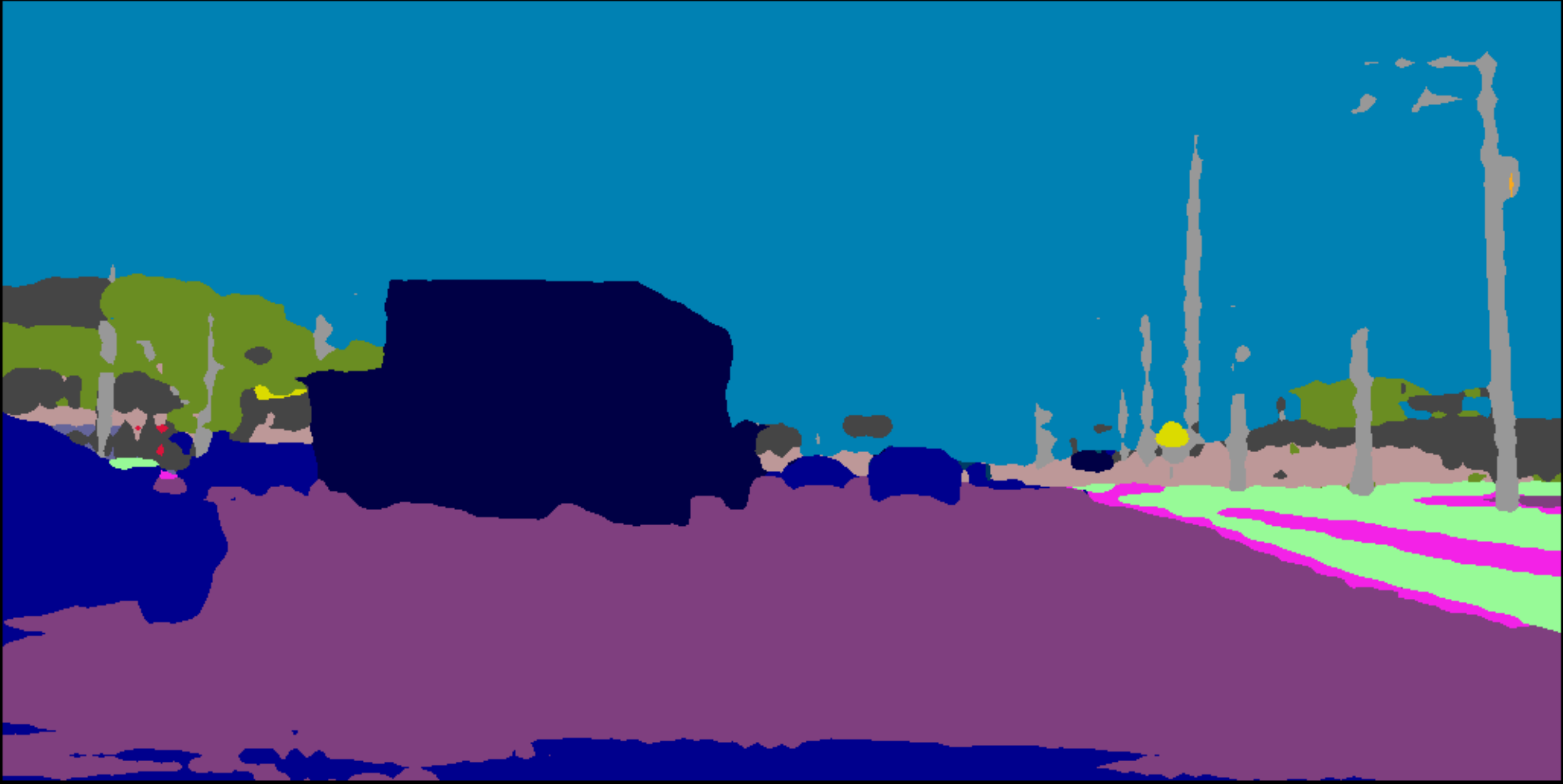}
    \\
    
    & \includegraphics[width=\w]{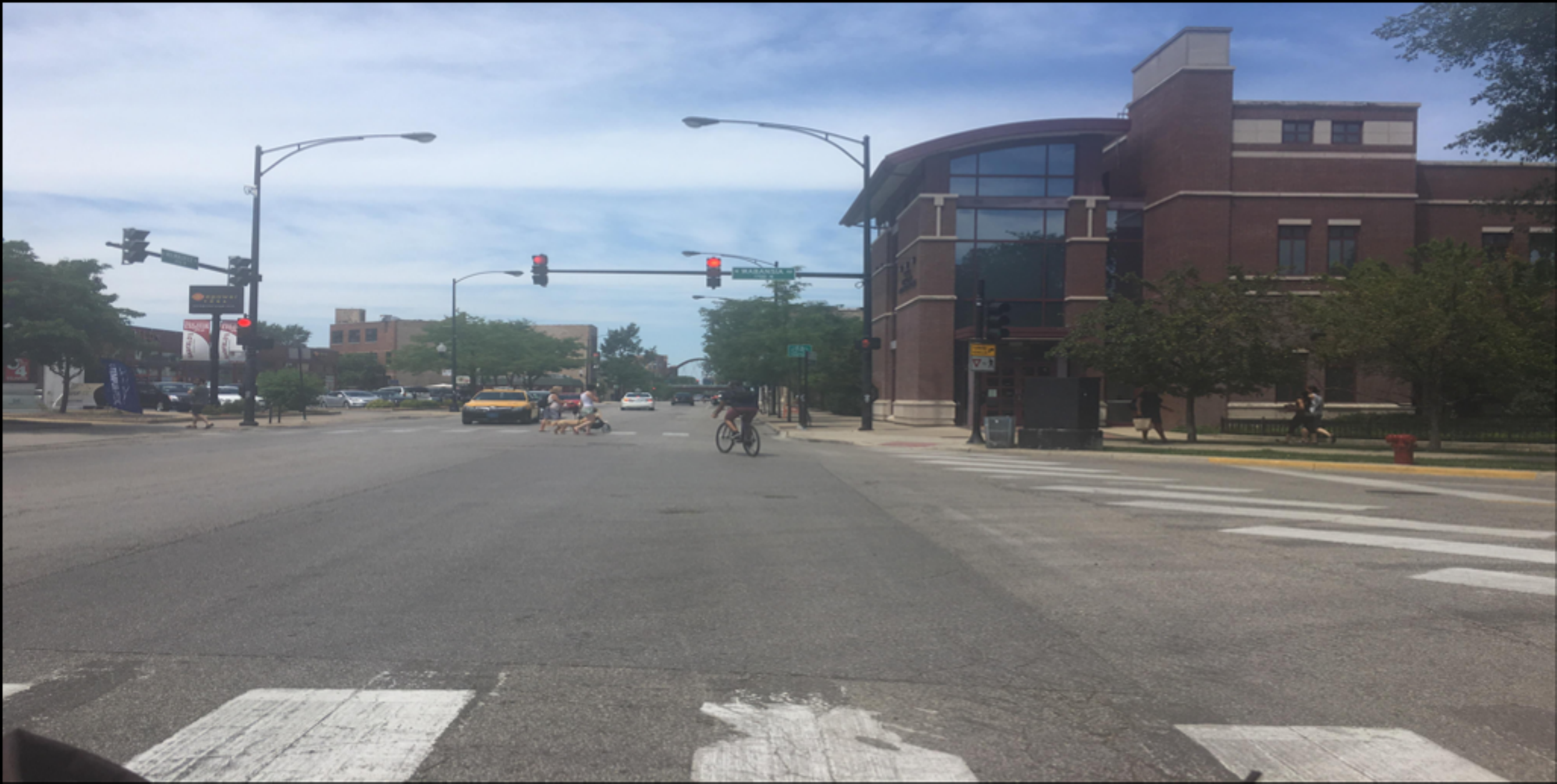}
    & \includegraphics[width=\w]{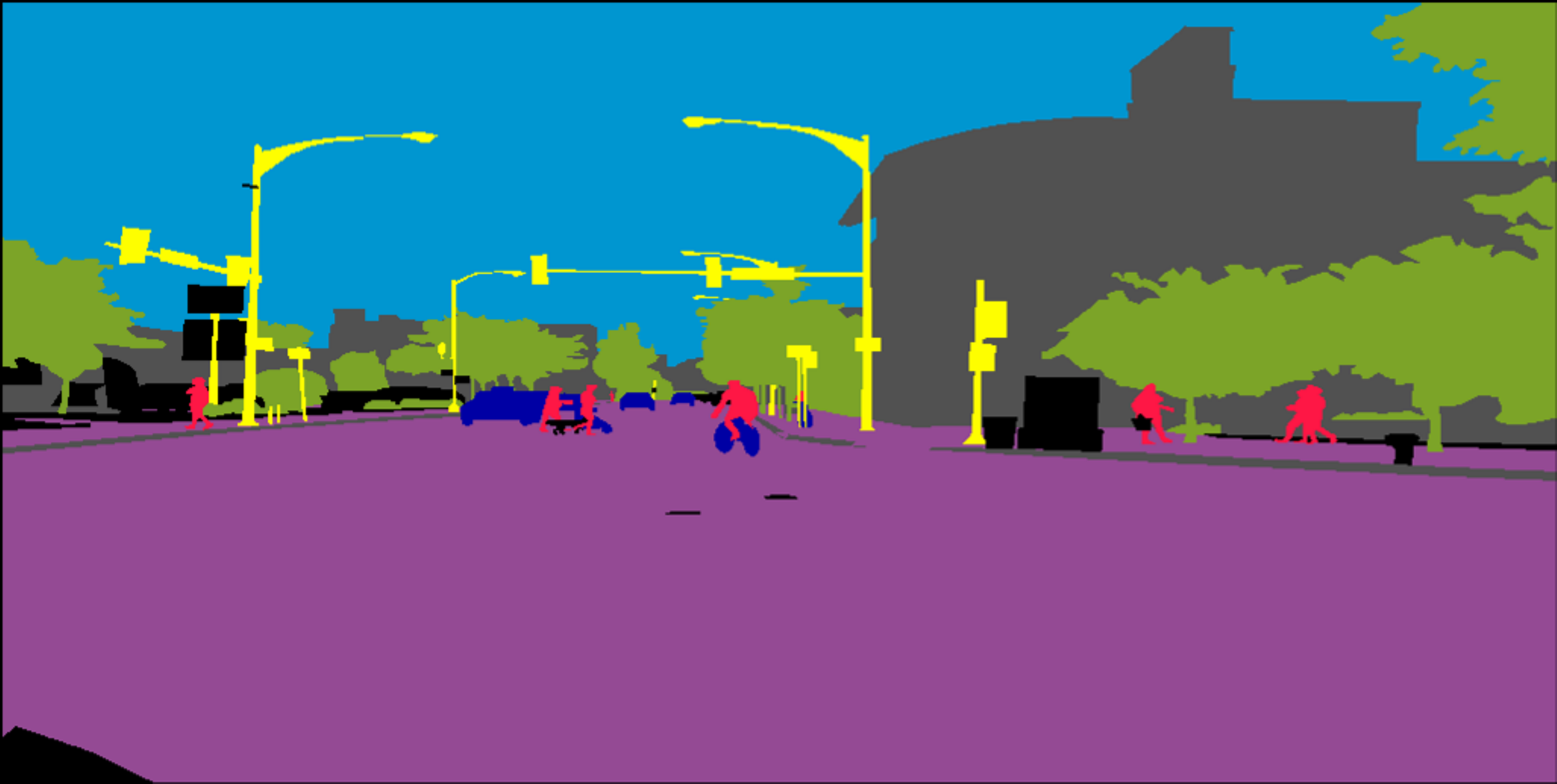}
    & \includegraphics[width=\w]{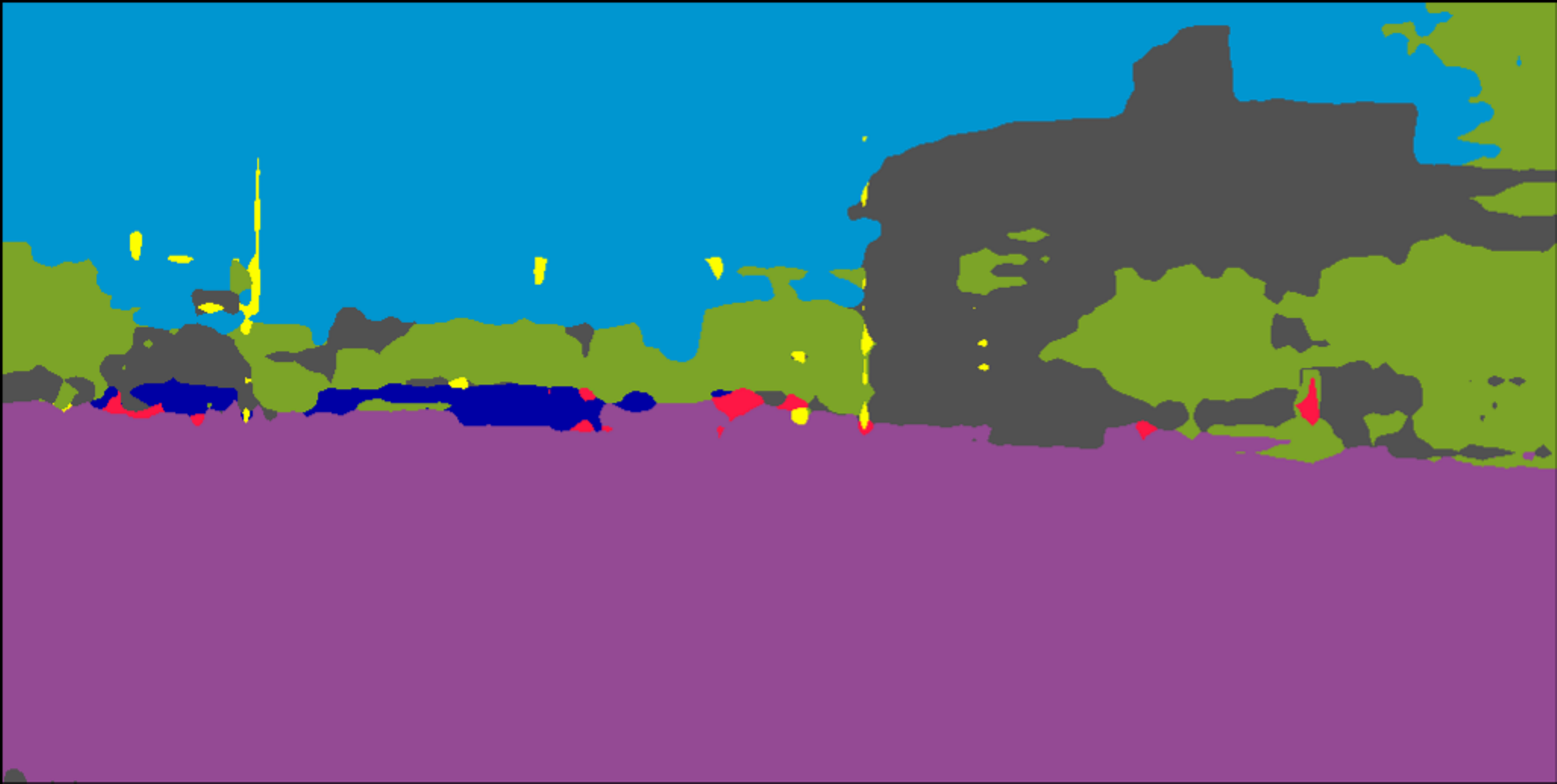}
    & \includegraphics[width=\w]{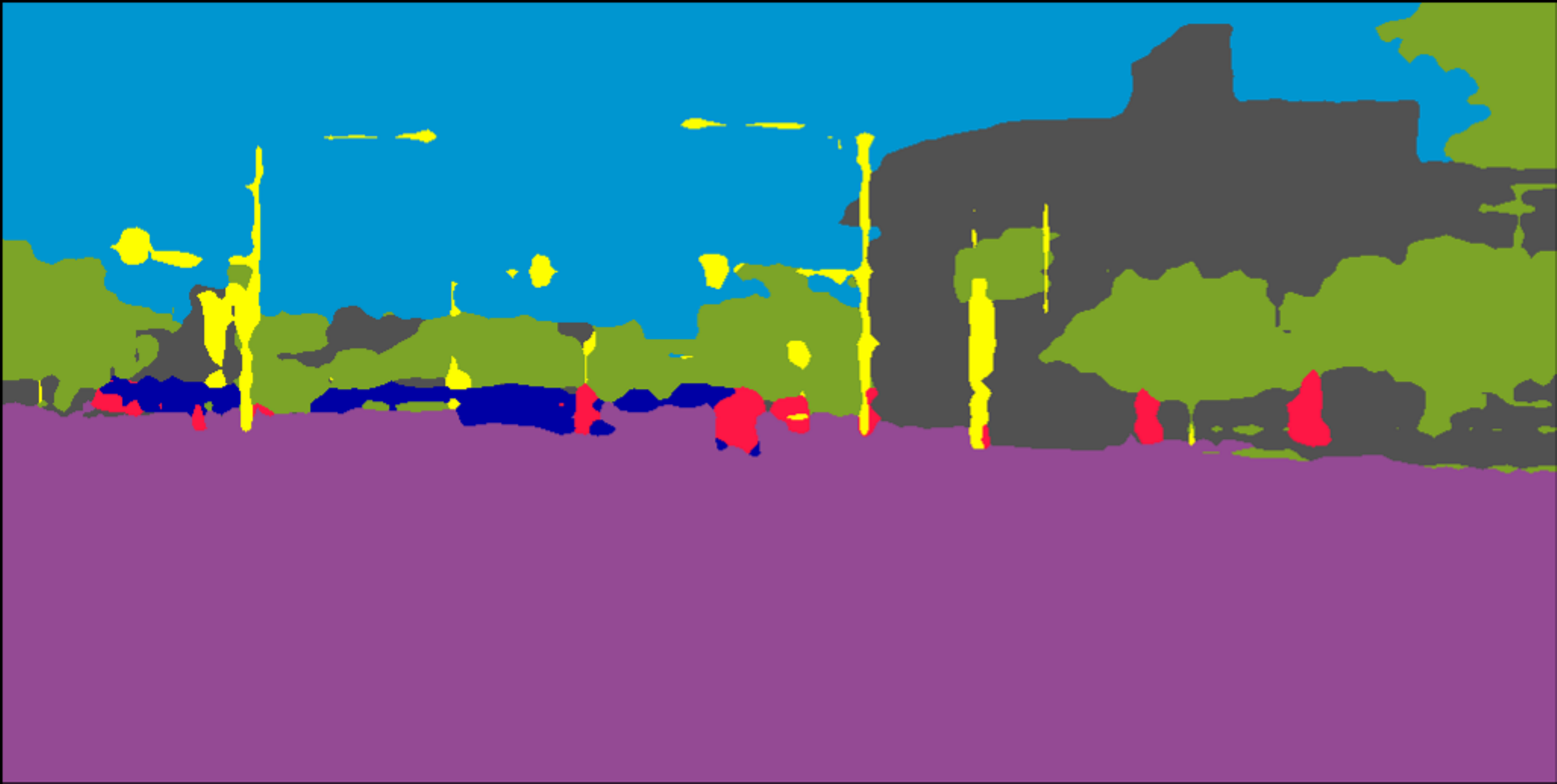}
    & \includegraphics[width=\w]{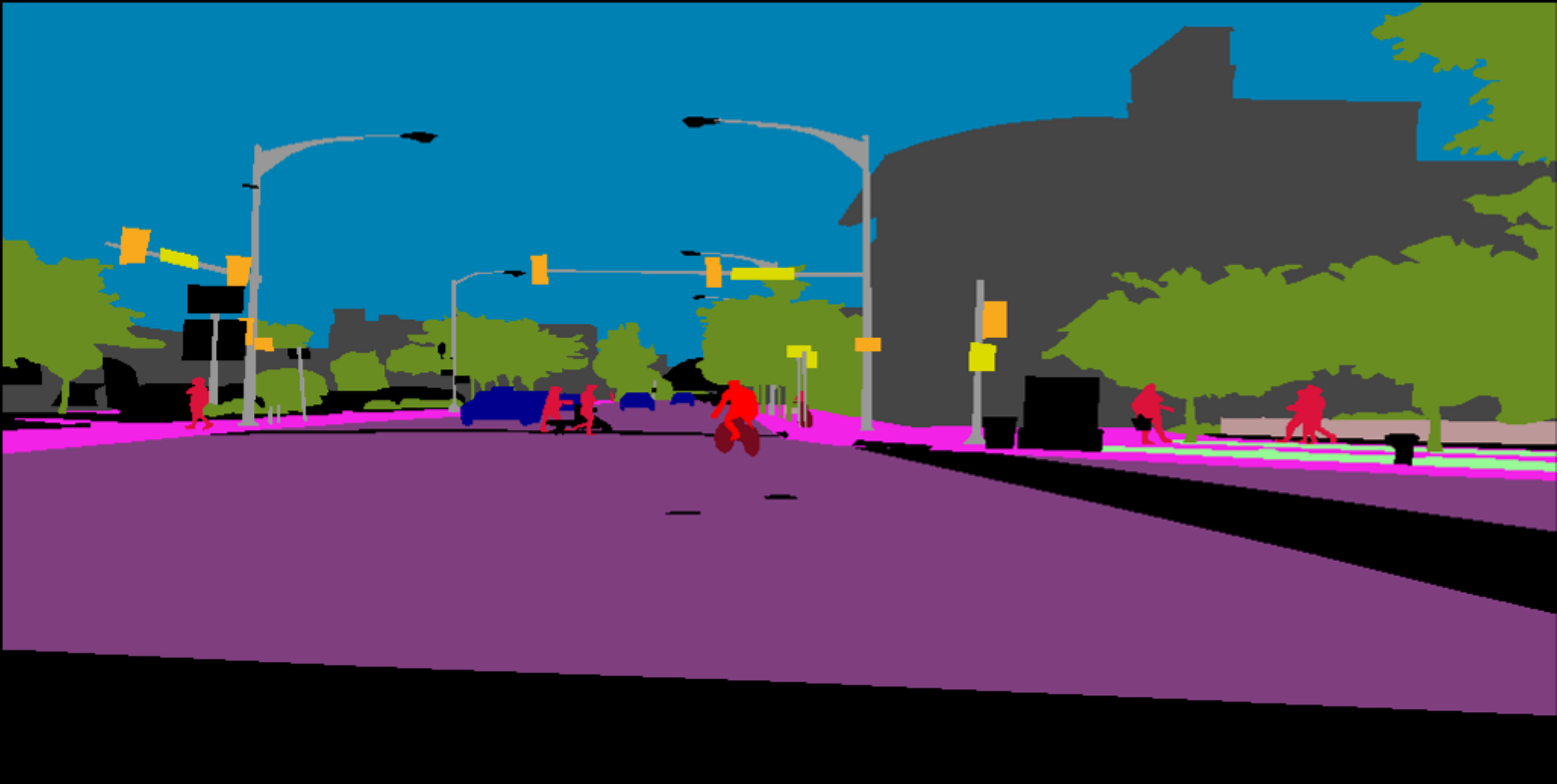}
    & \includegraphics[width=\w]{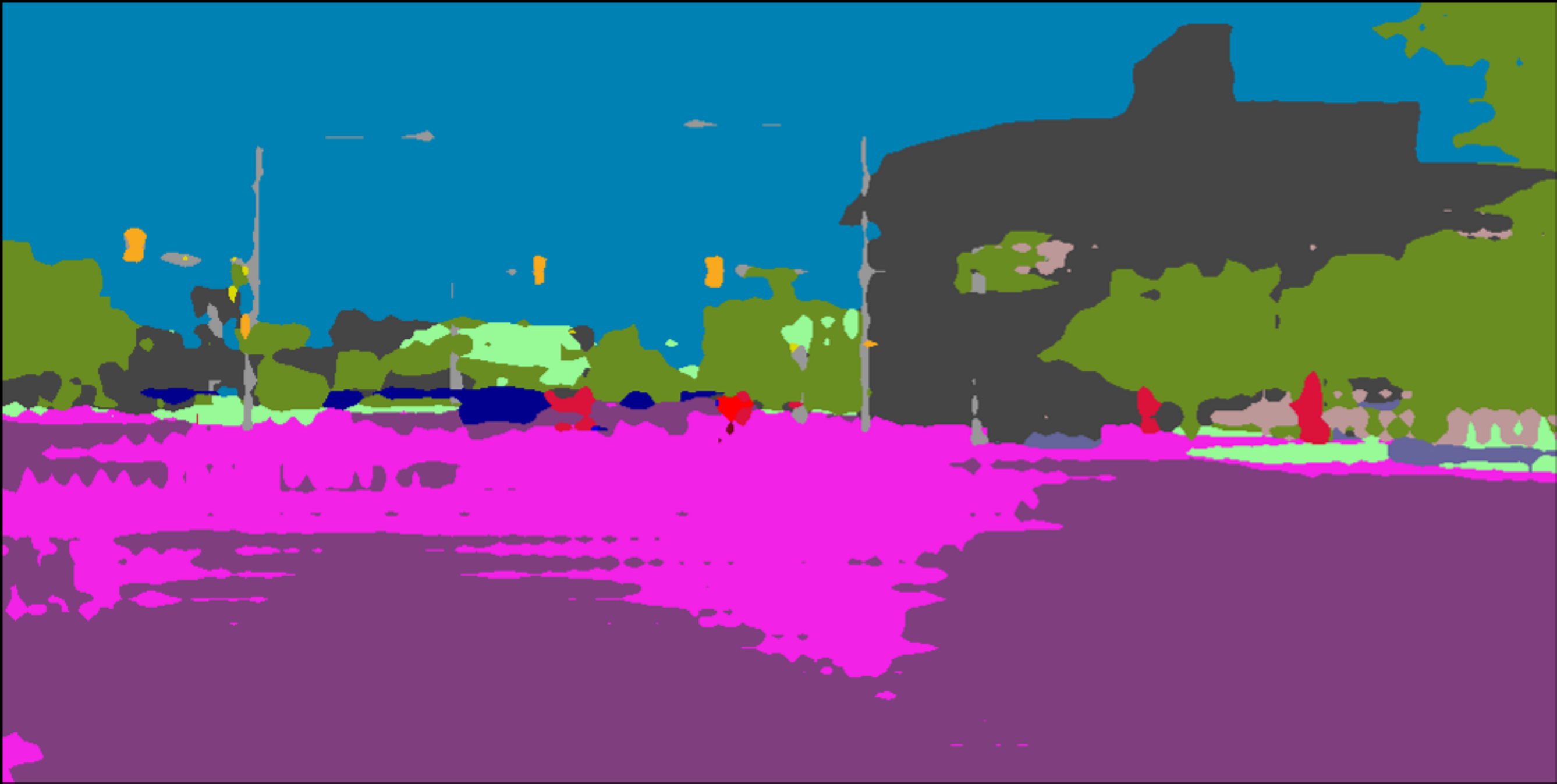}
    & \includegraphics[width=\w]{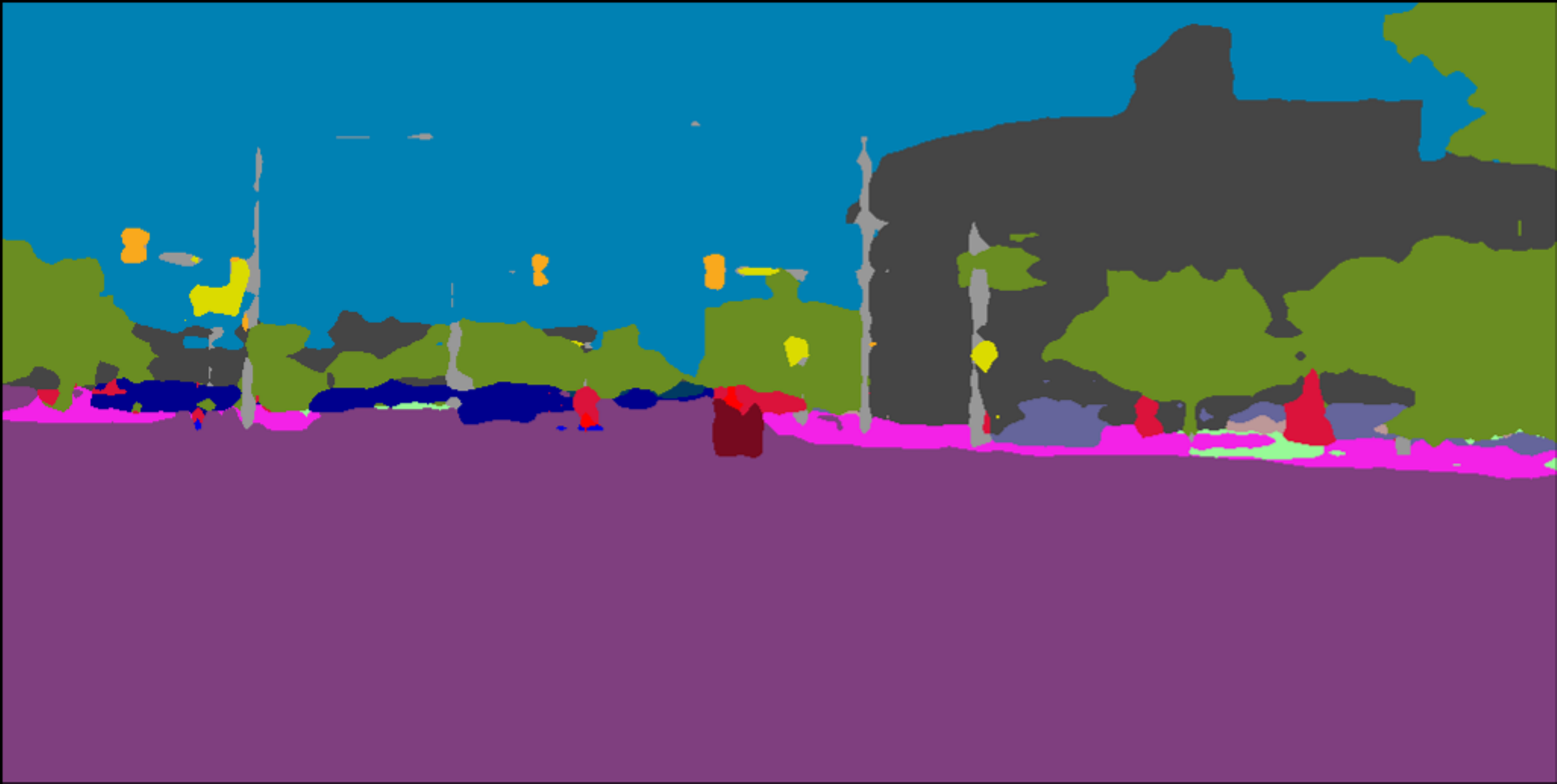}
    \\
    \end{tabular}
    \caption{Qualitative comparison between source only and our method on GTA5 (G) to Cityscapes (C), IDD (I), and Mapillary (M) with 7 classes and 19 classes setting.}
    \label{fig:seg}
    \vspace{-3mm}
\end{figure*}

%%%%%%%%% Experiment
\section{Experiments}
\label{sec:experiments}

In this section, we describe the implementation details and experimental results of the proposed ADAS.
We evaluate our method on a semantic segmentation task in both the synthetic-to-real adaptation in \secref{sec:syn2real} and the real-to-real adaptation in \secref{sec:real2real} with multiple target domain datasets.
We also conduct an extensive study to validate each submodule, MTDT-Net and BARS, in \secref{sec:Further}.

\begin{table}[t]
    \centering
    \begin{tabular}{c|c|ccc|c}
    \toprule
        & \multirow{2}{*}{Method} & \multicolumn{3}{c|}{mIoU} & mIoU \\
        & & C & I & M & Avg.\\
        \hline
        \multirow{2}{*}{G$\to$C, I} & CCL\cite{isobe2021multi} & 45.0 & 46.0 & - & 45.5 \\
        & Ours & \textbf{45.8} & \textbf{46.3} & - & \textbf{46.1} \\
        \hline
        \multirow{2}{*}{G$\to$C, M} & CCL\cite{isobe2021multi} & 45.1 & - & 48.8 & 46.8 \\
        & Ours & \textbf{45.8} & - & \textbf{49.2} & \textbf{47.5} \\
        \hline
        \multirow{2}{*}{G$\to$I, M} & CCL\cite{isobe2021multi} & - & 44.5 & 46.4 & 45.5 \\
        & Ours & - & \textbf{46.1} & \textbf{47.6} & \textbf{46.9} \\
        \hline
        \multirow{2}{*}{G$\to$C, I, M} & CCL\cite{isobe2021multi} & 46.7 & 47.0 & 49.9 & 47.9 \\
        & Ours & \textbf{46.9} & \textbf{47.7} & \textbf{51.1} & \textbf{48.6} \\
    \bottomrule
    \end{tabular}
    \caption{Results of adapting GTA5 to Cityscapes (C), IDD (I), and Mapillary (M) with 19 classes setting.}
    \label{tab:19classes}
    \vspace{-3mm}
\end{table}

\subsection{Training Details}
\label{sec:training_details}

\noindent\textbf{Datasets}
We use four different driving datasets containing one synthetic and three real-world datasets, each of which has a unique scene structure and visual appearance.
\begin{itemize}
\setlength\itemsep{0em}
\item GTA5\cite{richter2016playing} is a synthetic dataset of 24,966 labeled images captured from a video game.
\item Cityscapes\cite{cordts2016cityscapes} is an urban dataset collected from European cities, and includes 2,975 images in the training set and 500 in the validation set.
\item IDD\cite{varma2019idd} has total 10,003 Indian urban driving scenes, which contains 6,993 images for training, 981 for validation and 2,029 for test.
\item Mapillary Vista\cite{neuhold2017mapillary} is a large-scale dataset that contains multiple city scenes from around the world with 18,000 images for training and 2,000 for validation.
\end{itemize}

For a fair comparison with the recent MTDA methods~\cite{vu2019advent,saporta2021multi,isobe2021multi}, we follow the segmentation label mapping protocol of 19 classes and super classes (7 classes) proposed in the papers.
We use mIoU (\%) as evaluation metric for all domain adaptation experiments.

\begin{figure}[t] 
% 	\centering
	\newcommand\w{2.5cm}
% 	\newcolumntype{M}[1]{>{\centering\arraybackslash}m{#1}}
	\begin{tabular}{l@{\hspace{1mm}}
	>{\centering\arraybackslash}m{\w}@{\hspace{1mm}}
	>{\centering\arraybackslash}m{\w}@{\hspace{1mm}}
	>{\centering\arraybackslash}m{\w}}
	& C$\to$ I,M & I$\to$ C,M & M$\to$ C,I \\
    C & \includegraphics[width=\w]{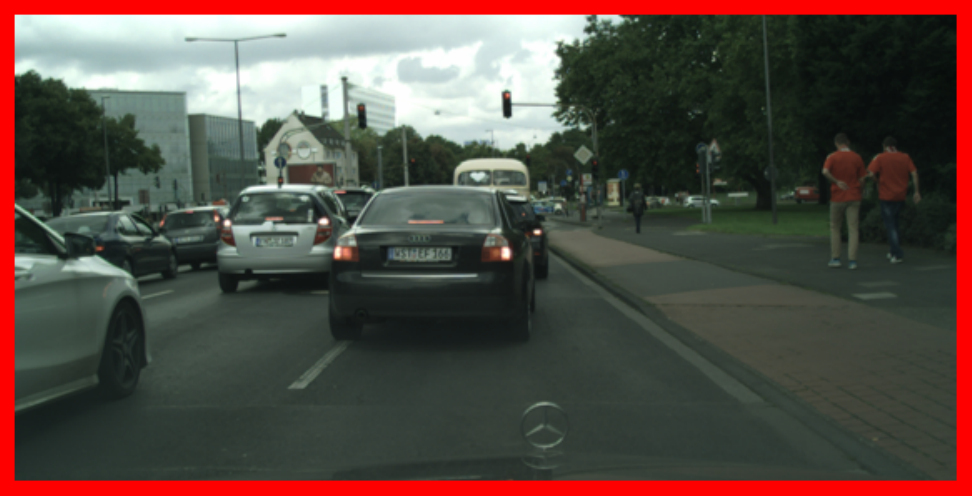}
    & \includegraphics[width=\w]{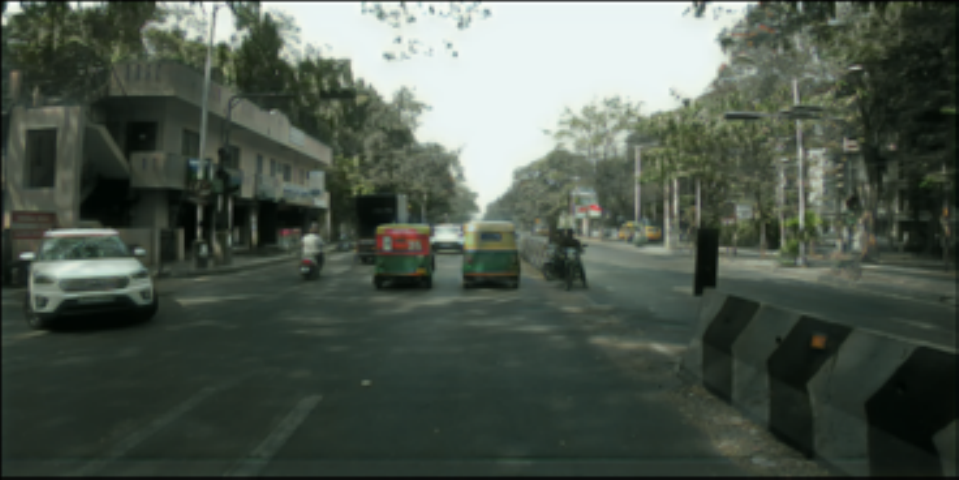}
    &\includegraphics[width=\w]{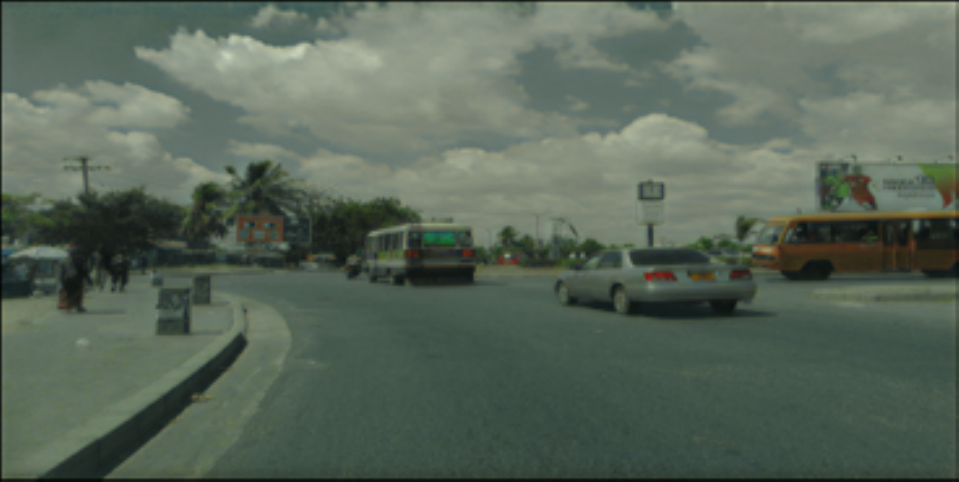}\\
    
    I & \includegraphics[width=\w]{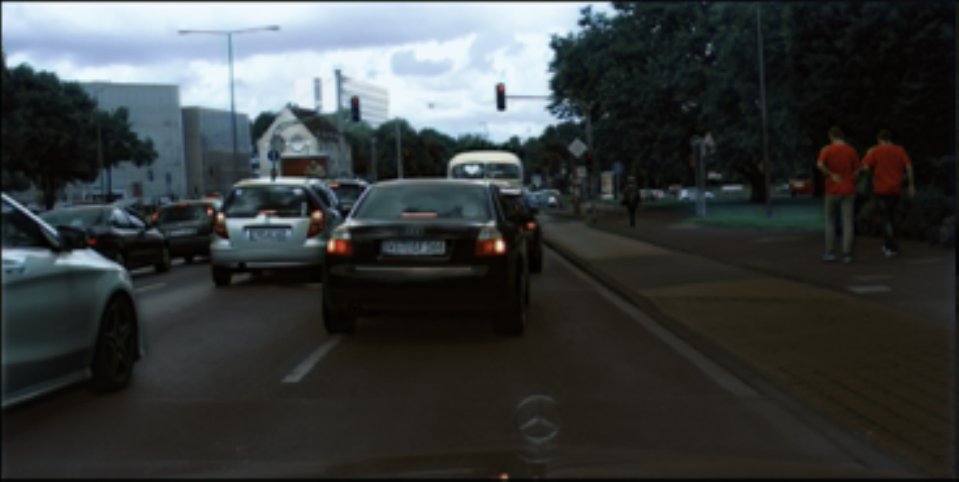}
    & \includegraphics[width=\w]{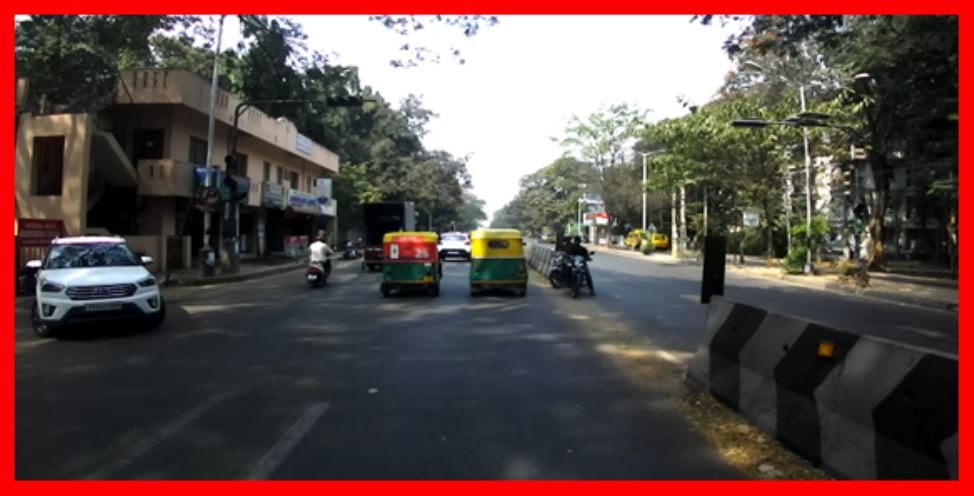}
    & \includegraphics[width=\w]{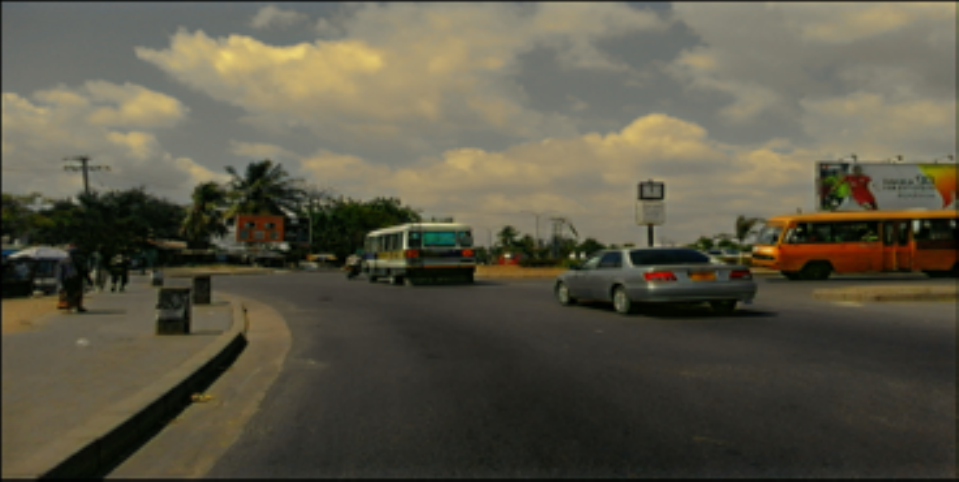}\\
    
    M & \includegraphics[width=\w]{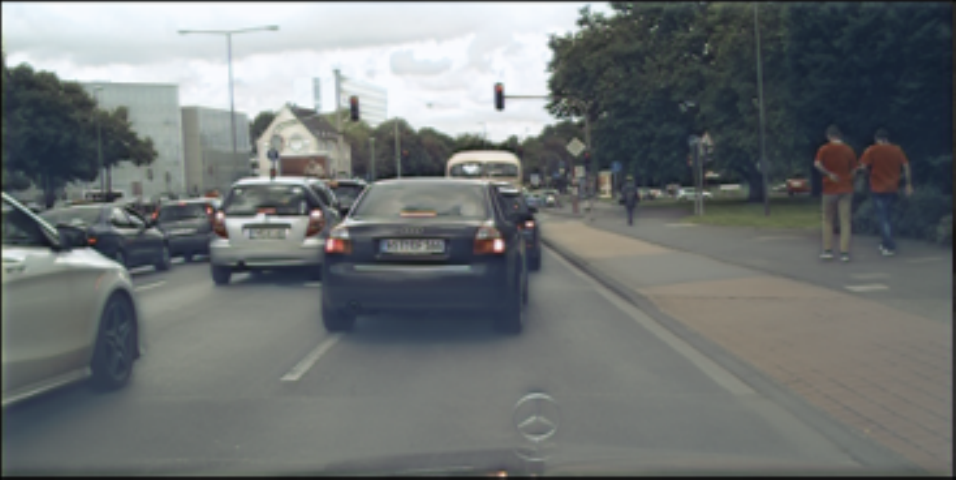}
    & \includegraphics[width=\w]{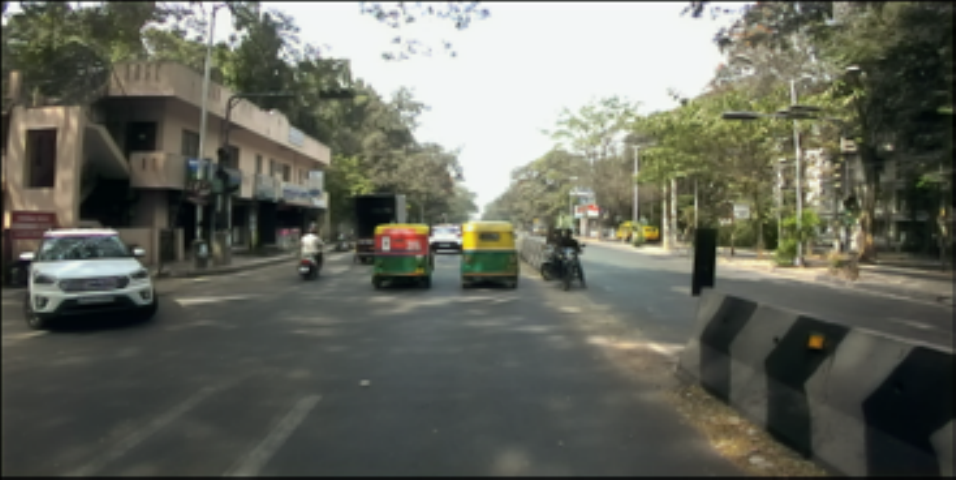}
    & \includegraphics[width=\w]{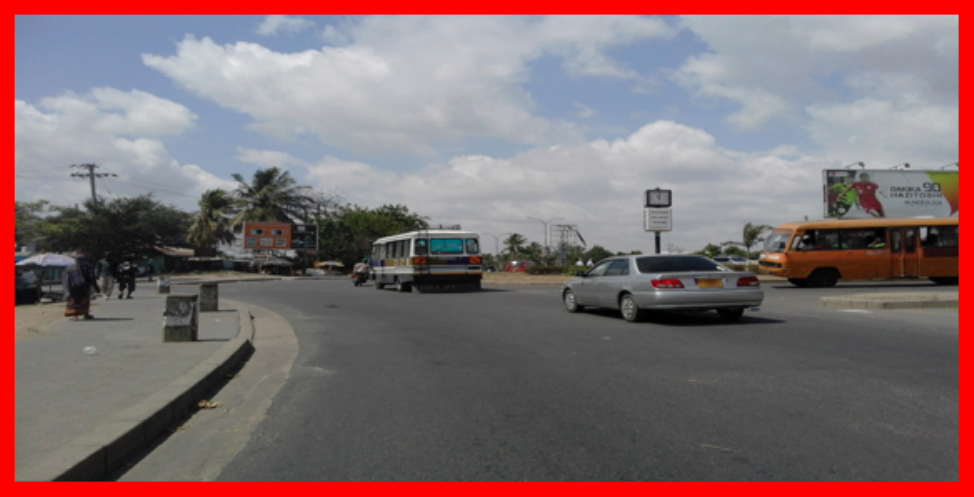} \\
    \end{tabular}
    \vspace{-2mm}
	\caption{Real-to-real domain transfer results with Cityscapes (C), IDD (I), and Mapillary (M). Red boxed images are the input.}
	\label{fig:Real2Real}
 	\vspace{-3mm}
\end{figure}

\noindent\textbf{Implementation Details}
We use the Deeplabv2+ResNet-101 \cite{chen2017deeplab,he2016deep} architecture for our segmentation network, as used in other conventional works \cite{isobe2021multi, saporta2021multi}.
We use the same encoder and generator structure of DRANet \cite{lee2021dranet} with group normalization \cite{wu2018group}.
For our multi-head discriminator, we use the patchGAN discriminator \cite{isola2017image} and two fully connected layers as the domain classifier.
We use ImageNet-pretrained VGG19 networks \cite{simonyan2014very} as the perceptual network and compute the perceptual loss at layer relu$\_$4$\_$2.
We use a stochastic gradient descent optimizer\cite{bottou2010large} with a learning rate of $2.5 \times 10^{-4}$, a momentum of 0.9 and a weight decay of $5 \times 10^{-4}$ for training the segmentation network. 
We use Adam\cite{kingma2014adam} optimizer with a learning rate of $1 \times 10^{-3}$, momentums of 0.9 and 0.999 and a weight decay of $1 \times 10^{-5}$ for training all the networks in MTDT-Net. 
% We use a stochastic gradient descent optimizer\cite{bottou2010large} with a learning rate of $2.5 \times 10^{-4}$ (momentum=0.9, weight decay=$5 \times 10^{-4}$) for training the segmentation network. 
% We use the Adam\cite{kingma2014adam} optimizer with a learning rate of $1 \times 10^{-3}$  ($\beta_1$=0.9, $\beta_2$=0.999, weight decay=$1 \times 10^{-5}$) for training all the networks in MTDT-Net.

% \begin{table}[t]
%     \centering
%     \begin{tabular}{c@{\hspace{2mm}}c|c|ccc|c}
%     & & & \multicolumn{3}{c|}{mIoU} & mIoU \\
%     \small{$\mathcal{Y}_S^{BARS}$} & \small{$\hat{\mathcal{Y}}_{T_k}^{BARS}$} & Mapping & C & I & M & Avg. \\
%     \hline
%     \multirow{2}{*}{}& & 19 &&&& \\
%     &&7&&&& \\
%     \hline
%     \multirow{2}{*}{\checkmark} &&&&&& \\
%     &&&&&& \\
%     \hline
%     & \multirow{2}{*}{\checkmark} &&&&& \\
%     &&&&&& \\
%     \hline
%     \multirow{2}{*}{\checkmark} & \multirow{2}{*}{\checkmark} &&&&& \\
%     &&&&&& \\
     
%     \end{tabular}
%     \caption{Ablation study}
%     \label{tab:ablation}
% \end{table}

\subsection{Synthetic-to-Real Adaptation}
\label{sec:syn2real}

We conduct the experiments on synthetic-to-real adaptation with the same settings as competitive methods \cite{saporta2021multi, isobe2021multi}.
We use GTA5 as the source dataset and a combination of Cityscapes, IDD and Mapillary as multiple target datasets.
We show the qualitative results of the multi-target domain transfer in \figref{fig:teaser}.
This demonstrates that our single MTDT-Net can synthesize high quality images even in multi-target domain scenarios. 
We report the quantitative results for semantic segmentation with 7 common classes in \tabref{tab:7classes}, and 19 classes in \tabref{tab:19classes}, respectively.
The results show that our method, composed of both MTDT-Net and BARS outperforms state-of-the-art methods by a large margin. 
Compared to ADVENT\cite{vu2019advent} calculating selection criterion value from incorrect target prediction, our BARS derives the criterion robustly from accurate source GT without class ambiguity.
Moreover, MTDT-Net aims to transfer visual attributes of domains rather than adapting color information using a color transfer algorithm \cite{reinhard2001color} proposed in CCL\cite{isobe2021multi}.
The new attribute alignment method improves the task performance over state-of-the-art methods.
%more adequate visual attributes of target domain than CCL\cite{isobe2021multi} that adapts only color information via color transfer algorithm\cite{reinhard2001color}.
%that is not enough to involve the distinctiveness of each domain.
%These difference lead to performance improvement compared to competitive methods.
Lastly, the qualitative results in \figref{fig:seg} demonstrates that our method produces reliable label prediction maps on both label mapping protocols.

\subsection{Real-to-Real Adaptation}
\label{sec:real2real}
To show the scalability of our model, we also conduct an experiment with real-to-real adaptation scenarios.
We set one of the real-world datasets, Cityscapes, IDD, and Mapillary, as the source domain and the other two as the target domains.
We conduct the experiments with all possible combinations of source and target.
The results in \figref{fig:Real2Real} show that our MTDT-Net also produces high fidelity images even across real-world datasets. 
As shown in \tabref{tab:real2real}, our method outperforms the competitive methods in the overall results.
The experiments demonstrate that our method achieves realistic image synthesis not only on synthetic-to-real but also on real-to-real adaptation, which validates the scalability and reliability of our model.

% generality를 증명하기 위해서 real to real adaptation실험도 했다. 
% 실험에 대한 내용. real dataset에 대해서 사용했다. 카메라 위치, 도시, illumination등이 다른 데이터셋에 각각 multi target domain adaptation을 진행했다.
% 우리 모델은 real to real domain transfer에서도 consistently 좋은 결과를 얻었다.
% 테이블 3에 나와있는 것 처럼 정량적으로도 좋은 평가를 얻었고 모든 실험에서 sota 방법들을 이겼다.
% 이 실험을 통해서 우리 모델은 generality가 잘된다는 걸 보였다.

% 표를 보고 뭔 말을 하기가 쫌.. 

%- 우리 model이 generalization을 확인하기 위해 real world에서의 adaptation 실험 및 비교.
%- 마찬가지로 19 class, 7 class 모두 비교.
%- C->IM, I->CM, M->CI 각각 MTDT-Net 학습. Figure7에서 알 수 있듯이 real-to-real에서도 퀄리티 좋음.
%- adaptation 결과 Table 3.
%- generalize 가능한 model임을 알 수 있음.

\begin{table}[t]
    \centering
    {\footnotesize
    \begin{tabular}{c|c|c|ccc|c}
    \toprule
        & \# of & \multirow{2}{*}{Method} & \multicolumn{3}{c|}{mIoU} & mIoU \\
        & classes & & C & I & M & Avg.\\
        \hline
        \multirow{4}{*}{\begin{turn}{90}C$\to$I, M\end{turn}}& \multirow{2}{*}{19} & CCL\cite{isobe2021multi} & - & \textbf{53.6} & 51.4 & \textbf{52.5} \\
        & & Ours & - & 48.3 & \textbf{53.6} & 50.5 \\
        \cline{2-7}
        & \multirow{2}{*}{7}& MTKT\cite{saporta2021multi} & - & 68.3 & 69.3 & 68.8 \\
        & & Ours & - & \textbf{70.4} & \textbf{75.1} & \textbf{72.7} \\
        \hline
        \multirow{4}{*}{\begin{turn}{90}I$\to$C, M\end{turn}}& \multirow{2}{*}{19} & CCL\cite{isobe2021multi} & 46.8 & - & 49.8 & 48.3 \\
        & & Ours & \textbf{49.1} & - & \textbf{50.8} & \textbf{50.0} \\
        \cline{2-7}
        & \multirow{2}{*}{7}& MTKT\cite{saporta2021multi} & - & - & - & - \\
        & & Ours & \textbf{79.5} & - & \textbf{77.9} & \textbf{78.7} \\
        \hline
        \multirow{4}{*}{\begin{turn}{90}M$\to$C, I\end{turn}}& \multirow{2}{*}{19} & CCL\cite{isobe2021multi} & 58.5 & \textbf{54.1} & - & 56.3 \\
        & & Ours & \textbf{58.7} & \textbf{54.1} & - & \textbf{56.4} \\
        \cline{2-7}
        & \multirow{2}{*}{7}& MTKT\cite{saporta2021multi} & - & - & - & - \\
        & & Ours & \textbf{75.8} & \textbf{81.1} & - & \textbf{78.5} \\
    \bottomrule
    \end{tabular}
    }
    \caption{Results of real-to-real MTDA on all possible combinations among Cityscape (C), IDD (I), and Mapillary (M). 
    % The splitted rows for each domain combination show the quantitative comparison between our model and the state-of-the-art method on 19 classes (upper row) and 7 classes (lower row) setting, respectively.
    }
    \label{tab:real2real}
    \vspace{-3mm}
\end{table}

\subsection{Further Study on MTDT-Net and BARS}
\label{sec:Further}
In this section, we conduct additional experiments to validate each sub-module, MTDT-Net and BARS.

\noindent\textbf{MTDT-Net}
We compare our MTDT-Net with a color transfer algorithm \cite{reinhard2001color} used in CCL \cite{isobe2021multi} and DRANet \cite{lee2021dranet} which are the most recent multi-domain transfer methods.
We conduct the experiment on a synthetic-to-real adaptation using GTA5, Cityscapes, IDD and Mapillary as in \secref{sec:syn2real}. 
We train the task network using synthesized images from each method with corresponding source labels. \tabref{tab:ablation_MTDT} shows the results for semantic segmentation with 19 classes setting.
% \figref{fig:DT_comparison} shows the domain transfer results of each method.
% The results in the second column show that DRANet fails to transfer the source image to each target domain, and we can even find some artifacts in the images.
Among the competitive methods, MTDT-Net shows the best performance. 
We believe the other two methods hardly transfer the domain-specific attribute of each target dataset.
The color transfer algorithm just shifts the distribution of the source image to that of the target image in color space, rather than aligning domain properties.
DRANet tries to cover the feature space of each domain using just one parameter, called the domain-specific scale parameter, resulting in unstable learning with multiple complex datasets.
On the other hand, MTDT-Net robustly synthesizes the domain transferred images by exploiting the target feature statistics, which facilitate better domain transfer.

% In \figref{fig:DT_comparison}, images in the first row are the output of DRANet, lower are output of MTDT-Net. (FIG 8 update needed)
% DRANet often fails to converge with multiple driving datasets in our experiments, because the model heavily depends on the domain normalization parameter.
% They try to cover the feature space of each domain using just one parameter, which results in unstable learning with multiple complex datasets.
% Moreover, in g2c, g2m images, there is the mixed content artifact in the sky regions. (reason?)
% 컨텐츠와 스타일의 완벽한 disentangling이 안돼서?
% Mixed content artifact도 생긴다!
%On the other hand, our MTDT-Net produces well-made images of multiple domains, because it stores domain statistics for each domain and transfer the style of the source domain to the target domain via DST, enabling powerful image transformation.
% On the other hand, MTDT-Net stores domain statistics for each domain and transfer the style of the source domain to the target domain via DST, so it can robustly generate well-made images of multiple domains.

% - 반면 우리 모델은...
% 여러 도메인의 이미지들을 대체로 잘 생성한다. 이는 각 도메인마다 domain statistic을 저장해놓고 DST를 이용하여 source domain의 style을 domain transferr하기 때문에 robust한 이미지 변환이 가능하기 때문이다.

\noindent\textbf{BARS}
To validate the effectiveness of the two filtered labels  $Y_{\mathcal{S}\to{\mathcal{T}_k}}^{BARS}$ and $\hat{Y}_{\mathcal{T}_k}^{BARS}$, we conduct a set of experiments with/without each component.
We train the segmentation network with the output images of MTDT-Net using a full source label in the experiments without $Y_{\mathcal{S}\to{\mathcal{T}_k}}^{BARS}$.
With just the $Y_{\mathcal{S}\to{\mathcal{T}_k}}^{BARS}$ or $\hat{Y}_{\mathcal{T}_k}^{BARS}$, the model achieves large improvements in \tabref{tab:ablation}, respectively.
% The experiment with either component shows better performance than the method without BARS. 
However, the region with ambiguous or noisy labels limits the model performance, so the network trained with both filtered labels achieves the best performance.

\begin{table}[]
    \centering
    {\small
    \begin{tabular}{c|ccc|c}
    \toprule
        & \multicolumn{3}{c}{mIoU} & mIoU \\
        Method & C & I & M & Avg. \\
        \hline
        Color Transfer \cite{reinhard2001color} & 33.8 & 37.4 & 42.1 & 37.8 \\
        DRANet \cite{lee2021dranet} & 37.3 & 39.3 & 43.2 & 39.9 \\
        MTDT-Net & \textbf{41.4} & \textbf{40.6} & \textbf{44.1} & \textbf{42.0} \\
    \bottomrule
    \end{tabular}
    }
    \caption{Comparison of MTDA-Net with competitive methods on synthetic-to-real adaptation with 19 classes setting.}
    \label{tab:ablation_MTDT}
    \vspace{1mm}
\end{table}

\begin{table}[t]
    \centering
    {\small
    \begin{tabular}{c@{\hspace{2mm}}c|ccc|c}
    \toprule
    & &  \multicolumn{3}{c|}{mIoU} & mIoU \\
    \small{$Y_{\mathcal{S}\to {\mathcal{T}_k}}^{BARS}$} & \small{$\hat{Y}_{\mathcal{T}_k}^{BARS}$} & C & I & M & Avg. \\
    \hline
    && 41.4 & 40.6 & 44.1 & 42.0 \\
    \checkmark && 43.1 & 44.0 & 46.9 & 44.7 \\
    & \checkmark & 45.0 & 44.9 & 47.5 & 45.8 \\
    \checkmark & \checkmark & \textbf{46.9} & \textbf{47.7} &  \textbf{51.1} & \textbf{48.6} \\
    \bottomrule
     
    \end{tabular}
    }
    \caption{Ablation study of BARS on synthetic-to-real adaptation with 19 classes setting.}
    \label{tab:ablation}
    \vspace{-3mm}
\end{table}

% Conclusion
%%%%%%%%% Conclusion
\section{Conclusion}
\label{sec:conclusion}
In this paper, we present ADAS, a new approach for multi-target domain adaptation, which directly adapts a single model to multiple target domains without relying on the STDA models.
For the direct adaptation, we introduce two key components: MTDT-Net and BARS.
MTDT-Net enables a single model to directly transfer the distinctive properties of multiple target domains to the source domain by introducing the novel TAD ResBlock.
BARS helps to remove the outliers in the segmentation labels of both the domain transferred images and the corresponding target images.
Extensive experiments show that MTDT-Net synthesizes visually pleasing images transferred across domains, and BARS effectively filters out the inconsistent region in segmentation labels, which leads to robust training and boosts the performance of semantic segmentation.
The experiments on benchmark datasets demonstrate that our method designed with MTDT-net and BARS outperforms the current state-of-the-art MTDA methods.

\vspace{2mm}

{\footnotesize
\noindent \textbf{Acknowledgement} This work was supported by Institute of Information \& Communications Technology Planning \& Evaluation(IITP) grant funded by the Korea government(MSIT) (No.2014-3-00123, Development of High Performance Visual BigData Discovery Platform for Large-Scale Realtime Data Analysis), and the National Research Foundation of Korea (NRF) grant funded by the Korea government (MSIT) (No. 2020R1C1C1013210).
}

%%%%%%%%% REFERENCES
\clearpage
{\small
\bibliographystyle{ieee_fullname}
\bibliography{egbib}
}

% \clearpage
% \noindent0. Domain\\
% - Source : $\mathcal{S}$\\
% - Target : $\mathcal{T}$\\\\
% 1. Input/Output data\\
% - Images : $I, I', I"$\\
% - Labels : $Y$\\\\
% 2. Models\\
% - Encoder : $E$\\
% - Generator : $G$\\
% - Style Encoder : $SE$\\
% - Domain Style Transfer module : $DST$\\
% - Task network : $T$\\
% - Discriminator : $D = \{D_P, D_C\}$\\
% - Perceptual network : $P$\\\\
% 3.Functions\\
% - Label embedding : $\phi$\\
% - Target Adaptive Denormalization : $TAD$\\
% - Fully Connected layer : $FC$\\
% - Loss : $\mathcal{L}$\\
% - L2 distance : $d$\\
% - Indicator function : $\mathbbm{1}$\\\\
% 4. Intermediate features\\
% - Encoder feature : $\mathcal{F}$\\
% - Task feature : $\mathcal{F^T}$\\
% - Content : $\mathcal{C}$\\
% - Centroid : $\mathcal{\dot{C}}$\\
% - gamma, beta, mean, variance : $\gamma, \beta, \mu, (\sigma)^2$\\
% - Encoder feature : $\hat{f}_\mathcal{S}$\\\\
% 5. etc\\
% - \# of targets : $N$\\
% - Pixel \# of class : $N_c$\\
% - Total \# of iterations : $N_{train}$\\
% - \# of layers : $L$\\
% - Indice of spatial coordinate : $i, j$\\

\end{document}